\documentclass[10pt,twocolumn,letterpaper]{article}

\usepackage{cvpr}              % To produce the CAMERA-READY version
\usepackage{graphicx}
\usepackage{amsmath}
\usepackage{amssymb}
\usepackage{booktabs}
\usepackage{algorithm}
\usepackage{algorithmic}
\usepackage[algo2e]{algorithm2e}
\usepackage{enumerate}
\usepackage{enumitem}
\usepackage{makecell}
\usepackage[numbers,sort&compress]{natbib}
\usepackage[usenames, dvipsnames]{color}
\usepackage{color}
\usepackage{subcaption,graphicx}
\usepackage[normalem]{ulem}
\usepackage[table]{xcolor}
\usepackage[usenames, dvipsnames]{color}
\usepackage{pifont}% http://ctan.org/pkg/pifont
\usepackage{color, colortbl}

\definecolor{Gray}{gray}{0.93}

\usepackage[pagebackref,breaklinks,colorlinks]{hyperref}

\usepackage[capitalize]{cleveref}
\crefname{section}{Sec.}{Secs.}
\Crefname{section}{Section}{Sections}
\Crefname{table}{Table}{Tables}
\crefname{table}{Tab.}{Tabs.}

\def\cvprPaperID{1752} % *** Enter the CVPR Paper ID here
\def\confName{CVPR}
\def\confYear{2022}

\begin{document}
\setlength{\abovedisplayskip}{3pt}
\setlength{\belowdisplayskip}{3pt}
\title{Distill and De-bias: Mitigating Bias in Face Verification using Knowledge Distillation}

\author{Prithviraj Dhar\textsuperscript{1}, Joshua Gleason\textsuperscript{2}, Aniket Roy\textsuperscript{1}, Carlos D. Castillo\textsuperscript{1}, P. Jonathon Phillips\textsuperscript{3}, Rama Chellappa\textsuperscript{1}\\
\textsuperscript{1}Johns Hopkins University,
\textsuperscript{2}Univ. of Maryland, College Park,
\textsuperscript{3}NIST\\
%{\tt\small \{pdhar1,rchella4,\}@jhu.edu, \{akumar14,kkaplan,khushigupta,rakeshr\}@fb.com}
}

\maketitle

%%%%%%%%% ABSTRACT
\begin{abstract}
    Face recognition networks generally demonstrate bias with respect to sensitive attributes like gender, skintone etc. For gender and skintone, we observe that the regions of the face that a network attends to vary by the category of an attribute. This might contribute to bias. Building on this intuition, we propose a novel distillation-based approach called Distill and De-bias (D\&D) to enforce a network to attend to similar face regions, irrespective of the attribute category.  In D\&D, we train a teacher network on images from one category of an attribute; e.g. light skintone. Then distilling information from the teacher, we train a student network on images of the remaining category; e.g., dark skintone. A feature-level distillation loss constrains the student network to generate teacher-like representations. This allows the student network to attend to similar face regions for all attribute categories and enables it to reduce bias. We also propose a second distillation step on top of D\&D, called D\&D++. Here, we distill the `un-biasedness' of the D\&D network into a new student network, the D\&D++ network, while training this new network on all attribute categories; e.g., both light and dark skintones. This helps us train a network that is less biased for an attribute, while obtaining higher face verification performance than D\&D.  We show that D\&D++ outperforms existing baselines in reducing gender and skintone bias on the IJB-C dataset, while obtaining higher face verification performance than existing adversarial de-biasing methods. We evaluate the effectiveness of our proposed methods on two state-of-the-art face recognition networks: ArcFace and Crystalface.
\end{abstract}

%-------------------------------------------------------------------------
\section{Introduction}
%While some works have proposed adversarial strategies to prevent face descriptors from encoding such attributes, removal of sensitive attributes from descriptors cripples the performance of these descriptors in identity recognition.
The accuracy of face recognition networks \cite{taigman2014deepface,ranjan2019fast,deng2018ArcFace,dhar2019measuring,meng2021magface} has significantly improved in the last few years. Because of this, such face recognition systems are being used in a large number of applications. This has raised concerns about bias against sensitive attributes such as age, gender or race. A recent study from NIST \cite{grother2019face} has shown that characteristics such as gender and ethnicity impact the verification and matching performance of existing algorithms. Several works \cite{wang2019racial,amini2019uncovering,krishnapriya2020issues,vangara2019characterizing,nagpal2019deep,phillips2011other,cavazos2021accuracy} have explored the issue of bias against gender, race and skintone in face recognition.\\ 
%Similarly, \cite{buolamwini2018gender} showed that most face-based gender classifiers perform significantly better on male faces with light skintone than female faces with dark skintone.
\begin{figure}
\centering
{\includegraphics[width=\linewidth]{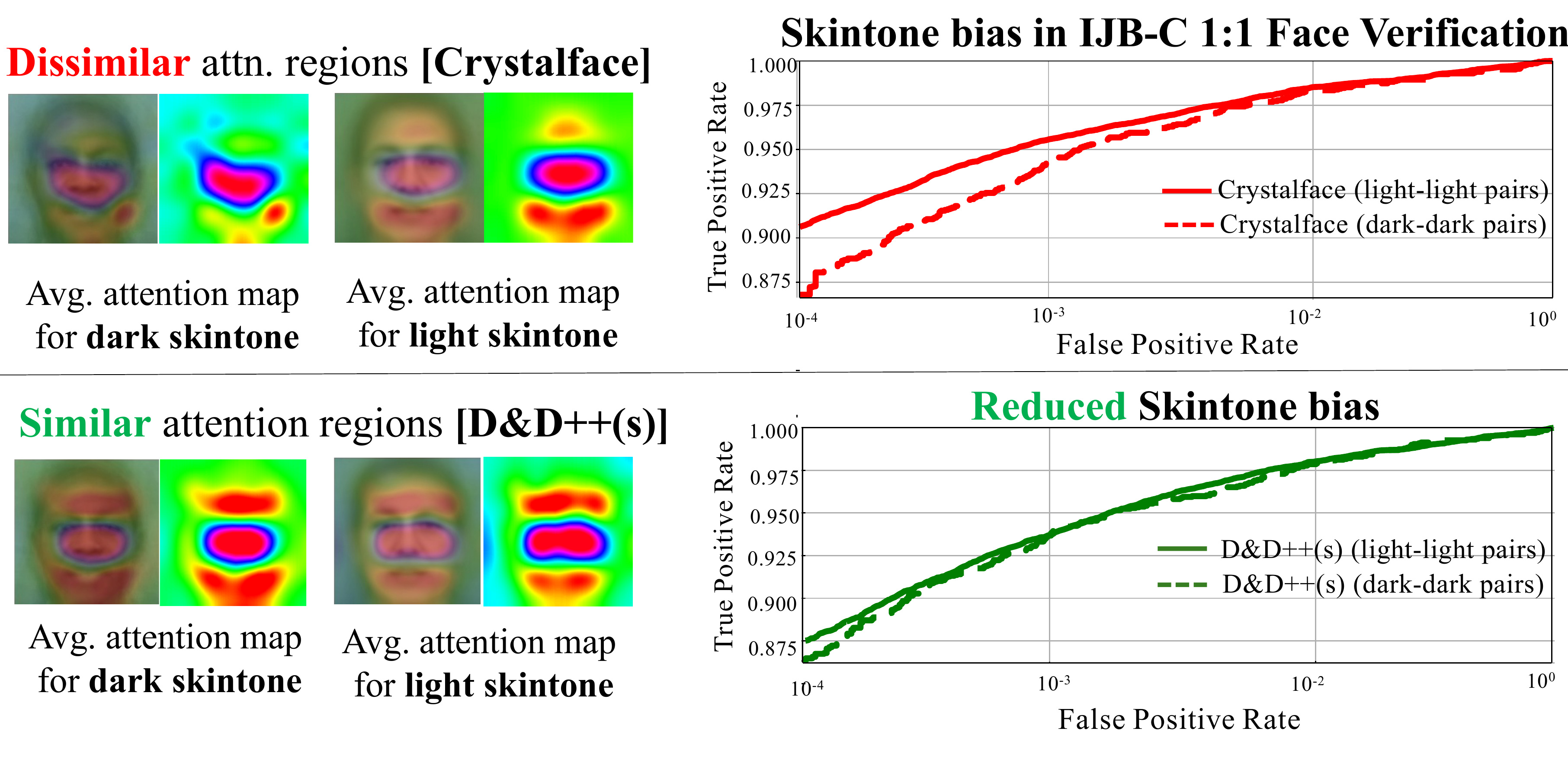}}
\vspace{-1.5em}
\caption{\small \textbf{(Top row)} Face recognition networks attend to different spatial regions in faces, depending on protected attributes (here, shown for skintone attribute). Here, we show the average attention maps generated using the pre-trained Crystalface network for frontal faces with light and dark skintone. This difference in processing faces with light and dark skintone might contribute to skintone bias. \textbf{(Bottom row)} Our proposed method D\&D++ enforces a network to attend to similar spatial regions for both light and dark skintones, and consequently reduces skintone bias. We report similar findings with respect to the gender attribute.}\vspace{-0.5cm}
\label{fig:teaserdnd}
\end{figure}
\-\hspace{\parindent} A possible approach to mitigate gender or skintone bias  would be to re-train a large scale face recognition network on a dataset which is balanced in terms of these attributes. However, as shown in \cite{albiero2020does,wang2019balanced,dhar2020adversarial}, training a network on a balanced dataset does not always lead to unbiased systems. \cite{dhar2020adversarial} points out that while we can balance the dataset in terms of gender or skintone, there exists appearance variation between demographic subgroups with respect to multiple factors such as pose, illumination etc., which may lead to a biased system. Some works \cite{gong2020jointly,Dhar_2021_ICCV} have proposed adversarial strategies to prevent face recognition networks from encoding sensitive attributes like gender and race. However, since gender and race are integral to the face identity, removing such attributes from face recognition features generally reduces their face verification accuracy. Among non-adversarial methods, GAC \cite{gac} proposes an adaptive filtering technique to mitigate racial bias. However, the effectiveness of GAC  applied to other attributes (such as gender) is currently unclear. Similar to GAC, we propose non-adversarial techniques to mitigate bias in face recognition. More specifically, we present two novel knowledge distillation-based techniques called D\&D and D\&D++ to incrementally learn different categories of a given sensitive attribute while significantly reducing bias with respect to that attribute.  We show that our proposed methods can be used to reduce bias with respect to either gender or skintone, and are therefore likely to generalize for other attributes. Our methods also generalize to different face recognition models. 

%While there is some correlation between race and skintone, we believe that race is still ill-defined, as pointed out in \cite{hazirbas2021casual}. For example, it is difficult to quantify the race category of multi-racial faces. Skintone, on the other hand, is more scientifically defined by the Fitzpatrick scale \cite{fitzpatrick1975soleil}. Therefore, inspired by previous work \cite{lu2019experimental, Dhar_2021_ICCV}, we have opted to mitigate skintone bias, as opposed to racial bias. In summary, we make the following contributions in our work:
Buolamwini \textit{et al.} \cite{buolamwini2018gender} introduced skintone as an alternative to race. It can be difficult to quantify the race category of multi-racial faces. Skintone, on the other hand, is more scientifically defined by the Fitzpatrick scale \cite{fitzpatrick1975soleil}. Therefore, following previous works \cite{lu2019experimental,Dhar_2021_ICCV}, we mitigate skintone bias, as opposed to racial bias. In summary, we make the following contributions in our work:
\begin{enumerate}[leftmargin=*]
\item  We observe that face recognition networks attend to different regions of the face, depending on the gender or skintone category. We illustrate this observation with GradCAM \cite{selvaraju2017grad} generated attention maps (Figs. \ref{fig:teaserdnd},\ref{fig:avgmap}, Section \ref{sec:ameffect}). These differences in how algorithms process faces from different demographic categories may lead to bias in face recognition.
\item Building on this observation, we propose a method, called \textbf{D}istill and \textbf{D}e-bias (D\&D), that enforces a network to attend to similar spatial regions in both male and female faces (and in both faces with light and dark skintones). We show the ability of D\&D to reduce gender and skintone bias in two state-of-the-art face recognition networks:  ArcFace \cite{deng2018ArcFace} and Crystalface \cite{ranjan2019fast}. To the best of our knowledge, \textit{we are the first to use knowledge distillation for designing bias mitigation strategies in face recognition}.
\item We propose D\&D++ to further improve the face verification performance, while inheriting the `unbiasedness' of D\&D. D\&D++, while being de-biased, achieves higher face verification performance than state-of-the-art adversarial de-biasing methods on the IJB-C \cite{maze2018iarpa} dataset. 
\end{enumerate}
\section{Related Work}
\textbf{Bias in face recognition:} Empirical studies \cite{grother2019face,buolamwini2018gender,drozdowski2020demographic,nagpal2020diversity,puc2021analysis,singh2021anatomizing,majumdar2021attention,chen2021understanding} have shown that many publicly available systems performing face recognition or face analysis demonstrate bias towards sensitive attributes such as race and gender. With respect to gender bias, \cite{albiero2020does,lu2019experimental} show that the performance of face recognition on females is lower than that of males. Use of cosmetics by females \cite{cook2019fixed,klare2012face,albiero2021gendered} and gendered hairstyles \cite{albiero2020face} has been assumed to play a major role in the resulting gender bias. However, \cite{albiero2020analysis} shows that cosmetics only play a minor role in the gender gap. \cite{wang2019racial,wang2020mitigating,gac,yang2021ramface,xu2021consistent} explore the issue of racial bias in face recognition, and propose strategies to mitigate the same. \cite{lu2019experimental} shows that face verification systems perform better on lighter skintones than darker skintones. \cite{dhar2020attributes,hill2019deep} show that face recognition networks implicitly encode information about sensitive attributes during training, which might lead to bias w.r.t these attributes.\\
\begin{table}
\scalebox{0.95}{
\begin{tabular}{ccc}
\hline
Method & Target task & Sensitive attribute\\ 
\hline
\cite{alvi2018turning}\hspace{-4pt} & Gender/Age pred. & Age/Gender\\
\cite{li2019deepobfuscator} & Smile, high-cheeks & Gender, make-up\\
\cite{amini2019uncovering} & Face detection & Skintone\\
%\cite{quadrianto2019discovering} & Face attractiveness & Gender\\
\cite{wang2019racial} & Face recognition & Race\\
\cite{xu2020investigating}&Expression pred.&Age, gender, race\\
\cite{gong2020jointly}&Face recognition&Age, gender, race\\
\cite{park2021learning}&Attractiveness pred.&Gender,age\\
\cite{robinson2021balancing} & Face recognition & Gender, race\\
\cite{Dhar_2021_ICCV} & Face recognition & Gender, skintone\\
\bottomrule
\end{tabular}
}
\vspace{-0.3cm}
\caption{\small Methods that adversarially remove sensitive attributes to reduce bias with respect to these attributes in the target task. \vspace{-0.3cm}}
\label{tab:rel}
\vspace{-15pt}
\end{table}
\textbf{Building fairer datasets:} It has been speculated that the imbalanced training datasets might lead to bias in face recognition. However, \cite{albiero2020does} shows that the gender bias is not mitigated when equal number of male and female identities are used for training. Similarly, \cite{gwilliam2021rethinking} shows that a network trained on racially balanced datasets does not necessarily obtain the lowest racial bias. \cite{hazirbas2021casual} introduces a dataset to measure the robustness of AI models to a diverse set of genders and skintones. \cite{robinson2020face} presents a race and gender-balanced evaluation dataset and provides the verification protocol for the same. \\
\textbf{Adversarial de-biasing:} Several researchers have proposed adversarial strategies to reduce the encoding of sensitive attributes (to reduce the bias with respect to these attributes), while performing a face-based target task. We provide a brief summary of these works in Table \ref{tab:rel}. \cite{gong2020jointly,park2021learning,Dhar_2021_ICCV} have reported that the face verification performance of the adversarially-debiased systems in the target task decreases due to the removal of sensitive attributes such as gender. \\
\textbf{Knowledge distillation (KD):} KD \cite{44873} has been primarily applied to continual learning tasks \cite{li2017learning,rebuffi2017icarl}. In a KD framework, a student network (initialized using a pre-trained teacher network) is trained to learn new classes or tasks that are not recognizable by the teacher network, and mimic the output score distribution of the teacher model for preventing forgetting of the classes/tasks that the teacher network was trained on. Several works \cite{romero2014fitnets,zagoruyko2016paying,dhar2019learning} have shown that directly matching feature activations of teacher and student networks from their intermediate layers is also an effective way to distill the teacher's knowledge into the student. This finding is used in \cite{dhar2021eyepad++} that proposes a two step sequential distillation process for disjoint multitask learning (MTL). Even though our aim and motivation is different from \cite{dhar2021eyepad++}, we use a similar sequential distillation pipeline, wherein we employ feature-level KD in our work to enforce a student trained on a specific attribute category (e.g. females) to mimic a teacher that is trained on a different attribute category (e.g. males). \cite{jung2021fair} also uses KD for mitigating bias with respect to sensitive attributes (e.g. gender)  that are independent of target task (e.g. attractiveness prediction).
\section{Problem Statement}
\label{sec:probstatement}
Given a binary face attribute $A$ with categories $a_1$ and $a_2$, our goal is to enforce a network to process faces with $A=a_1$ and $A=a_2$ in a similar way. We hypothesize that \textit{a network that processes faces with attribute $A=a_1$ and faces with attribute $A=a_2$ in a similar way will demonstrate lower bias with respect to attribute $A$}. For attribute $A=\text{skintone}$, $(a_1, a_2)=$(Light, Dark). For attribute $A=\text{gender}$, $(a_1, a_2)=$(Male, Female).\\
%In our preliminary experiments (in Sec. \ref{subsec:prednbias}), we find that face descriptors from which the gender/skintone is more difficult to predict generally demonstrate lower gender/skintone bias in face verification tasks. Therefore, we hypothesize that \textit{reducing the ability to predict protected attributes (gender and skintones) in face descriptors will reduce gender/skintone bias in face verification tasks}. To test this hypothesis, we present an adversarial technique that transforms face descriptors so that they can be used to accurately classify identity but not gender/skintone. Reducing the predictability of gender and skintone in face descriptors will impede malicious attribute classifiers, thus reducing the possibility of leakage of protected information in face descriptors. \\
\textbf{Bias measure}: Following \cite{Dhar_2021_ICCV}, we define gender and skintone bias, at a given false positive rate (FPR) as follows:  
 \begin{equation}
     \text{Gender Bias}^{(F)} = |\text{TPR}_{m}^{(F)} - \text{TPR}_{f}^{(F)}|
 \label{eq:gbias}
 \end{equation}
 \begin{equation}
     \text{Skintone Bias}^{(F)} = |\text{TPR}_{l}^{(F)} - \text{TPR}_{d}^{(F)}|
 \label{eq:stbias}
 \end{equation}
 where $(\text{TPR}_{m}^{(F)}, \text{TPR}_{f}^{(F)}, \text{TPR}_{l}^{(F)}, \text{TPR}_{d}^{(F)})$ denote the true positive rates for the verification of male-male, female-female, light-light and dark-dark pairs respectively at FPR $F$. Our goal is to train face recognition networks that reduce the bias (Eq. \ref{eq:gbias} or \ref{eq:stbias}), while maintaining reasonable face verification performance in face verification. The reasoning for using this measure instead of difference between AUC or EER is provided in \cite{Dhar_2021_ICCV}.\\
 %Some works such as \cite{gong2020jointly} evaluate bias as the difference between area under ROC curves (AUC), but AUC fails to capture the bias at low FPRs. Since most real world verification systems tend to operate at very low FPR \cite{ngan2020ongoing}, we focus on FPR values and not AUC. \\
 %In the supplementary material, we add a novel interpretation of these bias measures and show that it may be viewed as a measure of \textit{equality of odds} \cite{hardt2016equality}.
%In some works such as \cite{gong2020jointly}, bias is evaluated as the difference between area under ROC curves (AUC). While this can be viewed as an aggregate of our measure, such an aggregation fails to meaningfully capture the bias at realistic operating points as it marginalizes the performance at low FPR. In our experience, most real world verification systems tend to operate at very low FPR, i.e. less than $10^{-4}$, which is not meaningfully captured with AUC. In this work, we focus on FPR values that we consider to be realistic operating conditions.
\textbf{Measuring bias/performance trade-off:}
%Several methods that reduce bias demonstrate a slight drop in the overall performance of the system \cite{Dhar_2021_ICCV, gong2020jointly}.
We also adopt the tradeoff measure called \textit{bias performance coefficient} (BPC) from \cite{Dhar_2021_ICCV}. This is a measure of the trade-off between bias reduction and drop in face verification performance and is defined as
\begin{equation}
\label{eq:bpc}
\text{BPC}^{(F)}= \frac{\text{Bias}^{(F)}-\text{Bias}^{(F)}_{deb} }{\text{Bias}^{(F)}}-\frac{\text{TPR}^{(F)} - \text{TPR}^{(F)}_{deb} }{\text{TPR}^{(F)}}.
\vspace{-0.1cm}
\end{equation}

 Here, ($\text{TPR}^{(F)},\text{Bias}^{(F)}$) refer to the \textit{overall} TPR obtained by original features and the corresponding bias (Gender/Skintone bias) at FPR of $F$.  ($\text{TPR}^{(F)}_{deb}$, $\text{Bias}^{(F)}_{deb}$) denote their de-biased counterparts. We aim \textit{to build systems that achieve high BPC values}, since a higher BPC denotes high bias reduction and low drop in face verification performance. The original network (without any de-biasing) would have a zero BPC (since $\text{Bias}^{(F)}=\text{Bias}^{(F)}_{deb}$ and $\text{TPR}^{(F)}=\text{TPR}^{(F)}_{deb}$). We note that a system with $\text{TPR}^{(F)}_{deb} >> \text{TPR}^{(F)}$ and  $\text{Bias}^{(F)}_{deb} > \text{Bias}^{(F)}$ can have a high BPC value, which is not desirable. But, as pointed out in \cite{Dhar_2021_ICCV}, most de-biasing systems have  $\text{TPR}^{(F)}_{deb} < \text{TPR}^{(F)}$ and  $\text{Bias}^{(F)}_{deb} < \text{Bias}^{(F)}$. An ideal system that reduces the bias to 0 and does not reduce the TPR will have BPC $=1$. A negative BPC denotes that the percentage drop in TPR is higher than the percentage reduction in bias. We denote the BPC for skintone as `BPC\textsubscript{st}' and that for gender as `BPC\textsubscript{g}'.\vspace{-0.25cm}
\section{Proposed approach}
%\vspace{-0.35cm}
\subsection{Motivation}
\label{subsec:approach}
%\vspace{-0.25cm}
Let us consider a binary face attribute $A$ with two categories: $a_1$ and $a_2$. Suppose that we have a face image $I_1$, with attribute $A=a_1$ and image $I_2$ with attribute $A=a_2$. We hypothesize that a relatively unbiased face recognition network must attend to similar facial regions for both $I_1$ and $I_2$, irrespective of their attribute $A$ categories. However, our initial experiments show that the spatial regions that a network attends to vary for different genders (Figs. \ref{fig:arcamgender},\ref{fig:amgender}-first rows), and for different skintones (Figs. \ref{fig:arcamst},\ref{fig:amst}-first rows). A network that does not attend to similar face regions for males and females (or faces with light and dark skintones) might exhibit bias, as shown in Fig. \ref{fig:teaserdnd} (top row). More details regarding these results are presented in Section \ref{sec:ameffect}. From these results, it appears that male and female images are processed differently by a face recognition network. Similarly, images with light and dark skintones are also processed differently. Therefore, we propose KD-based methods that enforce a network to process faces from all attribute categories in a similar way, so that the network attends to similar regions of the face, irrespective of the attribute category. 
\begin{figure}
\centering
{\includegraphics[width=\linewidth]{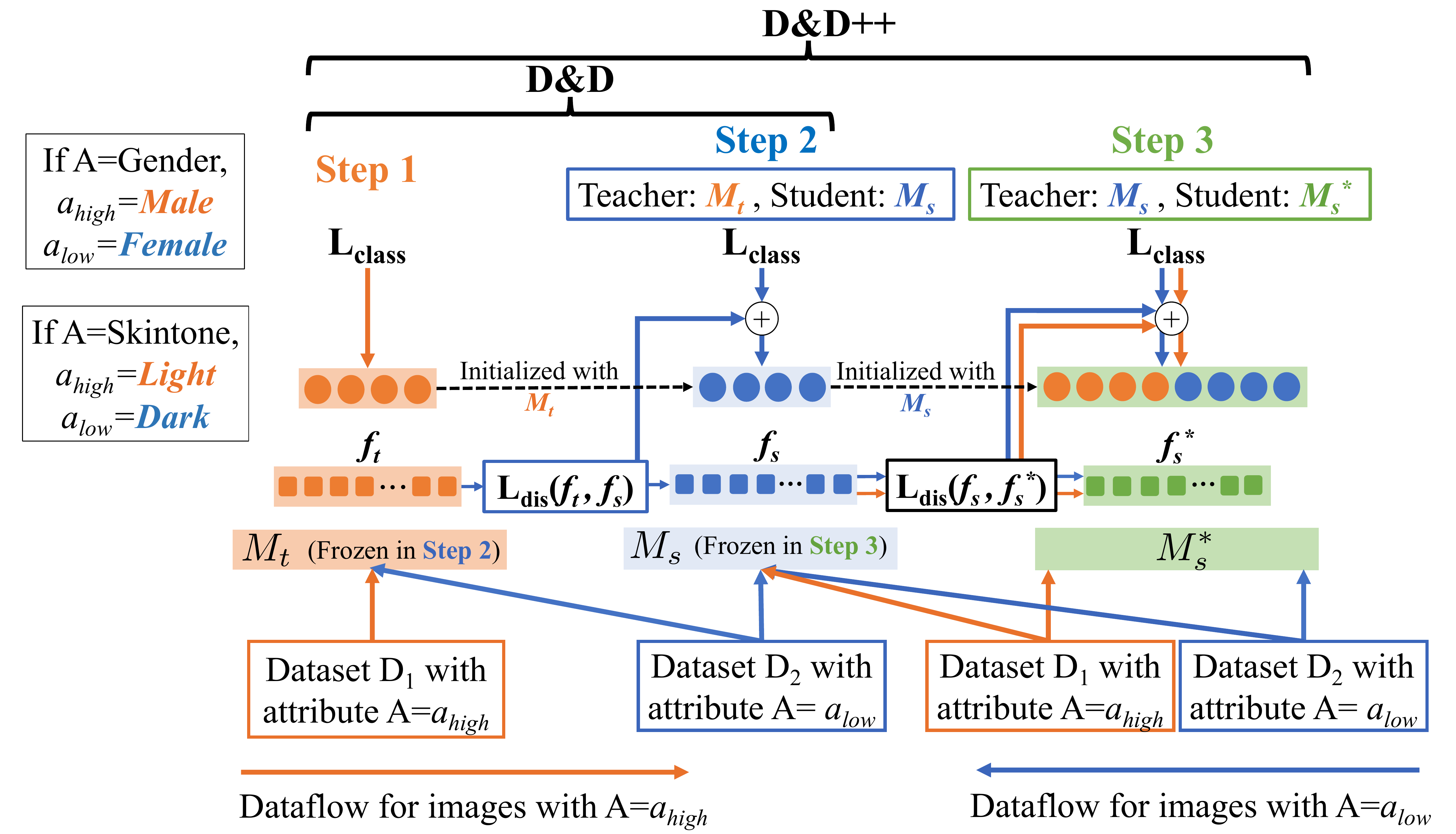}}
\vspace{-0.4cm}
\caption{\small Proposed approach. \textbf{Step 0}: We binarize attribute $A$ and assign $a_{high},a_{low}$ \textbf{Step 1}: We train $M_t$ on faces with $A=a_{high}$. \textbf{Step 2 (D\&D)}: We initialize $M_s$ using $M_t$ and train it using faces with $A=a_{low}$. Here, $M_s$ is enforced to generate teacher-like features using $L_{dis}$, and thus process faces belonging any attribute category ($a_{high}$ or $a_{low}$) in a similar way. \textbf{Step 3 (D\&D++)}: We initialize a new student network $M^{*}_s$ using $M_s$. $M^{*}_s$ is trained on the entire dataset to improve its recognition performance, while it inherits the `unbiasedness' of $M_s$ using $L_{dis}$. }\vspace{-0.5cm}
\label{fig:dndplus}
\end{figure}
\vspace{-0.35cm}
\subsection{Distill and De-bias (D\&D) and D\&D++}
\label{subsec:dnd}
%Thus, $M_s$ would be able to classify faces of both categories $a_1$ and $a_2$. More importantly, since $M_s$ would be using a single reference for processing faces, it would attend to similar spatial regions in face images, irrespective of attribute category $a_1$ or $a_2$. This is a potential way of reducing bias.
%Suppose we have a teacher network $M_t$ that is trained using images with attribute $A=a_1$. Our goal is to train a student network $M_s$ using images with $A=a_2$, such that $M_s$ learns to process the faces with $A=a_2$ in the same way $M_t$ processes a face with $A=a_1$. This is done by distilling information about how $M_t$ processes a face to $M_s$. 
Here, we explain the steps in D\&D and D\&D++ (Fig. \ref{fig:dndplus}):  \\
\textbf{Step 0}:~\textbf{(i) Binarizing attribute $A$:} We start with a binary attribute $A$ (such as gender). If attribute $A$ is non-binary, we regroup its categories into two categories. We explain this in more detail in Sec. \ref{subsec:archdata}, as we re-group the non-binary `race' attribute in the training dataset into binary skintone attribute.

\textbf{(ii) Assigning $a_{high}, a_{low}$:} Assign $a_{high}$ to be the category for which the face verification performance of a network is higher, and $a_{low}$ to be the remaining category. Several works \cite{albiero2020does,lu2019experimental,Dhar_2021_ICCV} have shown that the performance of a network trained on unconstrained datasets is better for males as compared to females. Hence, for $A=$Gender, $a_{high}$=Male and $a_{low}$=Female. Similarly, \cite{lu2019experimental,Dhar_2021_ICCV} show that these networks perform better for faces with light skintone than those with dark skintone. So, for $A=$skintone, $a_{high}=$Light, and $a_{low}$=Dark. \\
\textbf{Step 1}: We train the teacher network $M_t$ using faces with attribute $A=a_{high}$. \\
\textbf{Step 2 (Feature-level knowledge distillation - D\&D)}: We initialize a student network $M_s$ using $M_t$, and train it using images with $A=a_{low}$. Let $I$ be such an image with $A=a_{low}$. $I$ is fed to both $M_t$ and $M_s$, to obtain features $f_t$ and $f_s$, extracted from the penultimate layer of the corresponding networks. To enforce the student $M_s$ to mimic $M_t$'s way of processing faces, we employ feature-level knowledge distillation to distill $M_t$'s knowledge of processing faces into $M_s$. We define the distillation loss $L_{dis}$ as
\begin{equation}
    L_{dis}(f_s,f_t) =1 - \frac{f_s\cdot f_t}{\|f_s\|\|f_t\|}.
\end{equation}
 To constrain $M_s$ to process a face like $M_t$ would, we minimize the cosine distance between $f_t$ and $f_s$ using $L_{dis}$. Our application of feature-level KD is inspired by \cite{romero2014fitnets}. Note that we do not apply KD on the output scores as done in \cite{li2017learning}, since face verification protocols like \cite{maze2018iarpa} use the face recognition features from the penultimate layer (and not the output score vector). Additionally, we would like $M_s$ to classify identities using $L_{class}$. This is the standard cross-entropy loss. Combining these constraints, we train $M_s$ using the bias reducing classification loss $L_{br}$ as
\begin{equation}
    L_{br} = L_{class} + \lambda_1 L_{dis},
\end{equation}

where $\lambda_1$ is used to weight $L_{dis}$. In this step, the teacher $M_t$ remains frozen. The distillation step helps in two ways:
\begin{enumerate}[leftmargin=*]
    %\item \textit{This allows $M_s$ to classify and process faces with $A=a_1$ as $M_t$ would process faces with $A=a_1$, even without being trained on the $a_1$ category.} This is because $M_s$ is initialized with $M_t$ (which was trained on faces with $A=a_1$) and $L_{dis}$ prevents $M_s$ from diverging too much from $M_t$. 
    \item $M_s$ is initialized with $M_t$ (which was trained on faces with $A=a_{high}$) and $L_{dis}$ prevents $M_s$ from diverging too much from $M_t$. Therefore, \textit{the distillation step allows $M_s$ to process faces with $A=a_{high}$ in the same way $M_t$ would process them, even though $M_s$ is never trained on images from $a_{high}$ category.}
    %\item \textit{This enforces $M_s$ to process faces with $A=a_2$ in the same way $M_t$ would process faces with $A=a_1$.} We consider $M_t$'s representation for a given image to be a reference for $M_s$. $L_{class}$ enforces $M_s$ to classify a given faces with $A=a_2$ using the reference representation. 
    \item $M_t$ is never trained on faces with $A=a_{low}$. So, $L_{class}$ allows $M_s$ to classify faces from $a_{low}$ category. But, \textit{$L_{dis}$ enforces $M_s$ to process faces from $a_{low}$ category in the same way $M_t$ and $M_s$ would process faces from $a_{high}$ category.}
\end{enumerate}
In this way, $M_s$ can process faces from both $a_{high}$ and $a_{low}$ categories as $M_t$ would process faces from $a_{high}$ category. As a result, $M_s$ learns to attend to similar face regions for faces belonging to any category of attribute $A$, as shown in the second rows of all the subfigures of Fig. \ref{fig:avgmap}. Consequently, we find that $M_s$ is able to significantly reduce bias with respect to attribute $A$ (Sec. \ref{subsubsec:biasres}).
%The basic idea here is that we consider $M_t$'s representation for a given image to be a reference for $M_s$. $M_s$ mimics $M_t$ to process an $a_2$ face, and learns to classify it using the reference representation. Note that, $M_s$ is initialized using $M_t$ that was trained on faces with $A=a_1$. Although $M_s$ is trained on faces with $A=a_2$, $L_{dis}$ prevents $M_s$ to forget the information about faces with $A=a_1$. In this way, $M_s$ is able to classify faces from both attribute categories ($a_1$ and $a_2$). More importantly, $M_s$ can process any face belonging to either $a_1$ or $a_2$ in a similar way because $M_s$ learns to classify a face with $A=a_2$ as how $M_t$ would classify a face with $A=a_1$
% feat lev kd
%\begin{figure}
%\centering
%\vspace{-0.3cm}
%{\includegraphics[width=0.65\linewidth]{images/kd_process.pdf}}
%\vspace{-0.9em}
%\caption{\small \textbf{Feature-level knowledge distillation in D\&D.} Teacher $M_t$ is trained on attribute $A=a_1$ (say, female). Student $M_s$ (initialized with $M_t$) is trained on male faces, and is enforced to generate teacher-like features using $L_{dis}$. Thus $M_s$ processes male faces in the same way $M_t$ processes female faces. }\vspace{-0.6cm}
%\label{fig:kdp}
%\end{figure}
%featlevkd
%{\let\thefootnote\relax\footnote{{\textsuperscript{1}Face descriptors refer to the features extracted from the penultimate layer of a previously trained face recognition network.}}} 
For inference, we extract features from the penultimate layer of the trained $M_s$ network for the evaluation dataset, and perform 1:1 face verification. We note that, while $M_s$ considerably reduces bias in face verification, it obtains lower overall face verification performance (Sec. \ref{subsubsec:biasres}). We believe this is because neither $M_s$ nor its teacher $M_t$ is ever trained on the entire dataset. Hence, \textit{we train a new student $M^{*}_s$ (initialized with $M_s$) on the entire dataset to improve the identity classifying ability of $M_s$, while distilling the `unbiasedness' of $M_s$ into $M^{*}_s$}. We call this method D\&D++. (Fig. \ref{fig:dndplus})\\
\textbf{Step 3 (D\&D++)}: Once $M_s$ is trained, we initialize a new student network $M^{*}_s$ using $M_s$. $M^{*}_s$ is trained on the entire dataset with faces from both categories $a_{high}$ and $a_{low}$, to improve its classification performance. During this, we use $M_s$ as the teacher network and distill its `unbiasness' to $M^{*}_s$. Here, we apply the same knowledge distillation used in step 2. We feed the training image to both $M_s$ and $M^{*}_s$ and obtain features $f_s$ and $f^{*}_s$ respectively. $M_s$ remains frozen in this step. We use them to compute $L_{dis}$ as: 
\begin{equation}
    L_{dis}(f_s,f^{*}_s) =1 - \frac{f_s\cdot f^{*}_s}{\|f_s\|\|f^{*}_s\|}.
\end{equation}
Combining $L_{dis}$ with $L_{class}$, we train $M^{*}_s$ with a bias reducing classification loss $L_{br}$ defined as
\begin{equation}
    L_{br} = L_{class} + \lambda_2 L_{dis},
\end{equation}
where $\lambda_2$ is used to weight $L_{dis}$in D\&D++. During inference, we use the trained $M^{*}_s$ to perform verification.\\
\textbf{D\&D and D\&D++ for gender and skintone}: In this work, we show usability of D\&D and D\&D++ to reduce gender and skintone bias (separately). We build two variants of our proposed frameworks: (i) D\&D(g) and D\&D++(g) for reducing gender bias, and (ii)  D\&D(s) and D\&D++(s) for reducing skintone bias, the results for which are presented in Sec. \ref{subsec:afres} and \ref{subsec:cfres}. More training and hyperparameter ($\lambda_1,\lambda_2$) details for D\&D and D\&D++ are provided in the supplementary material. The implementation code will be made publicly available upon publication. 
%\vspace{-0.75cm}
\section{Experiments}
\subsection{Network architectures and datasets}
\label{subsec:archdata}
 \textbf{Training dataset:} We use the BUPT-BalancedFace \cite{wang2020mitigating} dataset for training. For gender bias reduction, we create two subsets of this dataset: Male and female subset. We obtain the gender labels and perform face alignment in this dataset by using \cite{ranjan2017all}. Currently, for skintone bias reduction, there does not exist a large-scale training dataset with skintone labels. So, we use the race labels in BUPT-BalancedFace, as a proxy for skintone. However, the race attribute in this dataset is a non-binary attribute with four categories: African, Asian, Caucasian, Indian. Hence, we binarize this attribute and re-group this dataset into two skintone categories (as explained in Step 0 of Sec. \ref{subsec:dnd}): Light  (`Caucasian' $\cup$ `Asian') and Dark (`African' $\cup$ `Indian'). Although skintone is not perfectly correlated with race, we elect to use these labels due to the high correlation with skintone.\\
 \textbf{Evaluation dataset}: For evaluation, we use aligned faces from IJB-C, and follow the 1:1 face verification protocol from \cite{maze2018iarpa}. The alignment is done using \cite{ranjan2017all}. This dataset provides gender and skintone labels. There are six classes for the skintone attribute which we reorganize into three groups, (i) \textit{Light} (`light pink' $\cup$ `light yellow'), (ii) \textit{Medium} (`medium pink' $\cup$ `medium yellow'), (iii) \textit{Dark} (`medium dark' $\cup$ `dark brown'). For evaluating gender bias, we compute the face verification performance on male-male and female-female pairs separately (out of all the pairs defined in the IJB-C protocol \cite{maze2018iarpa}). For skintone bias, we compute the face verification performance on dark-dark and light-light pairs.\\
 \textbf{Network architecture}: We implement the baselines and our proposed methods (D\&D and D\&D++) using Resnet50 \cite{he2016deep} version of the ArcFace \cite{deng2018ArcFace} network, trained using the Arc-margin loss. To demonstrate the versatility of D\&D and D\&D++, we also perform similar experiments using Crystalface proposed in \cite{ranjan2019fast}, which is a Resnet-101 \cite{he2016deep} network trained using crystal loss.\vspace{-0.25cm}
% \subsection{D\&D and D\&D++ for gender and skintone}
% \label{sec:sequence}
% In Section \ref{subsec:approach}, we present D\&D and D\&D++ as general approaches to reduce bias with respect to any binary attribute.  In D\&D(g), the teacher $M_t$ is trained on the $a_{high}=$male subset of the dataset, and student $M_s$ (initialized with $M_t$) is trained on the $a_{low}$=female images of the dataset. Similarly, in D\&D(s), for $A$ = skintone,  $a_1$ = Light and  $a_2$ = Dark. For Step 3 in both D\&D++(s) and D\&D++(g), we train $M^{*}_s$ on the entire BUPT-BalancedFace dataset, while performing feature-level distillation from $M_s$. 
% \vspace{-0.3cm}
\subsection{Baseline methods}
\label{sec:baseline}
%\vspace{-0.25cm}
We compare our proposed methods with the following de-biasing methods. More training details and hyperparameter information for these baseline methods are provided in the supplementary material.\\
\textbf{PASS: } Protected Attribute Suppression System (PASS) \cite{Dhar_2021_ICCV} is a recently proposed SOTA feature-based adversarial de-biasing framework. Here, features are obtained from a pre-trained network $P$ for the images in the training dataset, and are adversarially made to reduce gender and race information.  The authors present two PASS-based systems: PASS-g (for reducing gender information) and PASS-s (for reducing skintone information) in Crystalface and ArcFace features. {\let\thefootnote\relax\footnote{{\textsuperscript{1}\url{http://umdfaces.io/} Datasets unavailable}}} Both PASS-g and PASS-s are built on top of features from the pre-trained network $P$, which is trained on a combination of UMDFaces\cite{bansal2017umdfaces}, UMDFaces-Videos\cite{bansal2017s} and MS1M \cite{guo2016ms} datasets. \textit{But, UMDFaces\cite{bansal2017umdfaces} and UMDFaces-Videos\cite{bansal2017s} datasets are no longer publicly available\textsuperscript{1}}. So, in our implementation, we train the network $P$ on the publicly available BUPT-BalancedFace \cite{wang2020mitigating} dataset, after which we extract features for this dataset and perform adversarial training to reduce gender (PASS-g) and skintone information (PASS-s) in the extracted features. This also makes PASS systems comparable with our proposed methods: D\&D and D\&D++. We use the official implementation of PASS \cite{passcode} for this task.\\
\textbf{Incremental Variable Elimination}
IVE \cite{terhorst2019suppressing} is an attribute suppression algorithm that excludes variables in the face representation that affect attribute classification. Similar to \cite{Dhar_2021_ICCV},  we use the official implementation of IVE \cite{ivecode} to construct two variants of IVE: IVE(g) and IVE(s) for gender and skintone bias mitigation, using features from ArcFace trained on BUPT-BalancedFace.\\
\textbf{Hair obscuration:} \cite{albiero2020face} shows that obscuring hair in facial images during evaluation reduces gender bias by improving the similarity scores of genuine female-female pairs. We construct a similar pipeline to obscure hair for gender-bias mitigation by using face border keypoints computed by \cite{ranjan2017all} for the images in the evaluation dataset (IJB-C). Following that, we extract features using ArcFace (trained on BUPT-BalancedFace) for all the images in the evaluation dataset and perform 1:1 verification.  \\
\textbf{One step distillation (OSD)} The training sequence in D\&D is $M_t \rightarrow M_s$, and for D\&D++ is $M_t \rightarrow M_s \rightarrow M^{*}_s$, where $M^{*}_s$ is trained on both attribute categories ($a_{high},a_{low}$). We construct a baseline called `One step distillation' (OSD) for which the training sequence is $M_t \rightarrow M^{*}_s$. Here, $M_t$ is trained on one attribute category ($a_{high}$) and $M^{*}_s$ (initialized with $M_t$) is trained on both attribute categories ($a_{high},a_{low}$). $M^{*}_s$ is constrained to mimic $M_t$ through KD used in D\&D and D\&D++. We build two variants of OSD: OSD(g) for reducing gender bias and OSD(s) for reducing skintone bias. In OSD(g) we train $M_t$ on male faces and $M^{*}_s$ on both male and female faces. In OSD(s), we train $M_t$ on light skintone faces and $M^{*}_s$ on both dark and light faces. \vspace{-0.2cm}
%The training sequence in D\&D is $M_t \rightarrow M_s$, where $M_s, M_t$ are trained on different attribute categories. The training sequence in D\&D++ is $M_t \rightarrow M_s \rightarrow M^{*}_s$, where $M^{*}_s$ is trained on both attribute categories ($a_1,a_2$).
\begin{figure*}
{
\centering
\subfloat[AF/Gender]{\includegraphics[width=0.25\linewidth]{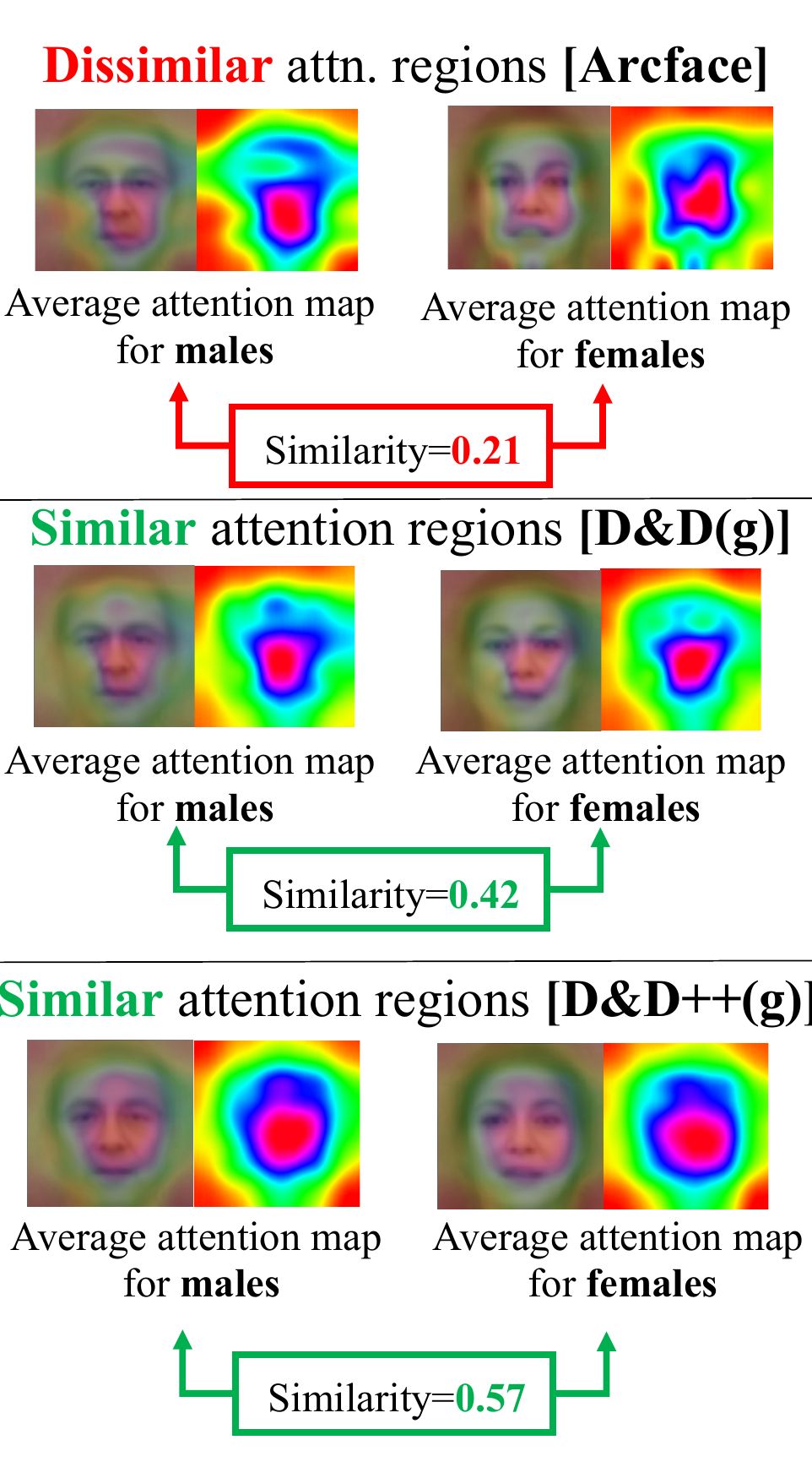}\label{fig:arcamgender}}~
%\rulesep
\subfloat[AF/Skintone]{\includegraphics[width=0.25\linewidth]{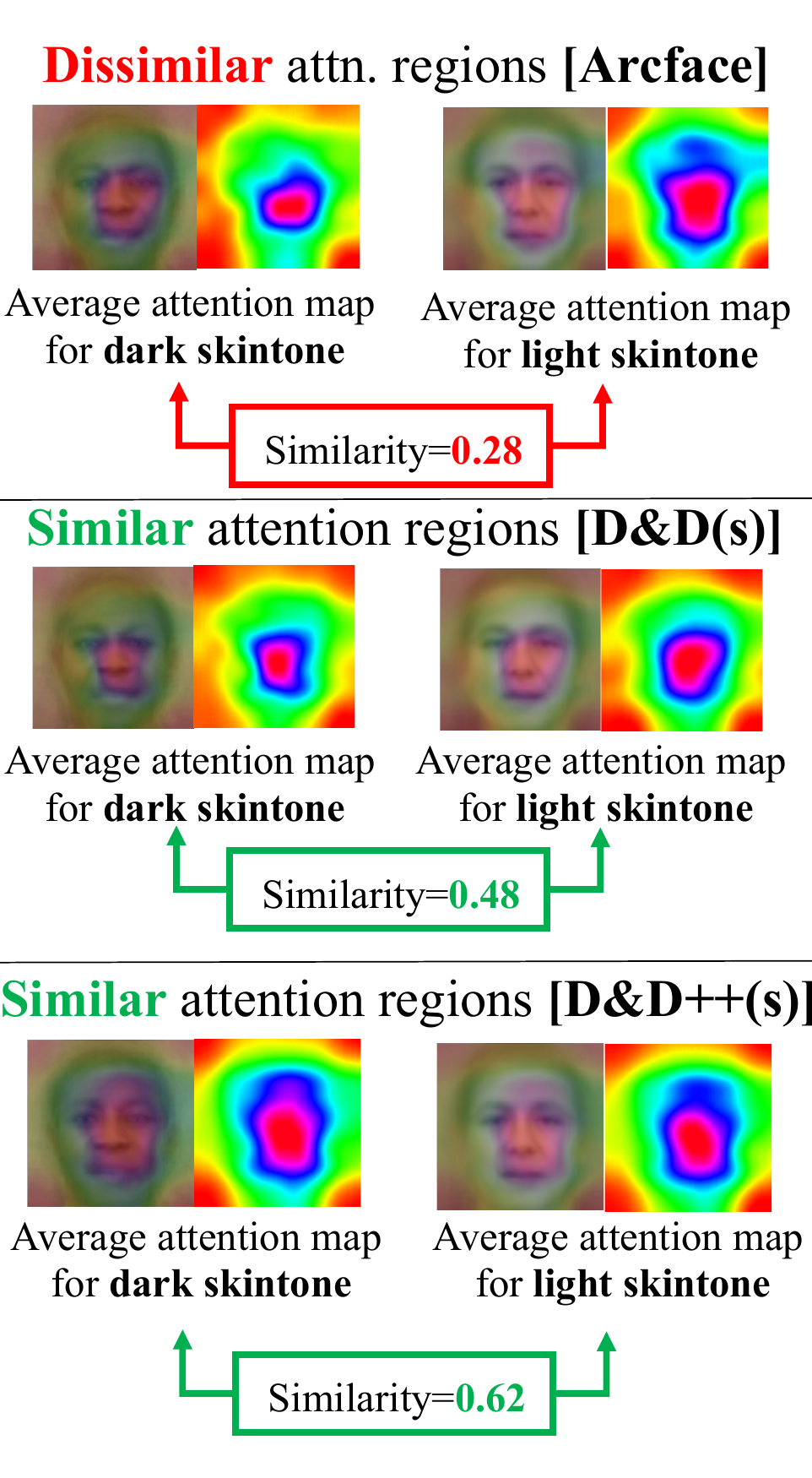}\label{fig:arcamst}}
%\rulesep
\subfloat[CF/Gender]{\includegraphics[width=0.25\linewidth]{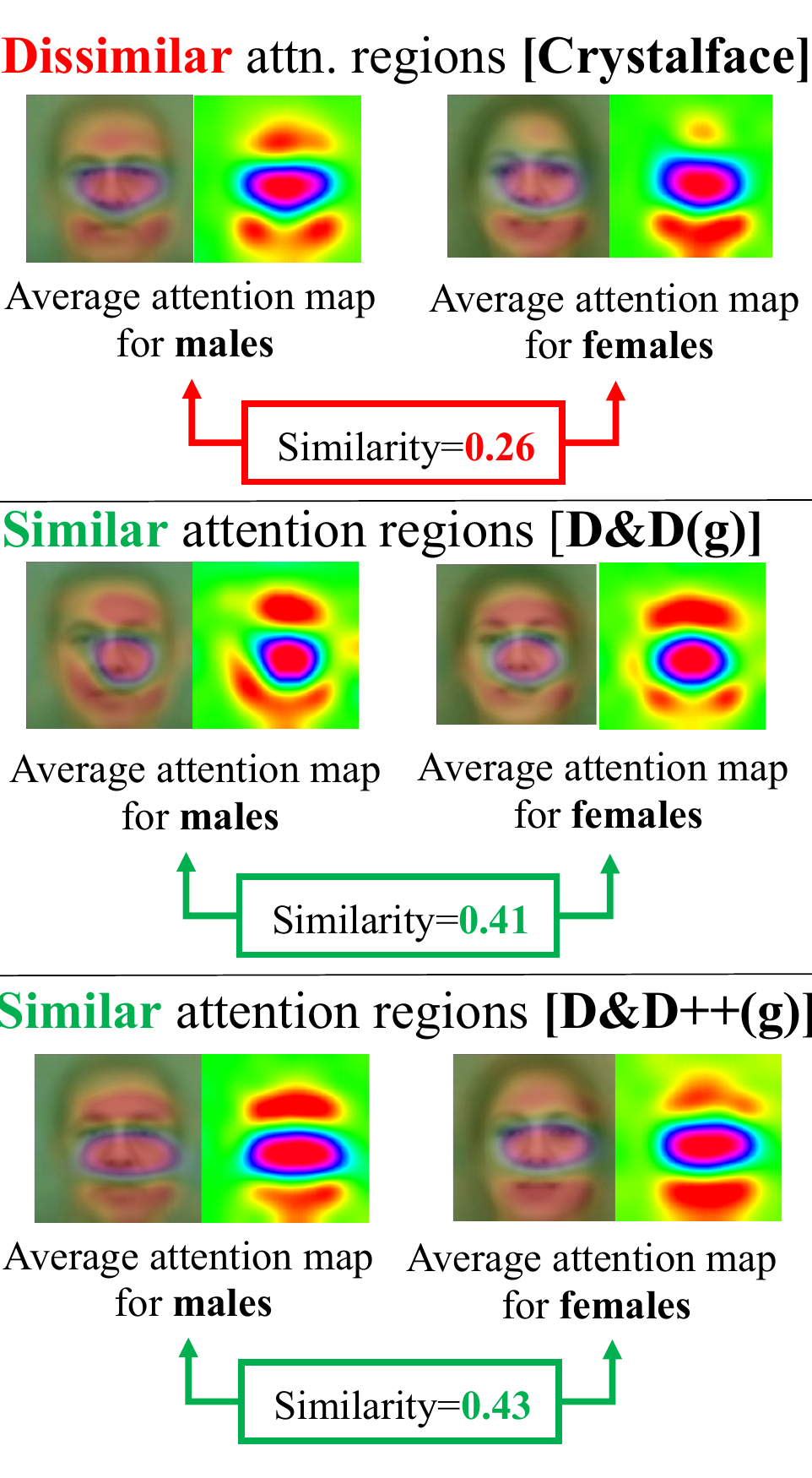}\label{fig:amgender}}
%\rulesep
\subfloat[CF/Skintone]{\includegraphics[width=0.25\linewidth]{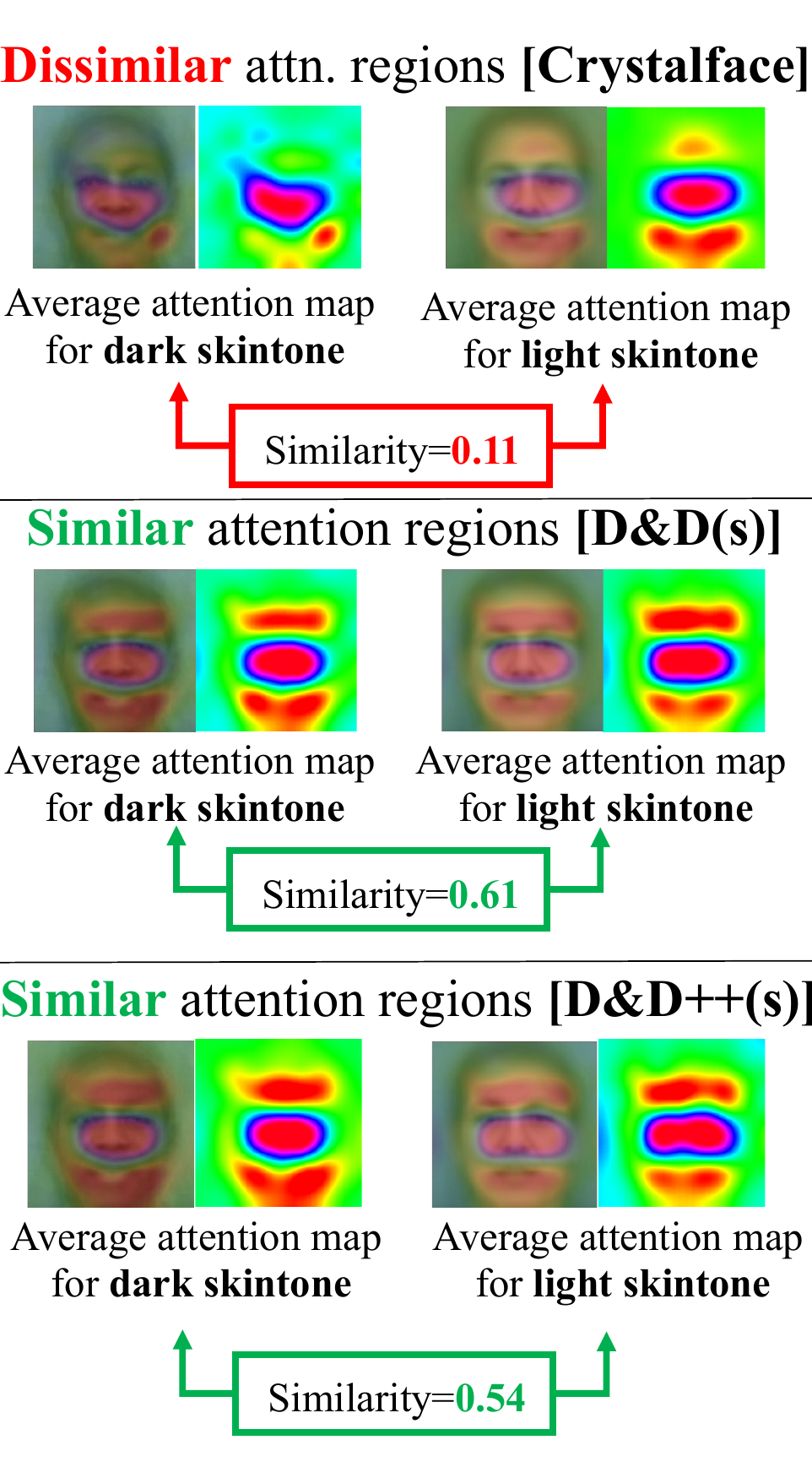}\label{fig:amst}}~
\vspace{-1em}
\caption{\small \textbf{(a,c)} D\&D(g) and D\&D++(g) generate more similar attention maps for male and female frontal faces, as compared to the original \textbf{(a)} ArcFace (AF), \textbf{(c)} Crystalface (CF) network. \textbf{(b,d)} D\&D(s) and D\&D++(s) generate more similar attention maps for dark and light frontal faces, than he original \textbf{(b)} ArcFace, \textbf{(d)} Crystalface network.\vspace{-0.2cm}}
\label{fig:avgmap}
}
\end{figure*}
\subsection{Results with ArcFace}
\label{subsec:afres}
\subsubsection{Effect on attention regions}
\label{sec:ameffect}
We first select frontal facial images in IJB-C dataset for which the yaw angles (computed using \cite{ranjan2017all}) lie between -5 to 5 degrees. We use GradCAM \cite{selvaraju2017grad} to generate the attention maps for all of these images using the last convolutional layer of the ArcFace network. These attention maps highlight important face regions relevant for the face recognition decision. After that we separate out the attention maps for males and females, and average them separately. Figure \ref{fig:arcamgender} (first row) shows the average attention maps for males and females, generated using ArcFace.\\
\textbf{Similarity between male and female attention maps}: We compute the cosine similarity between the flattened average attention maps for males and females, which turns out to be low (0.21). This implies that male and female faces are processed differently by ArcFace. This might lead to gender bias, when features from ArcFace are used in face verification. We then generate the average attention maps for males and females using ArcFace trained with D\&D(g) and D\&D++(g). From Fig. \ref{fig:arcamgender} (second and third row), it is clear that the male and female average attention maps are more similar when we use D\&D(g) and D\&D++(g), compared to ArcFace. This indicates that both male and female faces are processed in a more similar way with our distillation-based approaches.\\ 
\textbf{Similarity between light and dark attention maps:} We repeat this experiment to generate the average attention maps for light and dark frontal faces in IJB-C using ArcFace and its D\&D(s) and D\&D++(s) counterparts. Again, we find that the attention regions for dark and light faces generated using the original ArcFace network are dissimilar (cosine similarity=0.28), that might lead skintone bias. However, the attention maps for light and dark skintones generated using D\&D(s) and D\&D++(s) are much more similar (Fig. \ref{fig:arcamst})\vspace{-0.4cm}. 
\begin{table*}[]
\centering
%\scriptsize
\subfloat[\small Gender bias - ArcFace backbone (TPR\textsubscript{m}: male-male TPR, TPR\textsubscript{f}: female-female TPR)]{\scalebox{0.73}{
\hspace{-0.48cm}\begin{tabular}{c|ccccc|ccccc|ccccc}
\toprule
 FPR &  & {  } & {  $10^{-5}$} & {  } & {  } & {  } & {  } & {  $10^{-4}$} & {  } & {  } & {  } & {  } & {  $10^{-3}$} & {  } & {  } \\
 \midrule
  Method & TPR & TPR\textsubscript{m} & TPR\textsubscript{f} & Bias$(\downarrow)$ \hspace{-4pt} & \hspace{-4pt} BPC\textsubscript{g}$(\uparrow)$ \hspace{-4pt} & TPR & TPR\textsubscript{m} & TPR\textsubscript{f} & Bias$(\downarrow)$ \hspace{-4pt} & \hspace{-4pt} BPC\textsubscript{g}$(\uparrow)$ \hspace{-4pt} & TPR & TPR\textsubscript{m} & TPR\textsubscript{f} & Bias$(\downarrow)$ \hspace{-4pt} & \hspace{-4pt} BPC\textsubscript{g}$(\uparrow)$ \hspace{-4pt} \\
  \midrule
ArcFace & 0.879 & 0.884 & 0.841 & 0.042 & 0.00 & 0.914 & 0.922 & 0.890 &  0.032 & 0.00 & 0.944 & 0.946 & 0.928 & 0.017 & 0 \\
IVE(g)$\dag$\cite{terhorst2019suppressing} & 0.877 & 0.884 & 0.843 & 0.041 & 0.021 & 0.913 & 0.920 & 0.886 & 0.034 & -0.064 & 0.944 & 0.944 & 0.927 & 0.017&0 \\
W/o hair$\dag$\cite{albiero2020face}  & 0.726 & 0.412 &0.821 &0.409  & -8.91 & 0.883 &0.794  &0.888  & 0.094 & -1.97 & 0.926 &0.930  & 0.926 & \underline{0.004} &\underline{0.746} \\
PASS-g$\dag$\cite{Dhar_2021_ICCV} &0.798&0.681&0.768&0.087&-1.162&0.869&0.851&0.862&0.011&\underline{0.607}&0.909&0.916&0.902&0.013&0.198\\
\midrule
OSD(g) &  
0.778&
0.758&
0.780&
0.022&
0.361&
0.848&
0.849&
0.865&
0.017&
0.397&
0.898&
0.901&
0.917&
0.016&
0.010\\
\rowcolor{Gray}
D\&D(g)  & 
0.759&
0.754&
0.769&
\underline{0.016}&
\underline{0.483}&
0.830&
0.833&
0.843&
\underline{0.010}&
0.596&
0.889&
0.889&
0.897&
0.009&
0.412\\
\rowcolor{Gray}
D\&D++(g)  & 
0.825&
0.803&
0.800&
\textbf{0.002}&
\textbf{0.891}&
0.880&
0.879&
0.870&
\textbf{0.009}&
\textbf{0.682}&
0.920&
0.920&
0.918&
\textbf{0.002}&
\textbf{0.857}\\
\bottomrule
\end{tabular}
}}\\
\vspace{-0.4cm}
\subfloat[Skintone bias - ArcFace backbone (TPR\textsubscript{l}: light-light TPR, TPR\textsubscript{d}: dark-dark TPR)]{\scalebox{0.73}{
\hspace{-0.48cm}\begin{tabular}{c|ccccc|ccccc|ccccc}
\toprule
 FPR &  & {  } & {  $10^{-4}$} & {  } & {  } & {  } & {  } & {  $10^{-3}$} & {  } & {  } & {  } & {  } & {  $10^{-2}$} & {  } & {  } \\
 \midrule
  Method & TPR & TPR\textsubscript{l} & TPR\textsubscript{d} & Bias$(\downarrow)$ \hspace{-4pt} & \hspace{-4pt} BPC\textsubscript{st}$(\uparrow)$ \hspace{-4pt} & TPR & TPR\textsubscript{l} & TPR\textsubscript{d} & Bias$(\downarrow)$ \hspace{-4pt} & \hspace{-4pt} BPC\textsubscript{st}$(\uparrow)$ \hspace{-4pt} & TPR & TPR\textsubscript{l} & TPR\textsubscript{d} & Bias$(\downarrow)$ \hspace{-4pt} & \hspace{-4pt} BPC\textsubscript{st}$(\uparrow)$ \hspace{-4pt} \\
  \midrule
ArcFace & 0.914 & 0.912 & 0.883 & 0.029 & 0 & 0.944 &0.942 & 0.922
 & 0.021 & 0 & 0.964 & 0.964 & 0.950 & 0.014 & 0 \\
IVE(s)$\dag$\cite{terhorst2019suppressing} &0.913& 0.911 & 0.871 & 0.040 & -0.380 & 0.943 & 0.941 & 0.919 & 0.022
 & -0.049 & 0.964 & 0.962 & 0.951 & 0.011 &  0.214\\
PASS-s$\dag$\cite{Dhar_2021_ICCV} & 0.786& 0.778 & 0.738 & 0.041 &-0.554 & 0.861 & 0.859 & 0.846 & 0.014 & 0.245 & 0.920 &0.921 &0.922  & \textbf{0.001} & \textbf{0.883} \\
 \midrule
OSD(s) & 0.877 &0.864  & 0.859 & \underline{0.005} & \underline{0.787} & 0.923 & 0.918 & 0.901 & 0.016 & 0.216 & 0.956 & 0.953 & 0.944 & 0.009 & 0.349 \\
\rowcolor{Gray}
D\&D(s) & 0.855 & 0.836 & 0.851 & 0.015 & 0.418 & 0.913 & 0.906 & 0.895 & \textbf{0.011} & \underline{0.443} & 0.951 & 0.947 & 0.942 & 0.005 & 0.629\\
\rowcolor{Gray}
D\&D++(s) & 0.882 & 0.871 & 0.868 & \textbf{0.003} & \textbf{0.862} & 0.926 & 0.923 & 0.912 & \textbf{0.011} &  \textbf{0.457}& 0.957 & 0.954 & 0.951 & \underline{0.003} & \underline{0.778}\\
\bottomrule
\end{tabular}
}}
\caption{\small Bias analysis for \textit{ArcFace} network, and its de-biased counterparts on IJB-C. TPR: overall True Positive rate. \textbf{Bold}=Best, \underline{Underlined}=Second best. D\&D variants obtain highest BPC and lowest bias at most FPRs. \textsuperscript{$\dag$}=Our implementation of baselines (See \ref{sec:baseline} for details). All methods are trained on BUPT-BalancedFace \cite{wang2020mitigating} data.\vspace{-0.2cm}} \label{tab:afallbias}
\vspace{-0.2cm}
\end{table*}
\begin{figure*}
{
\centering
\subfloat[Gender bias(AF)]{\includegraphics[width=0.25\linewidth]{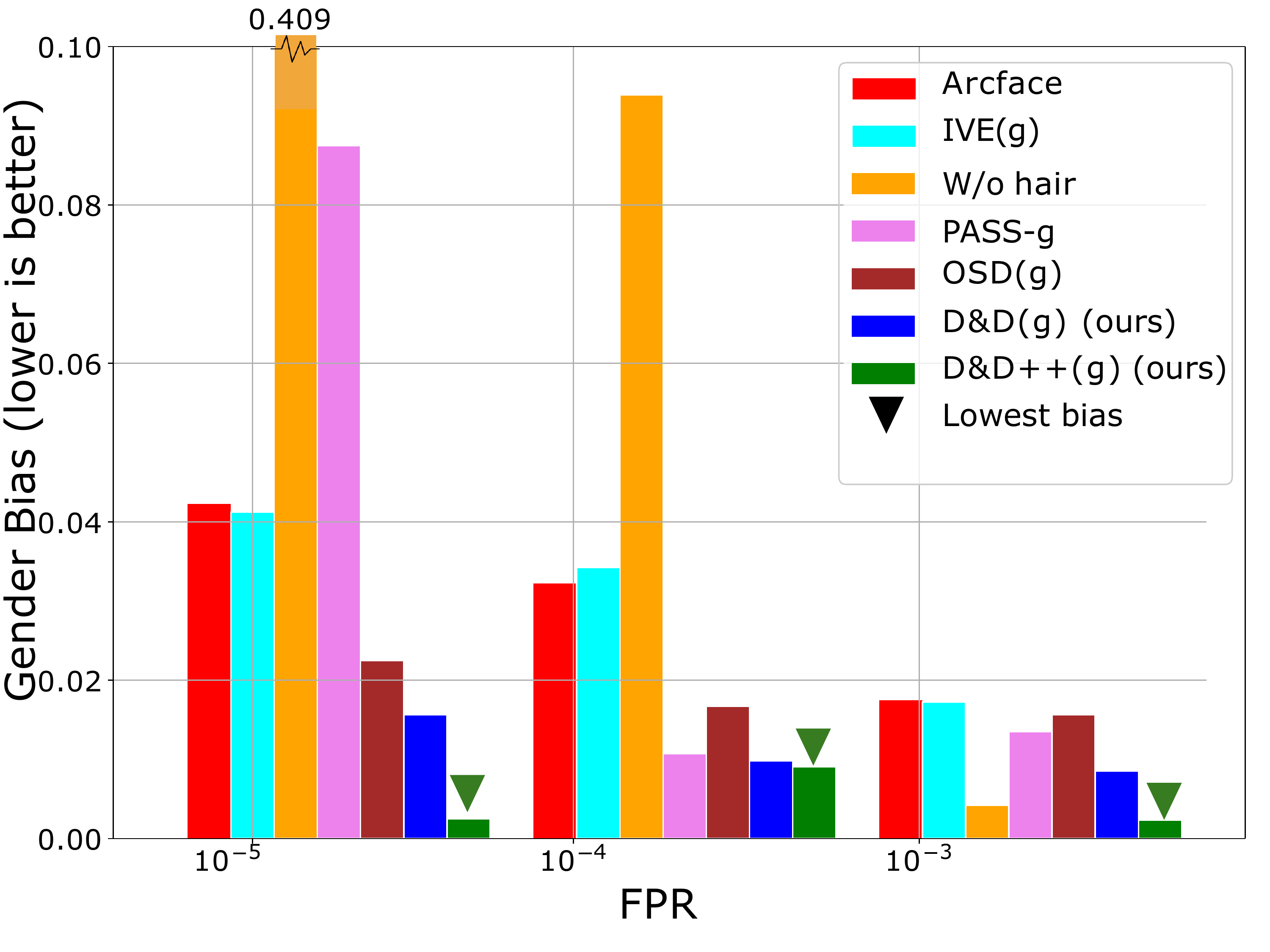}\label{fig:afgenplot}}
%\subfloat[]{\includegraphics[width=0.25\linewidth]{images/aaai_skintone_ArcFace_overall_dnd_bupt_skintone_latest.pdf}}
\subfloat[ST bias(AF)]{\includegraphics[width=0.25\linewidth]{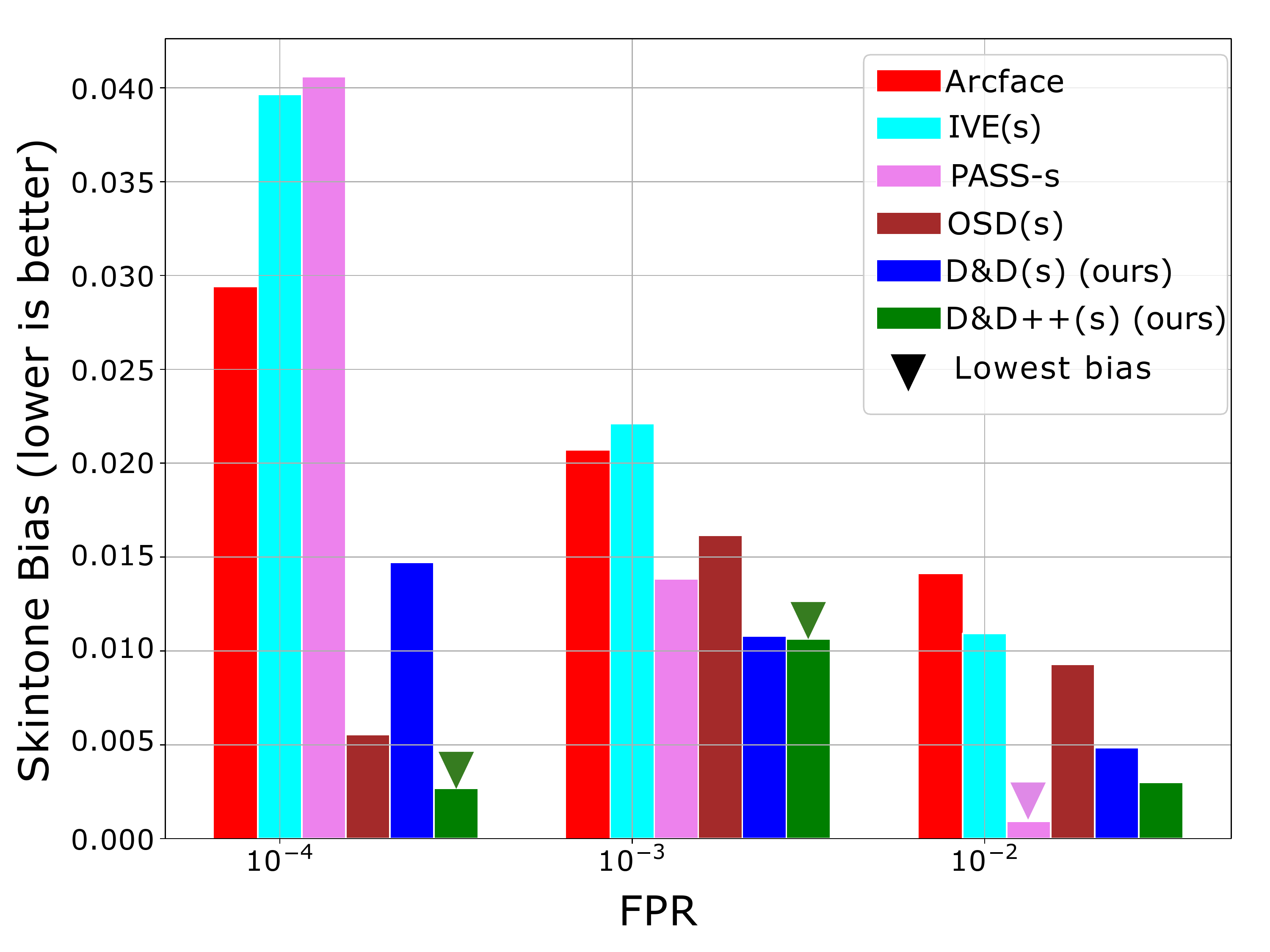}\label{fig:afstplot}}
\subfloat[Gender bias(CF)]{\includegraphics[width=0.25\linewidth]{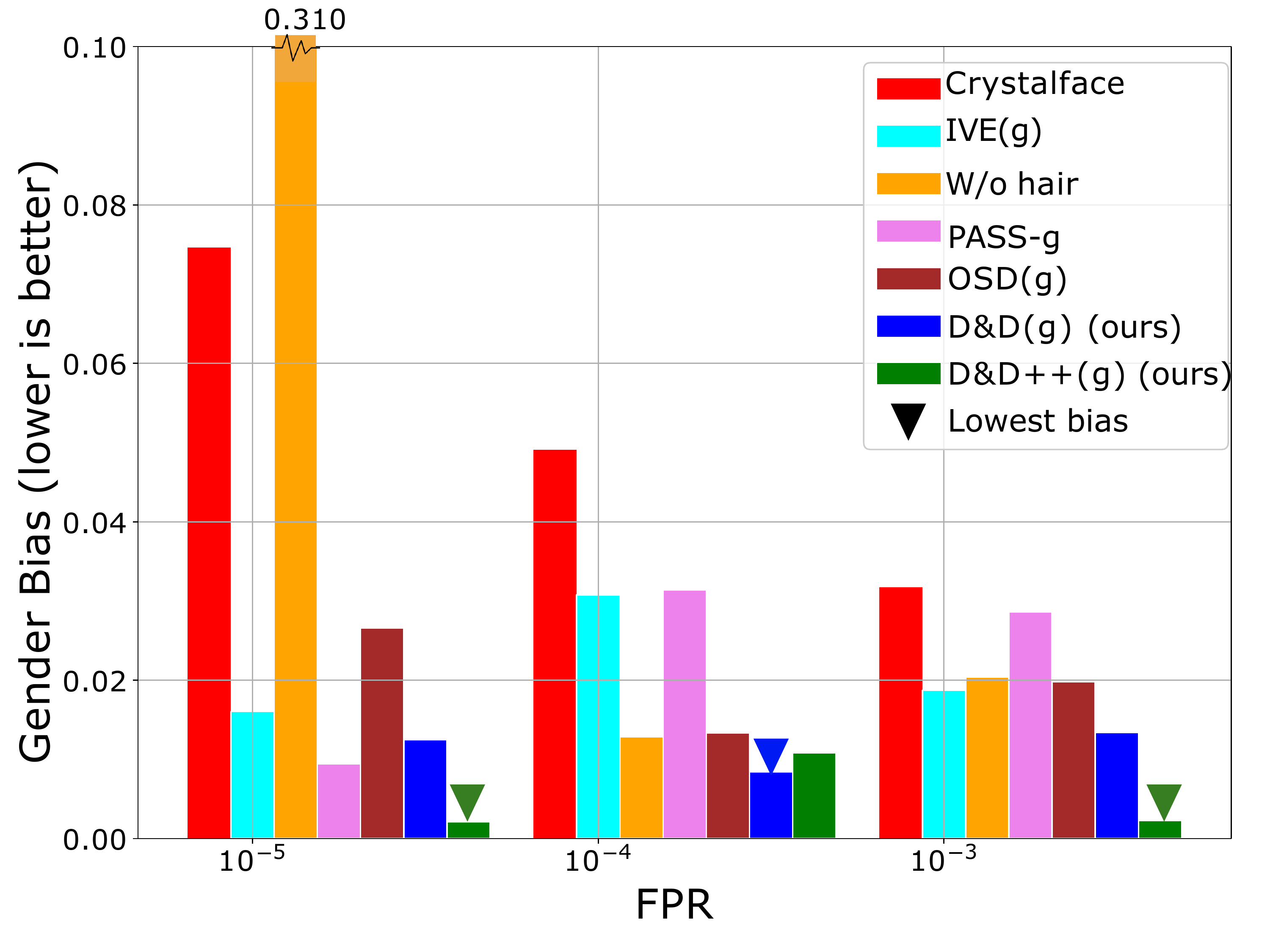}\label{fig:cfgenplot}}
\subfloat[ST bias(CF)]{\includegraphics[width=0.25\linewidth]{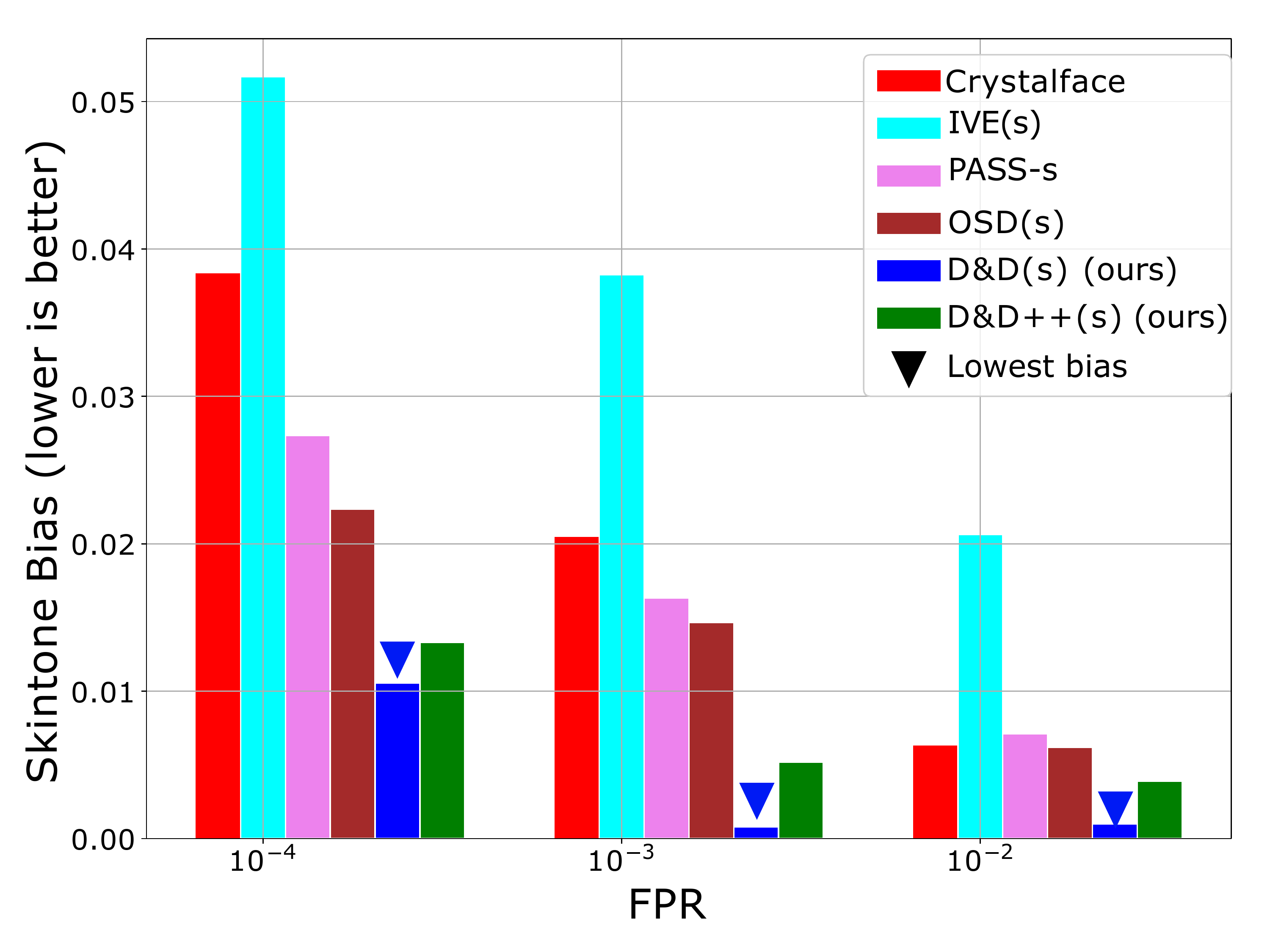}\label{fig:cfstplot}}
\vspace{-1em}
% \caption{\small Crystalface-based D\&D/D\&D++ obtain lowest (a)Gender and (b)Skintone bias in IJB-C than other baselines using Crystalface backbone at all FPRs.}\vspace{-0.75cm}
\caption{\small D\&D or D\&D++ obtain the lowest gender and skintone (ST) bias for \textbf{(a,b)} ArcFace (AF) and \textbf{(c,d)} Crystalface (CF) in IJB-C at most FPRs. \textit{Please see Sec. \ref{sec:baseline} for our implementation of baselines.} Best viewed when zoomed in.}\vspace{-0.7cm}
\label{fig:cfbiasplot}
}
\end{figure*}
\subsubsection{Evaluating gender and skintone bias}
\label{subsubsec:biasres}
We now evaluate the effectiveness of D\&D and D\&D++ to reduce bias in terms of a given attribute. From Table \ref{tab:afallbias} and Figs. \ref{fig:afgenplot},\ref{fig:afstplot}, we infer that ArcFace networks trained with D\&D/D\&D++ obtain the lowest gender/skintone bias at most FPRs. This confirms our hypothesis (in Sec. \ref{sec:probstatement}), that networks that process faces belonging to different attribute categories in a similar way demonstrate lower attribute-bias. Also, from Table \ref{tab:afallbias}, we also infer that \textit{D\&D-based frameworks obtain higher BPCs} (Eq. \ref{eq:bpc}) \textit{than the baselines at all FPRs}. Moreover, it is clear that \textit{D\&D++ obtains better face verification performance than D\&D and PASS \cite{Dhar_2021_ICCV}}, while maintaining low gender and skintone bias. This demonstrates the advantage of the additional step of training $M^{*}_s$ on the full dataset, which adds specificity and transfers the `unbiasedness' from $M_s$ to $M^{*}_s$ through distillation. Since most real-time face recognition systems are evaluated at low FPRs \cite{nistres}, it is important to reduce bias especially at low FPRs while maintaining high verification performance. D\&D++ clearly achieves this target. % Note that, similar to \cite{Dhar_2021_ICCV}, we report the bias and TPRs till FPR=$10^{-5}$ when evaluating gender bias whereas for skintone bias, we only report till FPR=$10^{-4}$. This is because the reported FPR is a function of the number of pairs, and IJBC has fewer dark subjects which does not provide enough pairs for statistically significant measurements at FPR=$10^{-5}$. We provide more qualitative results in Figs. \ref{fig:afqualresgender}, \ref{fig:afqualresst} to show that D\&D++ helps the network attend to similar spatial regions for both categories of the binary attribute under consideration. We provide the gender-wise,  skintone-wise ROC plots (similar to the ROC curves in Fig. \ref{fig:teaserdnd}) in the supplementary material. \vspace{-0.5cm}

Note that, similar to \cite{Dhar_2021_ICCV}, we report the bias and TPR down to FPR=$10^{-5}$ when evaluating gender bias, whereas for skintone bias we only report down to FPR=$10^{-4}$. The reason for this is the relative lack of dark-skintone examples in IJBC, which results fewer dark-dark pairs. The ROC curve begins to become quantized as the FPR nears $\frac{1}{\text{\# negatives}}$, resulting in less statistical significance. Since the FPR values become less reliable, so do the TPR measurements associated with them.

\begin{figure*}
\centering
\vspace{-0.3cm}
\subfloat[Male-female att\textsuperscript{n} map with AF backbone]{
{\includegraphics[width=0.5\linewidth]{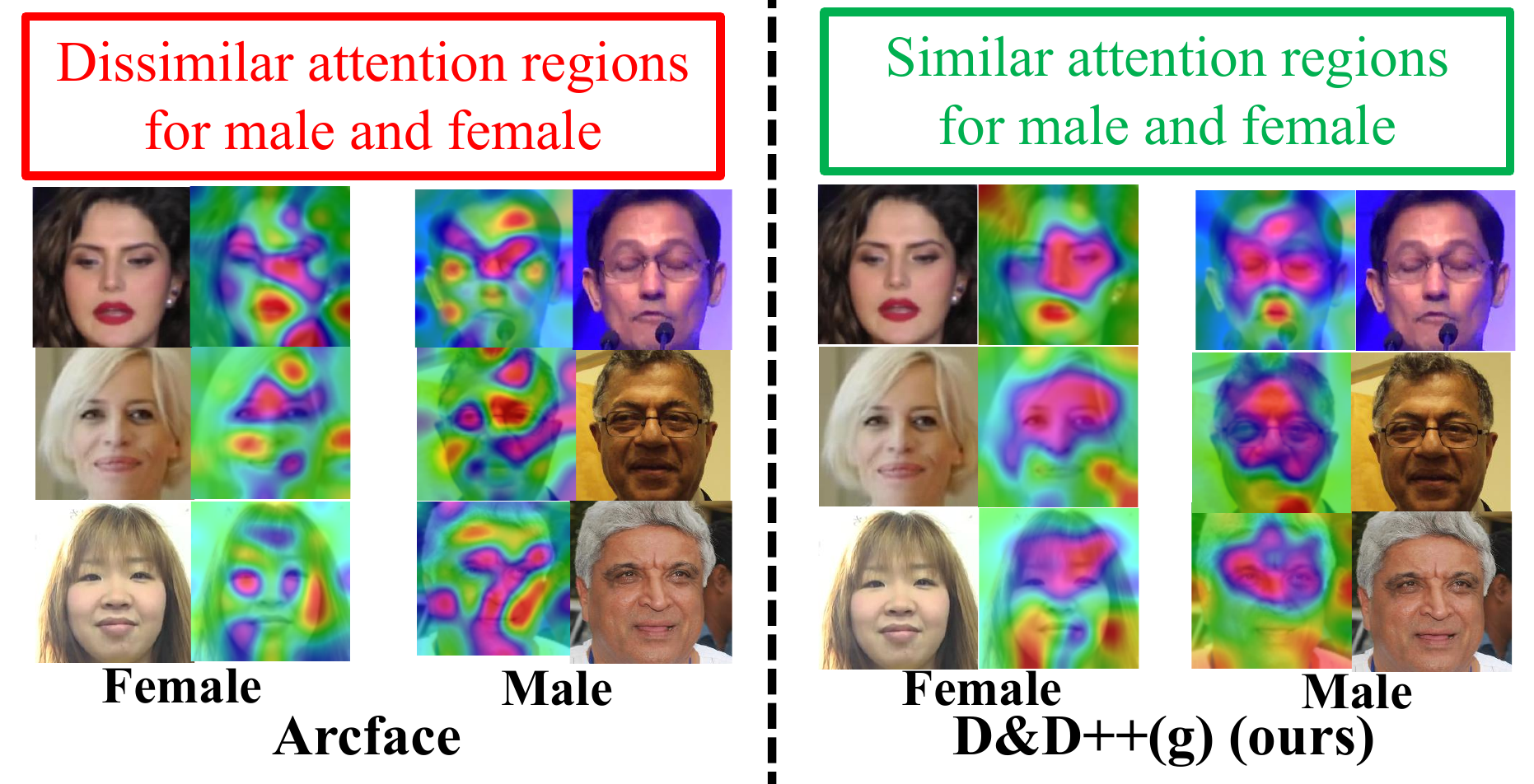}}\label{fig:afqualresgender}}~
\subfloat[Dark-light att\textsuperscript{n} map with AF backbone]{
{\includegraphics[width=0.5\linewidth]{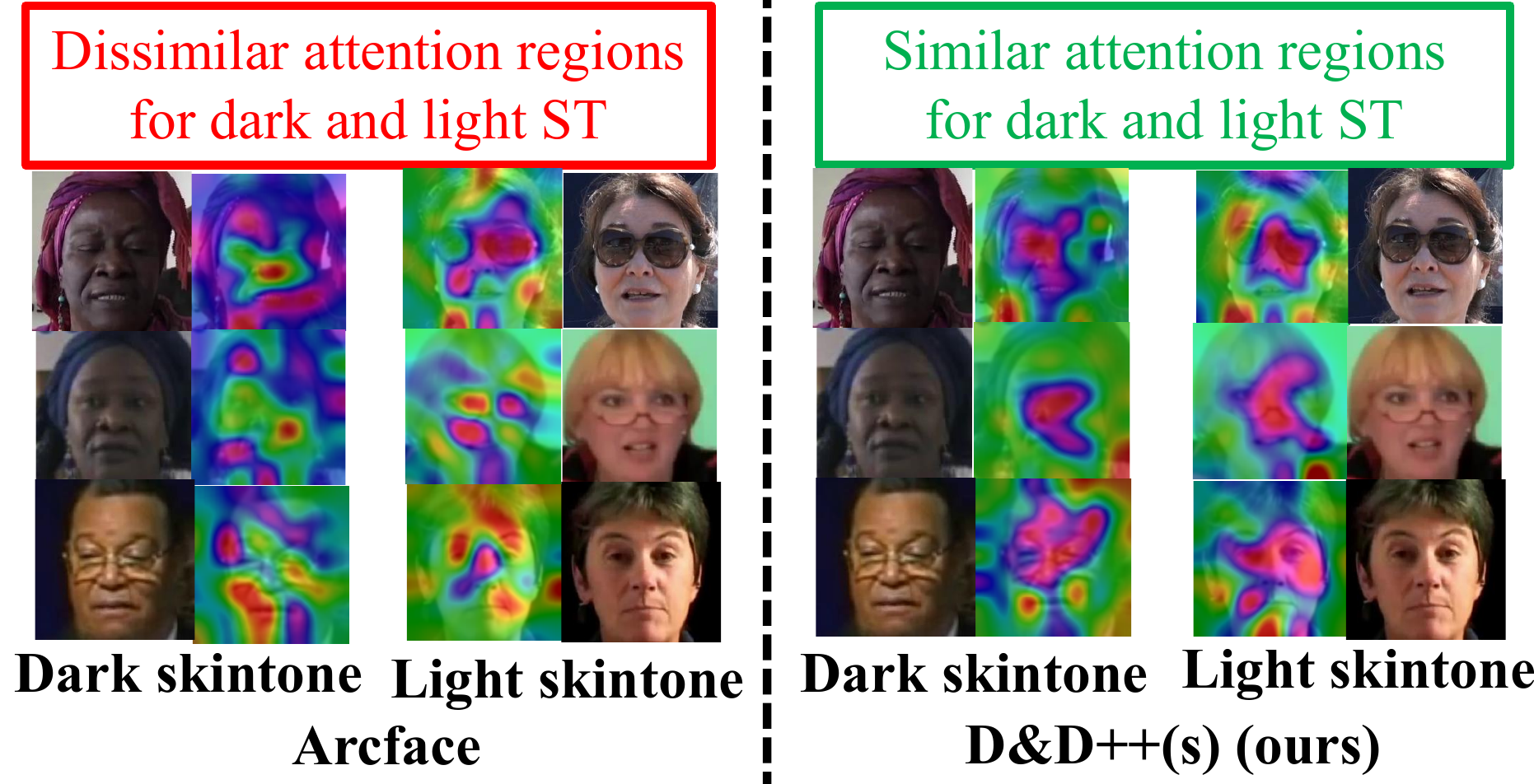}}\label{fig:afqualresst}}\\
\vspace{-0.4cm}
\subfloat[Male-female att\textsuperscript{n} map with CF backbone]{
{\includegraphics[width=0.5\linewidth]{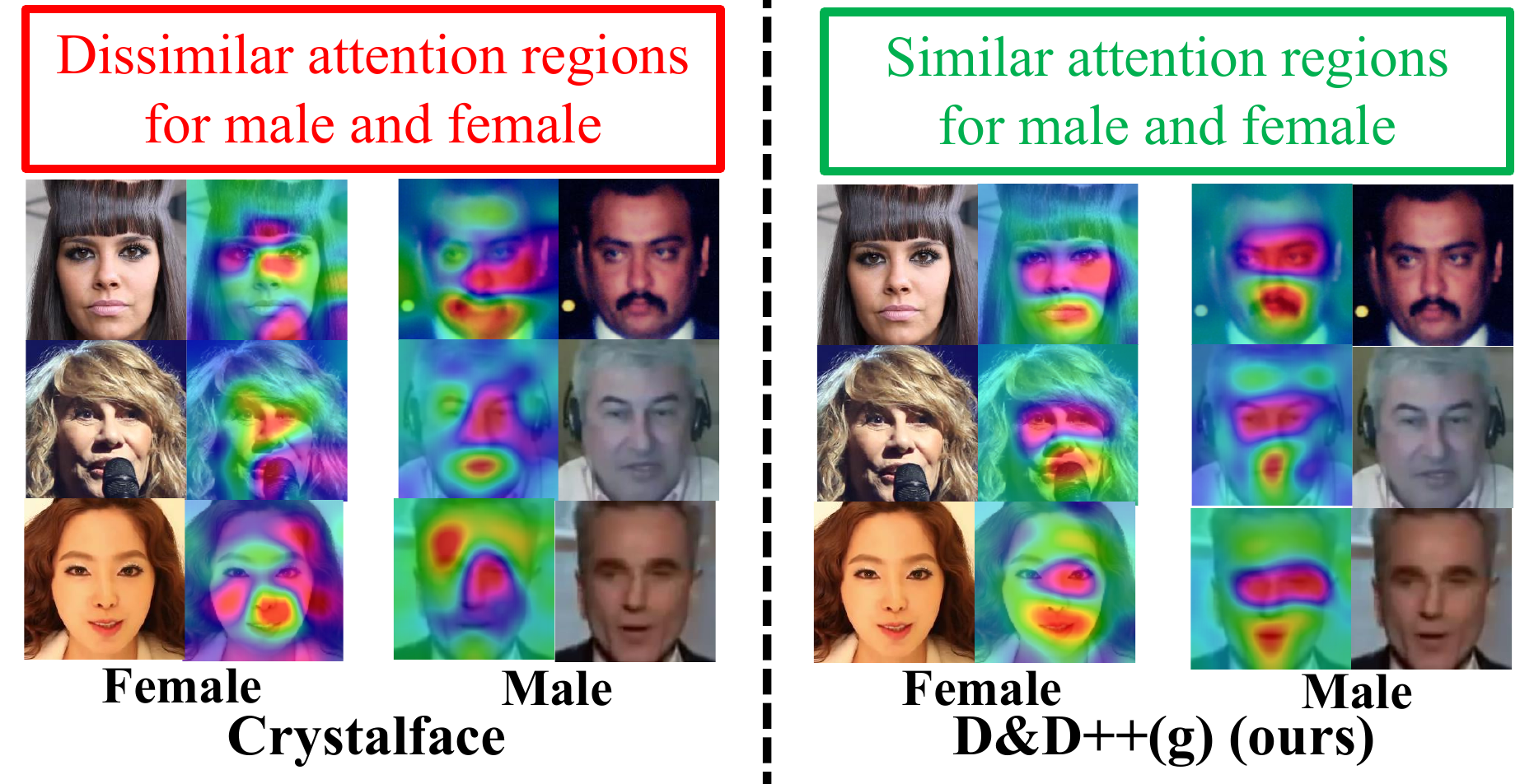}}\label{fig:cfqualresgender}}~
\subfloat[Dark-light att\textsuperscript{n} map with CF backbone]{
{\includegraphics[width=0.5\linewidth]{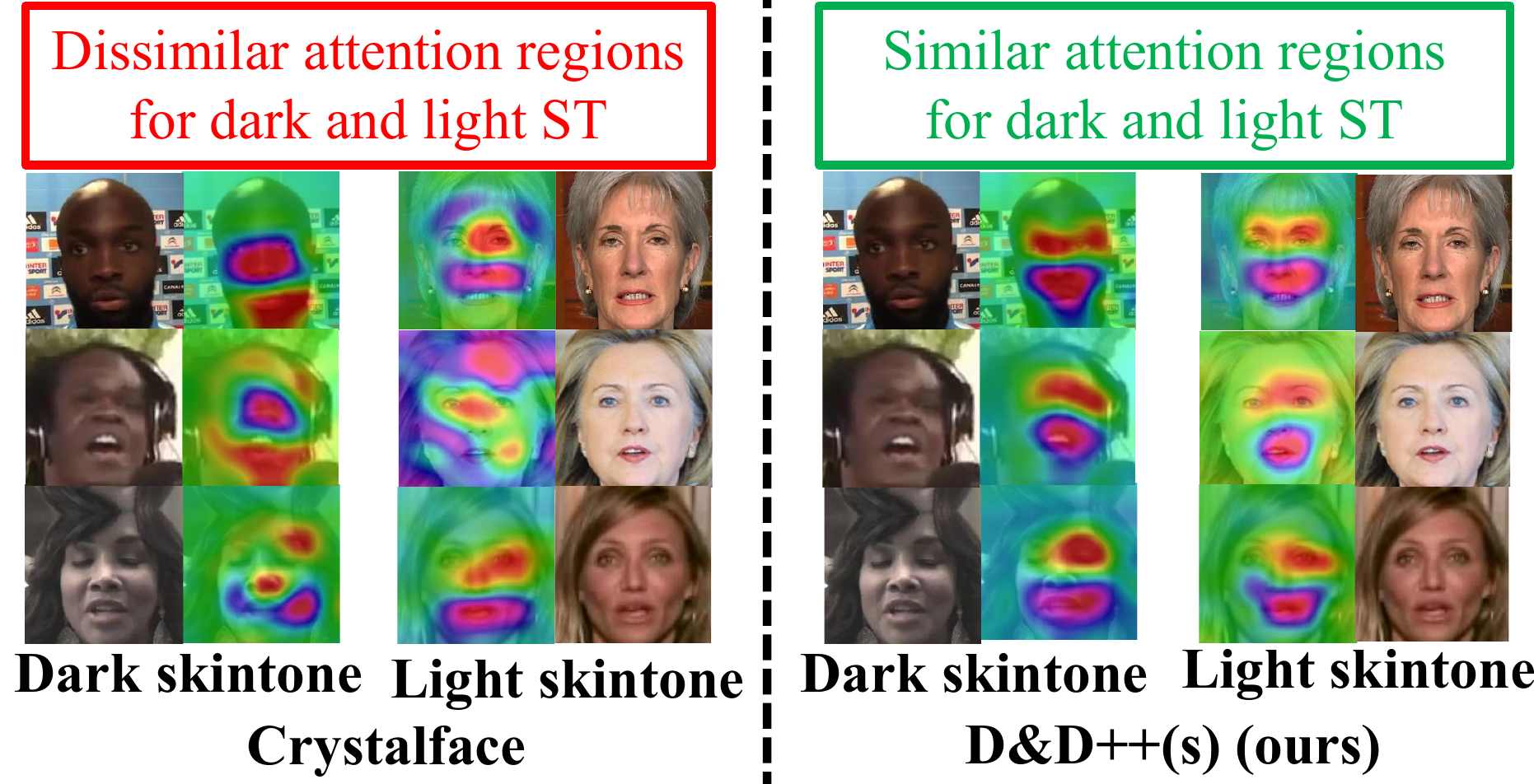}}\label{fig:cfqualresst}}
\vspace{-0.2cm}
\caption{\small \textbf{(a,c)} D\&D++(g) attends to similar face regions for males and females. \textbf{(b,d)} D\&D++(s) attends to similar face regions for faces with dark and light skintones. The attention maps are generated using GradCAM \cite{selvaraju2017grad}.AF=ArcFace, CF=Crystalface. }
\label{fig:qualres}
\end{figure*}

We provide more qualitative results in Figs. \ref{fig:afqualresgender}, \ref{fig:afqualresst} to show that D\&D++ helps the network attend to similar spatial regions for both categories of the binary attribute under consideration. We provide the gender-wise,  skintone-wise ROC plots (similar to the ROC curves in Fig. \ref{fig:teaserdnd}) in the supplementary material. \vspace{-0.5cm}

\subsubsection{Evaluating D\&D/D\&D++ on non-binary attributes}
\label{subsubsec:third}
\begin{figure*}
{
\centering
\subfloat[ArcFace]{\includegraphics[width=0.33\linewidth]{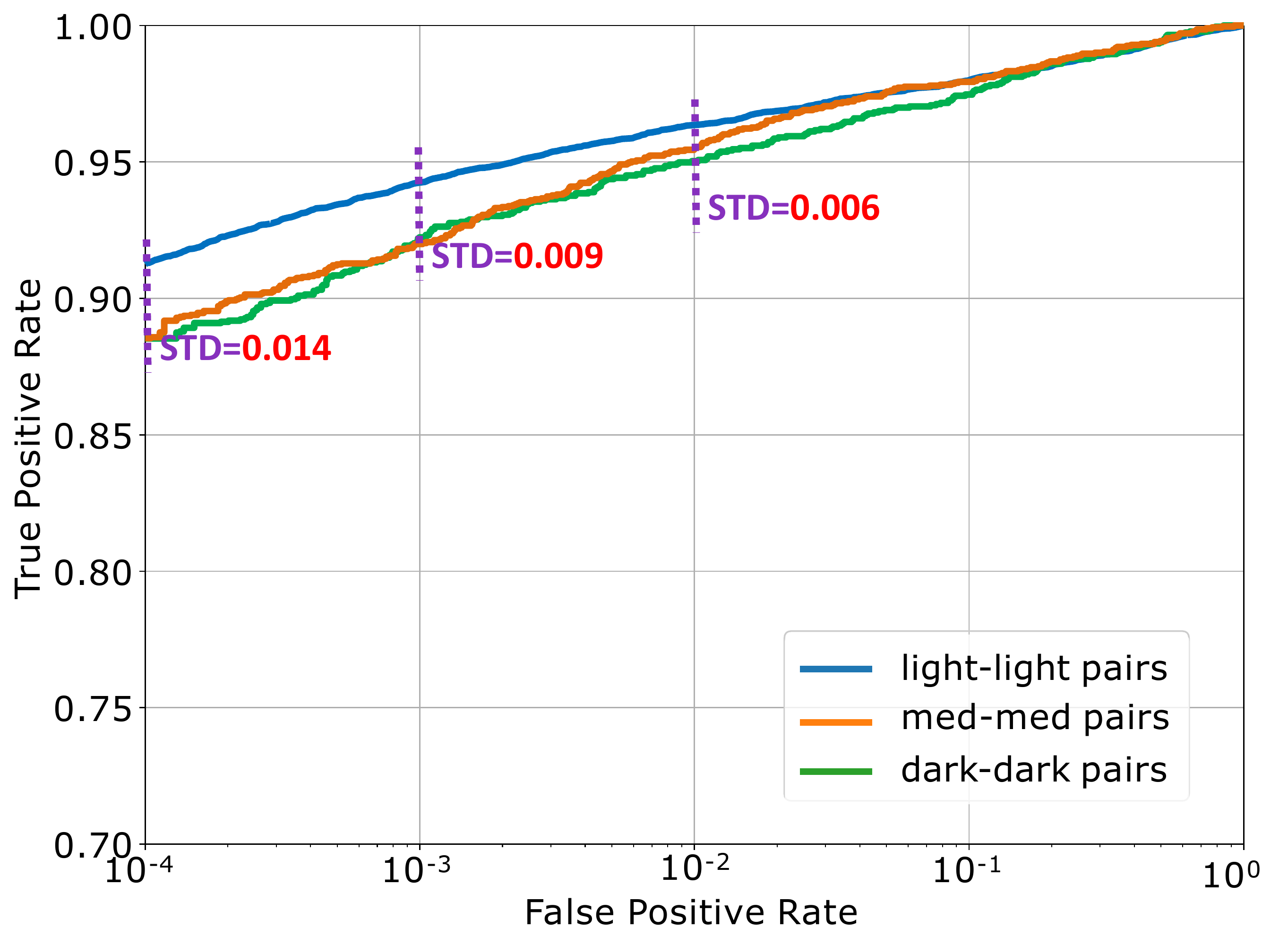}}
\subfloat[IVE(s)\cite{terhorst2019suppressing}]{\includegraphics[width=0.33\linewidth]{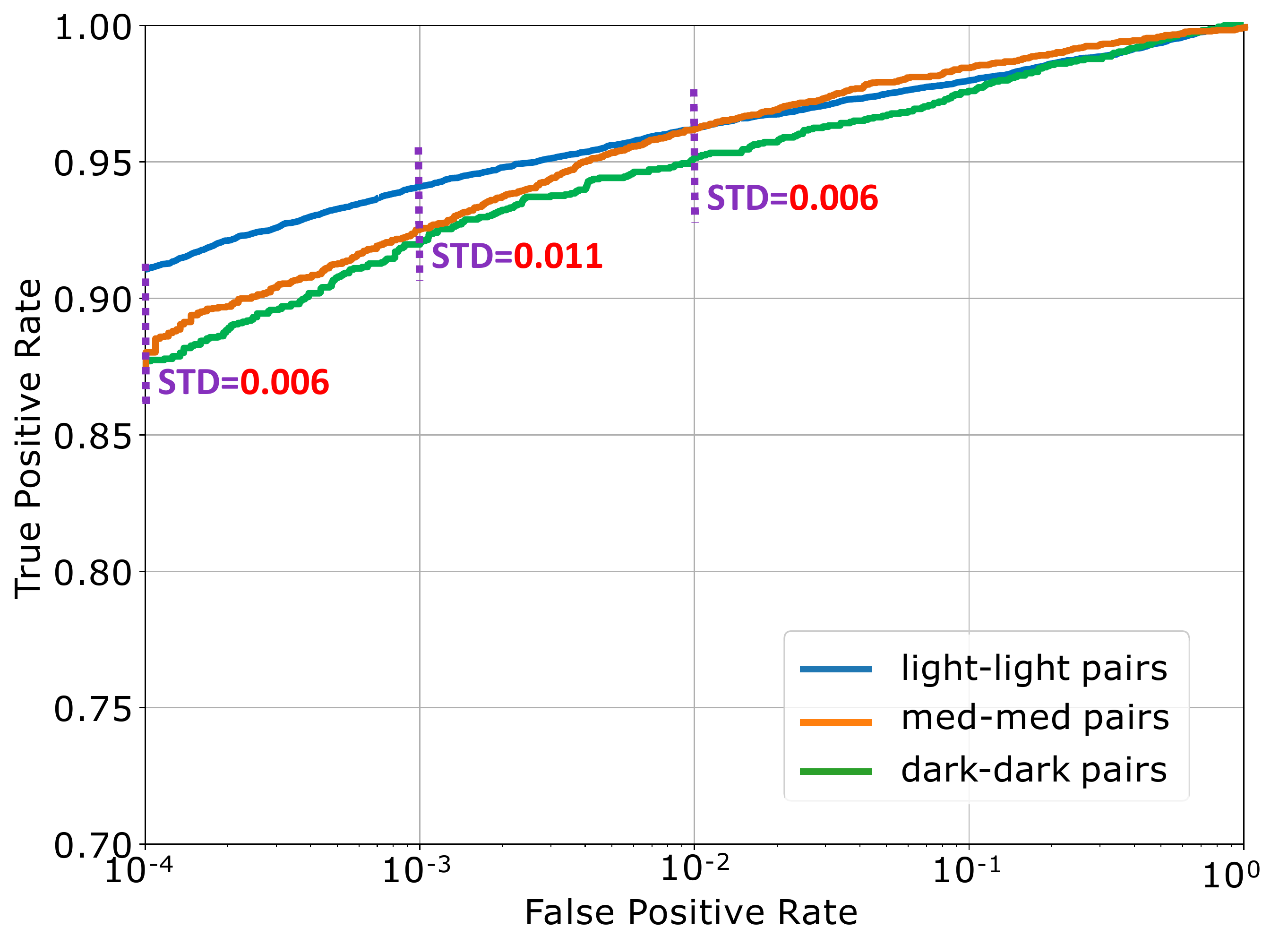}}
\subfloat[PASS-s\cite{Dhar_2021_ICCV}]{\includegraphics[width=0.33\linewidth]{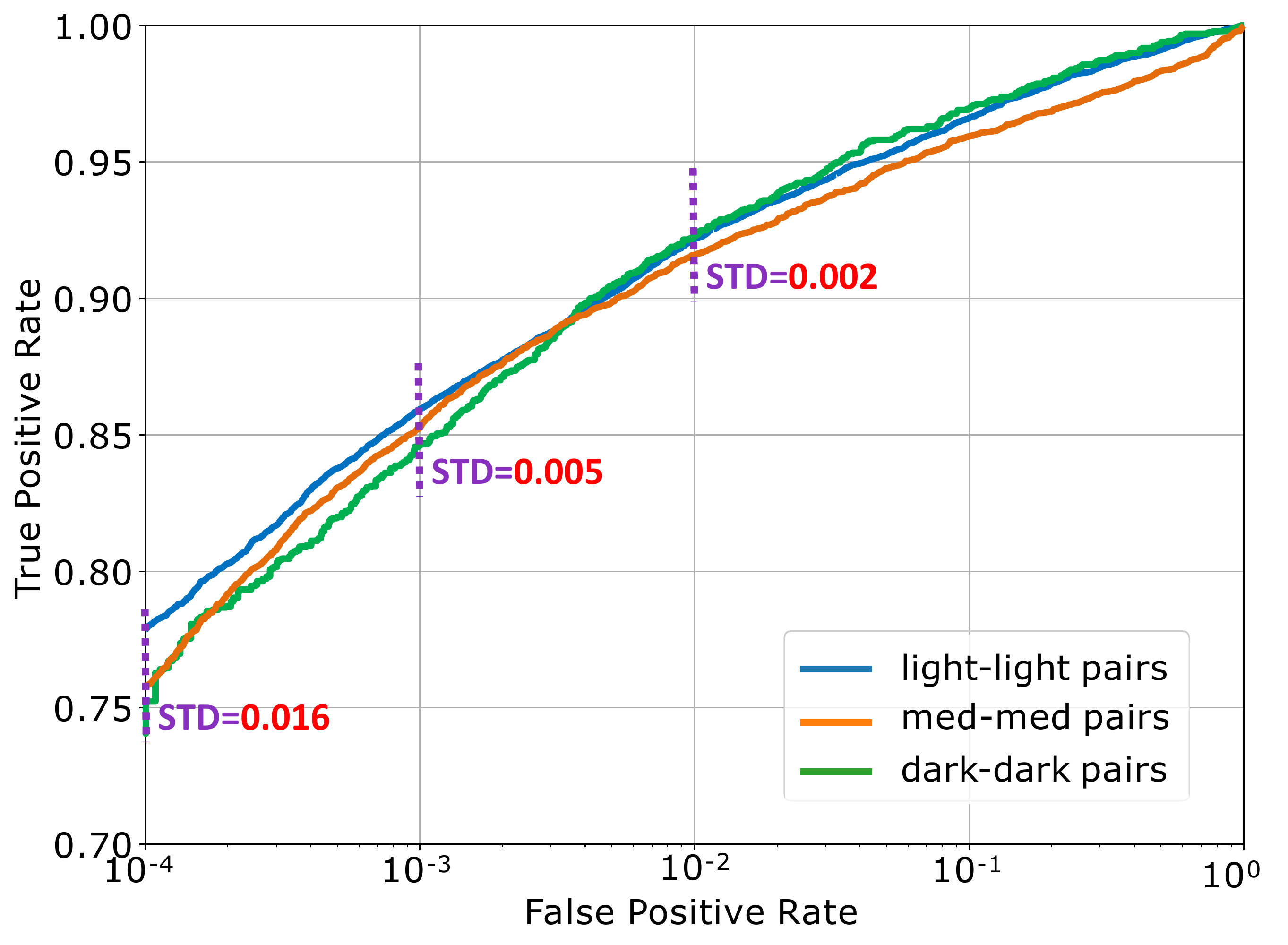}}\\
\vspace{-0.4cm}
\subfloat[OSD(s)]{\includegraphics[width=0.33\linewidth]{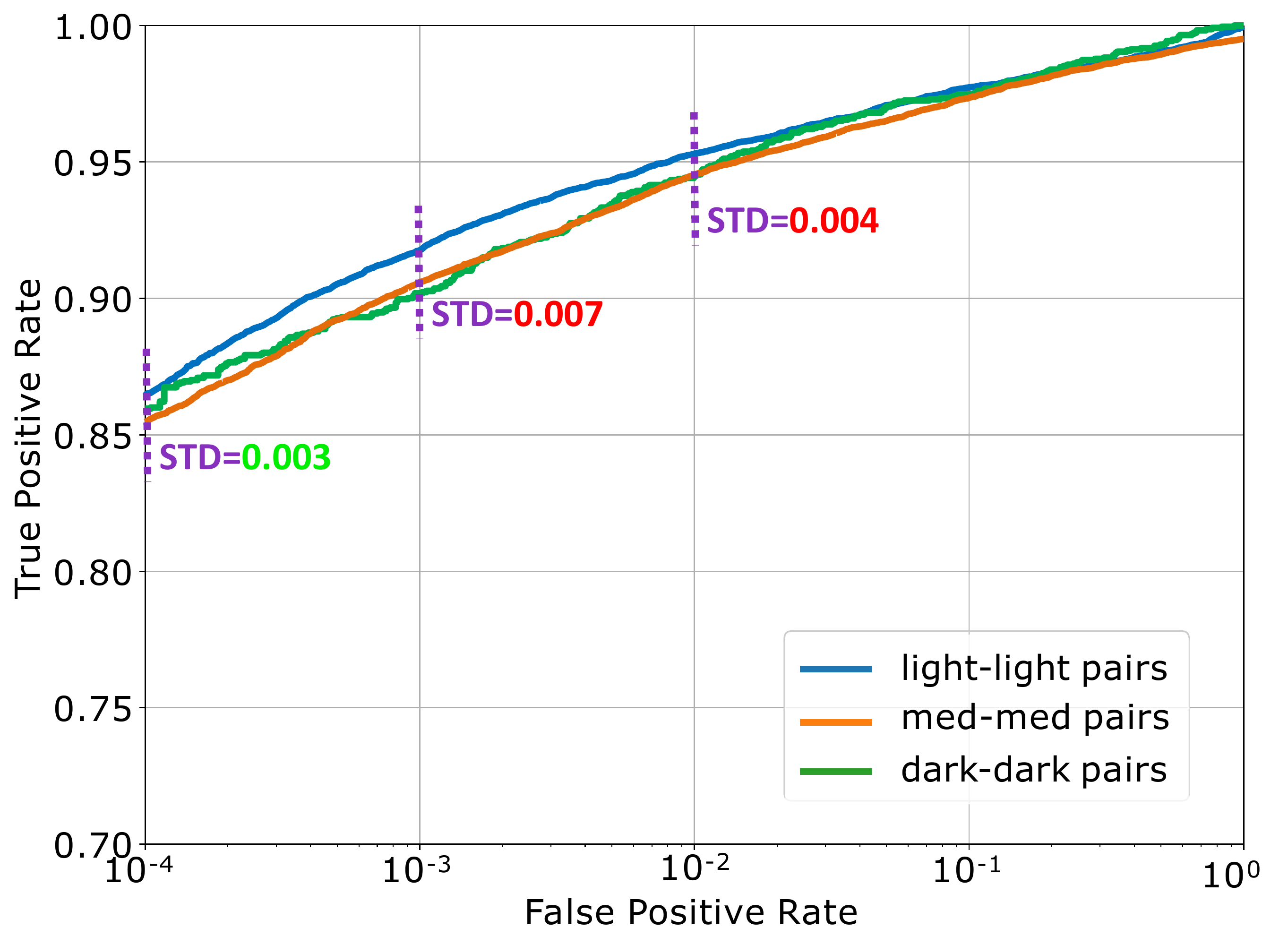}}
\subfloat[D\&D(s)]{\includegraphics[width=0.33\linewidth]{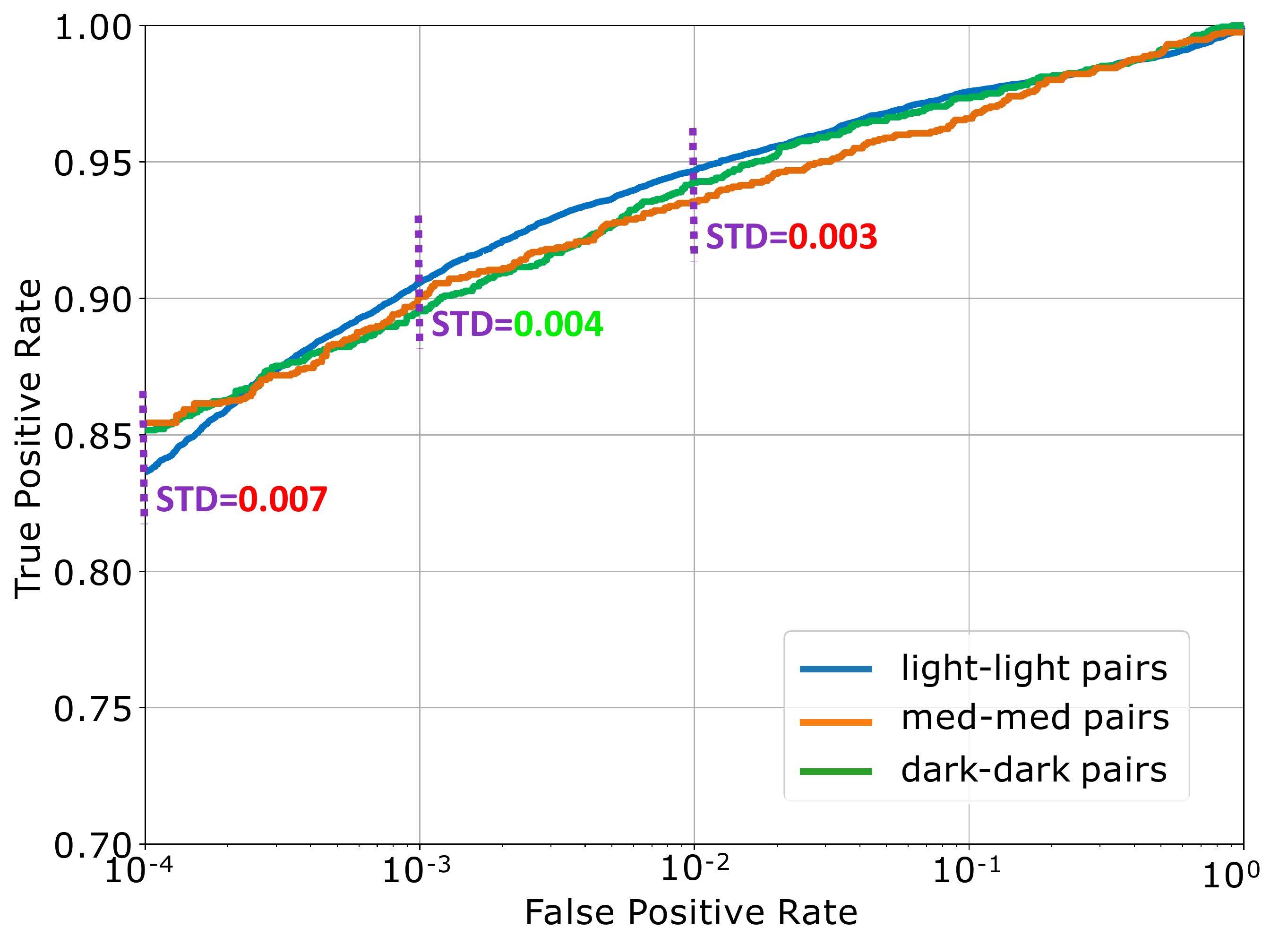}}
\subfloat[D\&D++(s)]{\includegraphics[width=0.33\linewidth]{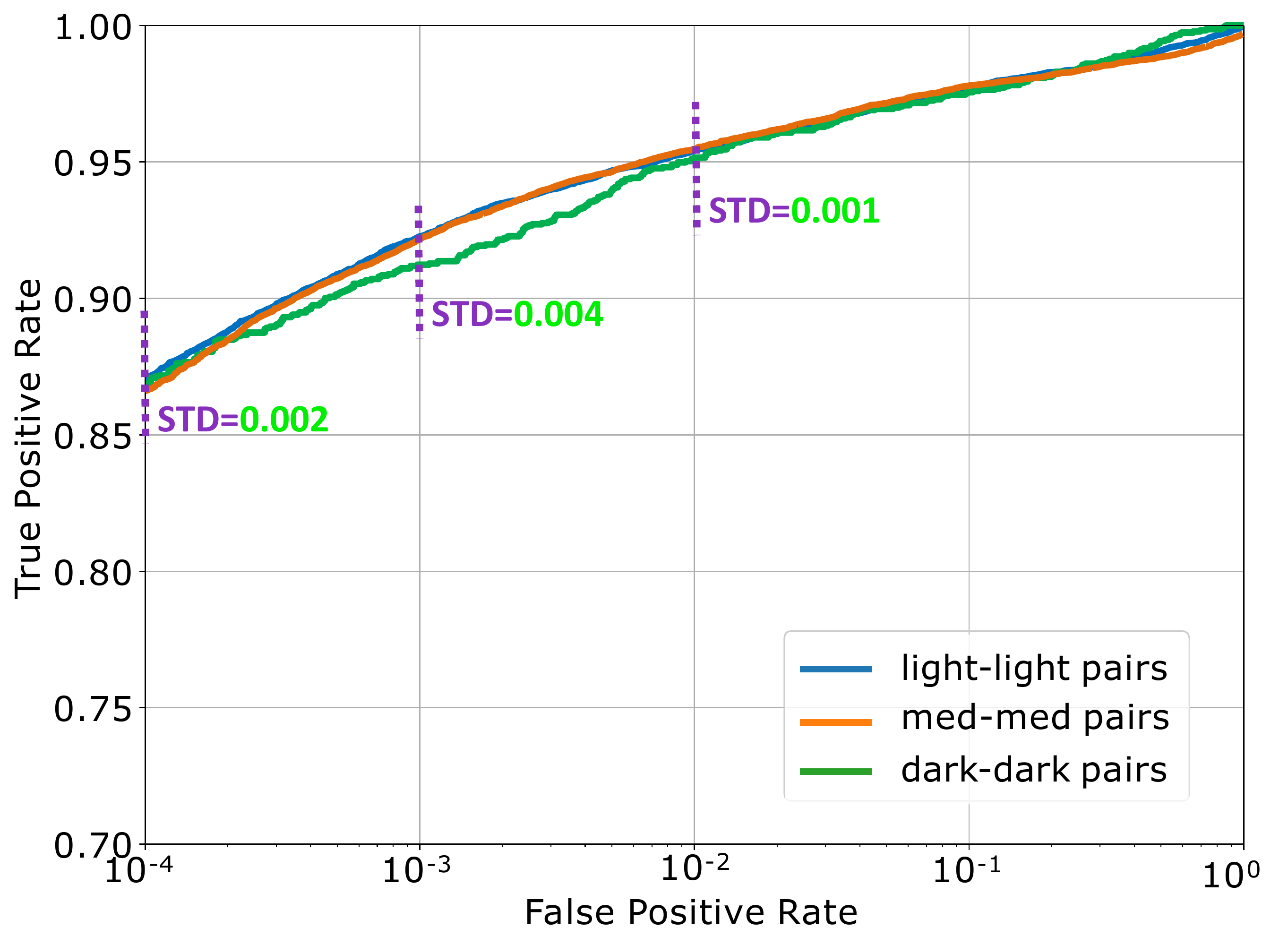}}
\vspace{-0.3cm}
\caption{\small Skintone-wise verification ROCs for all three skintones (light, medium, dark) in IJB-C for ArcFace and its de-biasing counterparts. The STD values in green denote the lowest STD at that FPR.  Best viewed when zoomed in. \vspace{-0.2cm}}
\label{fig:lmd}
}
\end{figure*}

 Even though we train D\&D(s) and D\&D++(s) on two skintone categories (Light and Dark), we also evaluate these systems on a third intermediate category i.e. `medium' skintone, provided by the IJB-C dataset. When considering all three skintone categories, Eq. \ref{eq:stbias} cannot be used to quantify bias, and hence we define skintone bias as the standard deviation (STD) among the verification TPRs of light-light pairs, medium-medium pairs and dark-dark pairs. This measure is inspired by previous works such as \cite{xu2021consistent,wang2020mitigating}. In Fig. \ref{fig:lmd}, we present the skintone-wise verification plots for all three categories and report the STD values among them. \textit{We find that our proposed D\&D(s)/D\&D++(s) obtain considerably lower STD than existing baselines, thus mitigating skintone bias.}
\subsection{Results with Crystalface}
\label{subsec:cfres}
\begin{figure*}
{
\centering
\subfloat[Crystalface]{\includegraphics[width=0.33\linewidth]{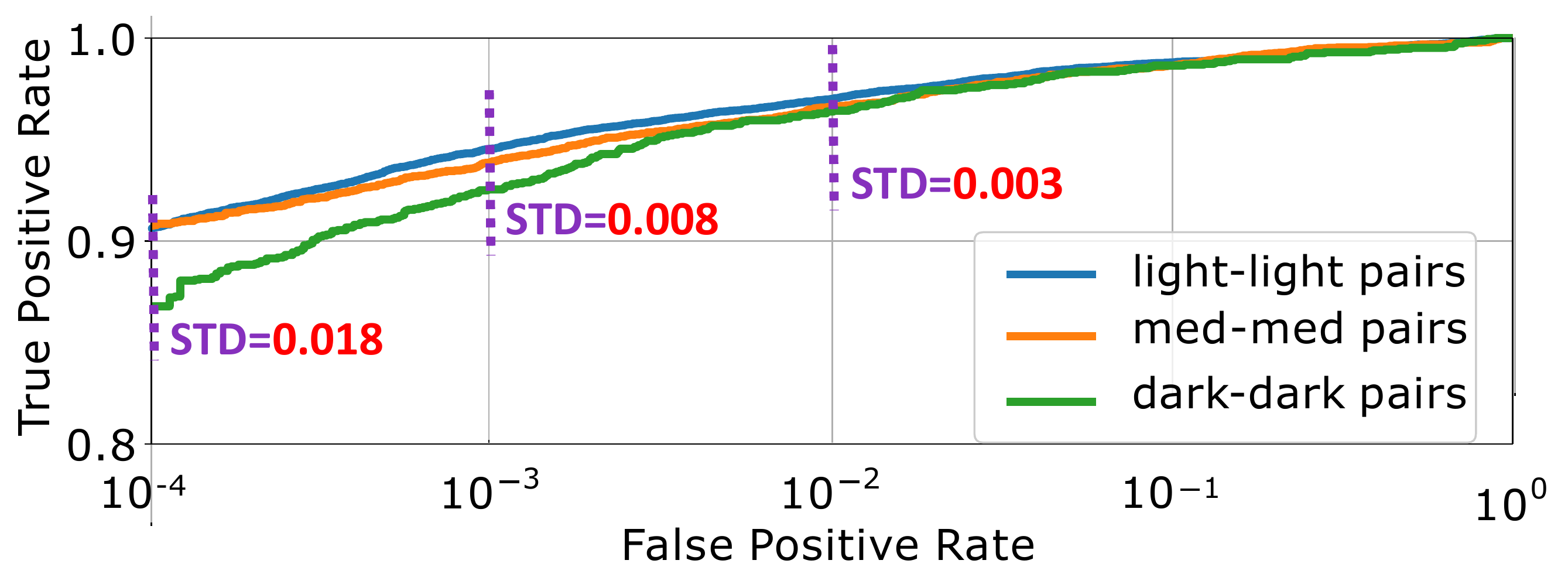}}
\subfloat[IVE(s)\cite{terhorst2019suppressing}]{\includegraphics[width=0.33\linewidth]{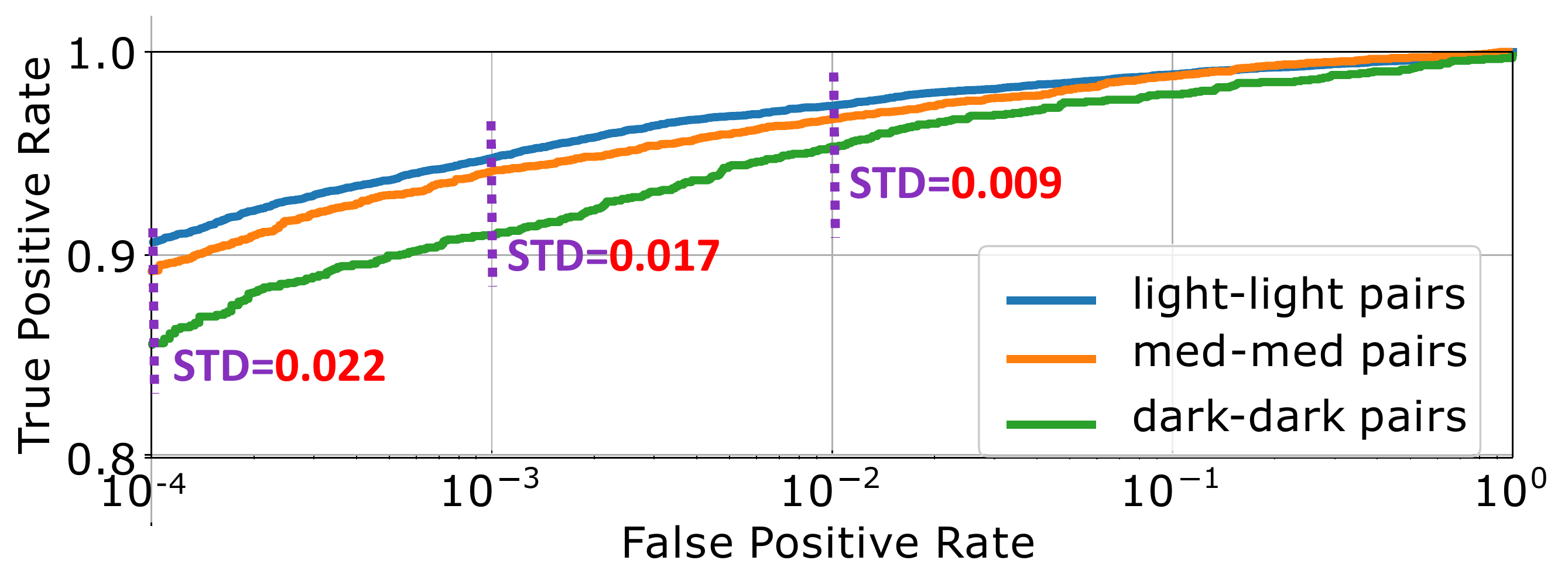}}
\subfloat[PASS-s\cite{Dhar_2021_ICCV}]{\includegraphics[width=0.33\linewidth]{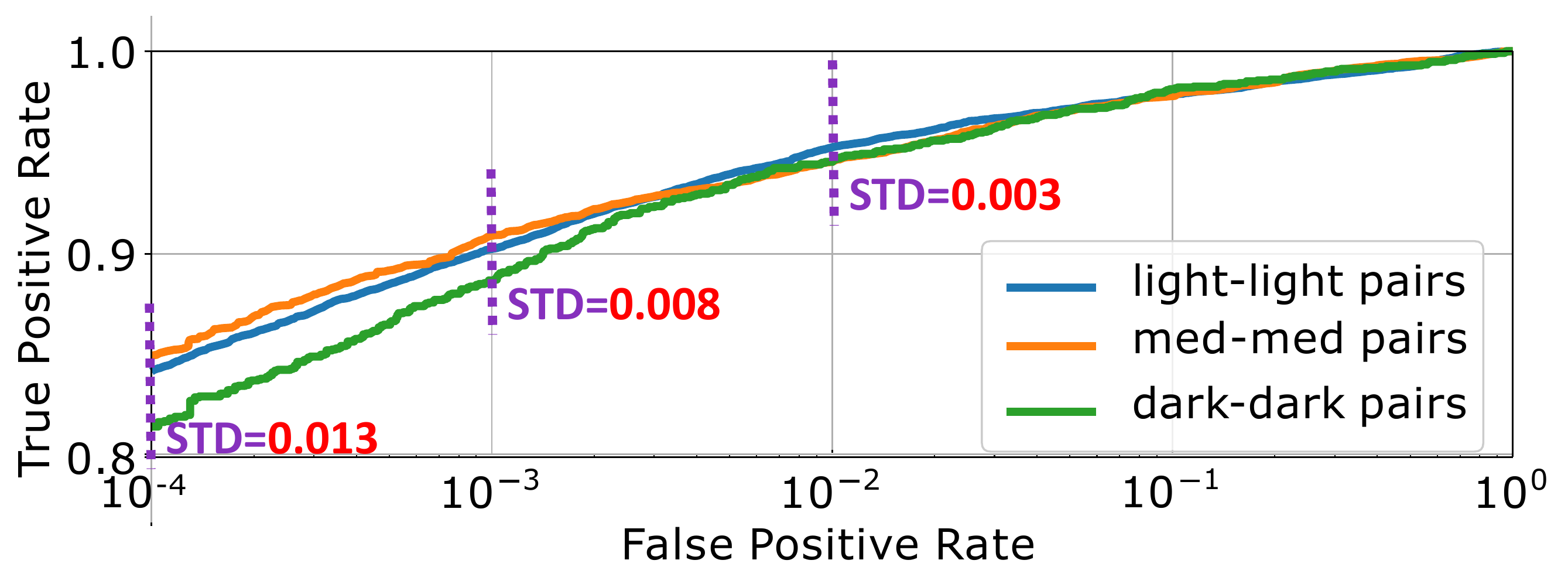}}\\
\vspace{-0.4cm}
\subfloat[OSD(s)]{\includegraphics[width=0.33\linewidth]{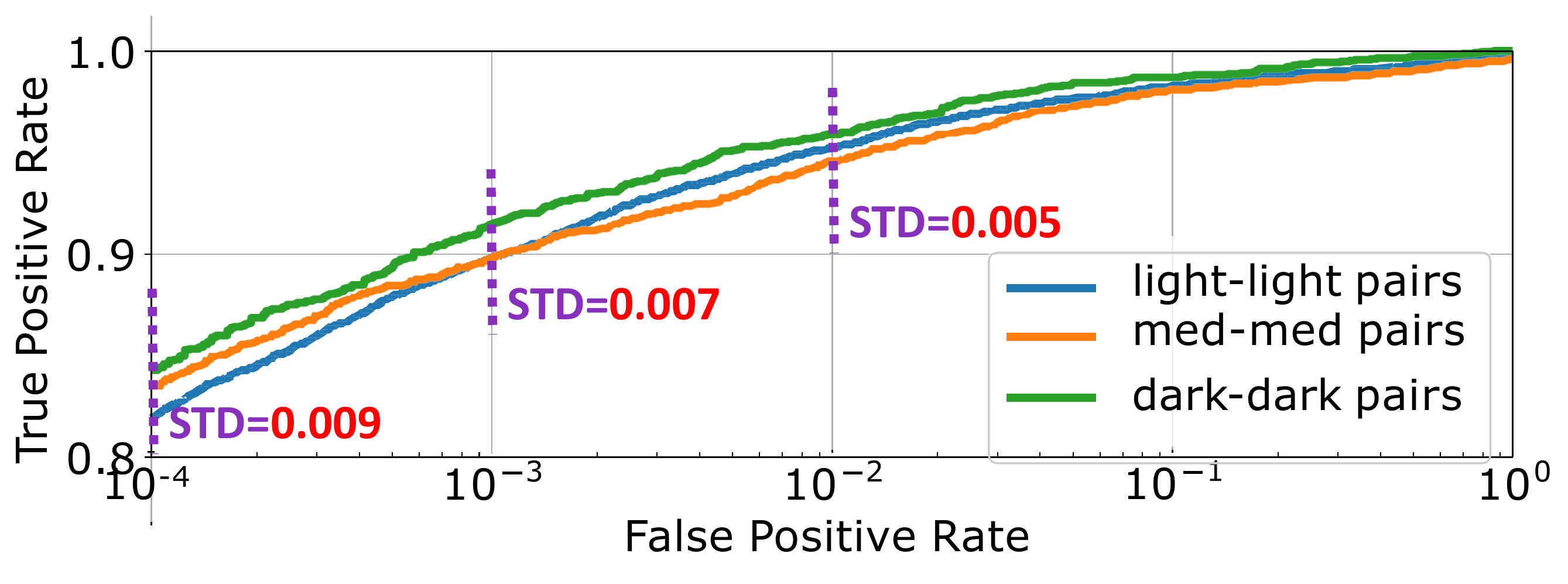}}
\subfloat[D\&D(s)]{\includegraphics[width=0.33\linewidth]{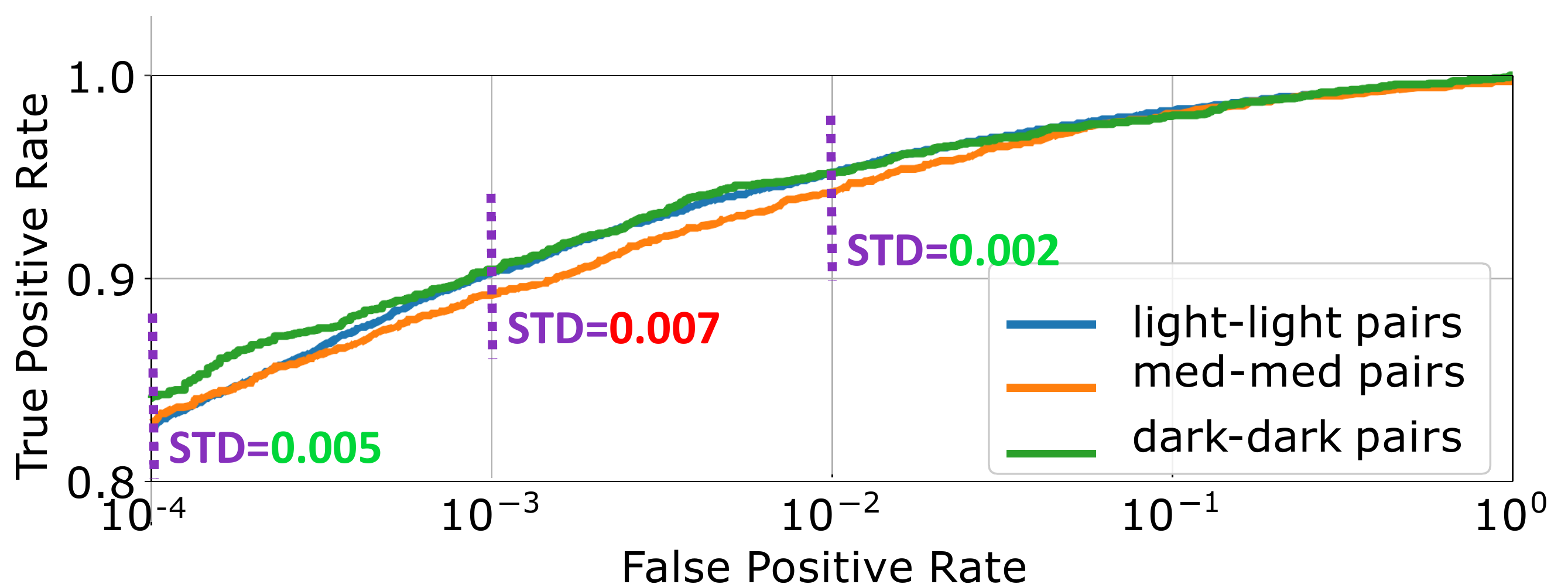}}
\subfloat[D\&D++(s)]{\includegraphics[width=0.33\linewidth]{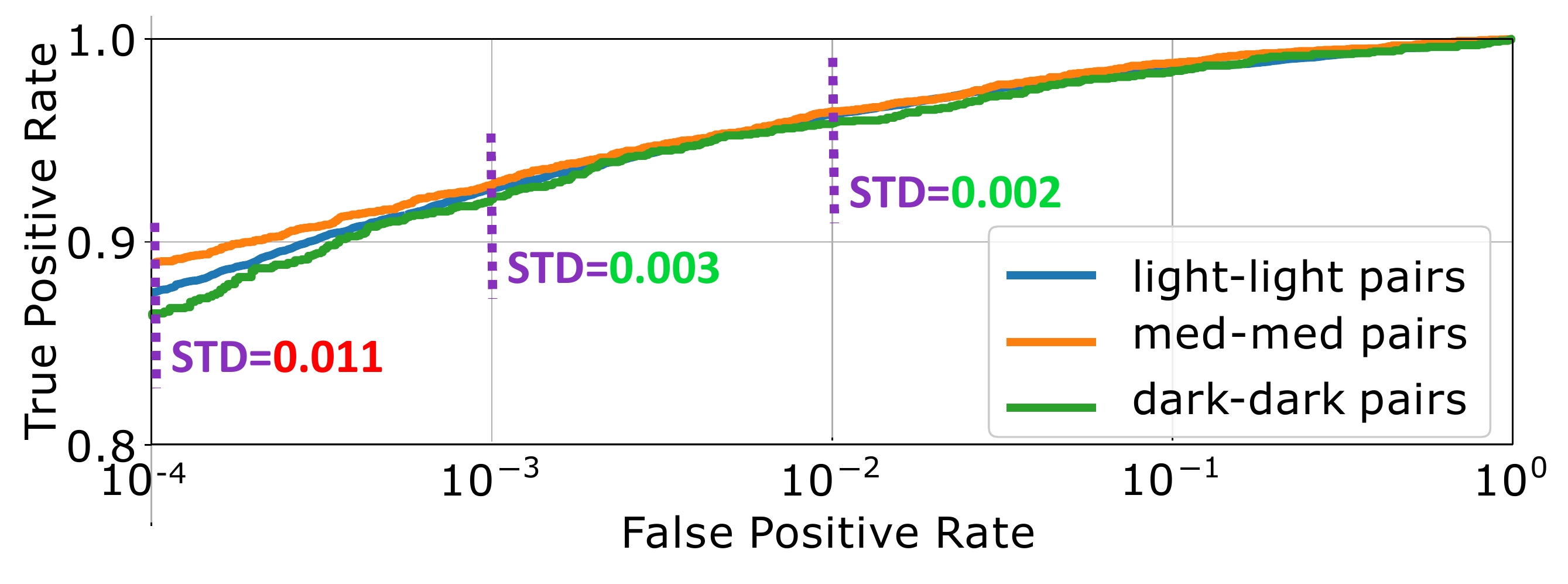}}
\vspace{-0.3cm}
\caption{\small Skintone-wise verification ROCs for all three skintones (light, medium, dark) in IJB-C for Crystalface and its de-biasing counterparts. Best viewed when zoomed in. \vspace{-0.2cm}}
\label{fig:lmdcrys}
}
\end{figure*}

% \begin{figure}
% {
% \centering
% \subfloat[]{\includegraphics[width=0.475\linewidth]{images/ArcFace_am_res_gender.pdf}\label{fig:arcamgender}}
% \rulesep
% \subfloat[]{\includegraphics[width=0.475\linewidth]{images/ArcFace_am_res_skintone.pdf}\label{fig:arcamst}}
% \vspace{-1em}
% \caption{\small (a) D\&D(g) and D\&D++(g) generate more similar attention maps for male and female frontal faces, as compared to ArcFace. (b)  D\&D(s) and D\&D++(s) generate more similar attention maps for dark and light frontal faces, than ArcFace.}
% \vspace{-0.3cm}
% \label{fig:avgmapArcFace}
% }
% \end{figure}
%\begin{figure}
%{
%\centering
%\subfloat[Gender bias plots]{\includegraphics[width=0.5\linewidth]{images/aaai_af_gender_bupt_dnd_barplot_all7.pdf}}
%\subfloat[]{\includegraphics[width=0.25\linewidth]{images/aaai_skintone_ArcFace_overall_dnd_bupt_skintone_latest.pdf}}
% \subfloat[Skintone bias plots]{\includegraphics[width=0.5\linewidth]{images/aaai_af_skintone_bupt_dnd_barplot_all6_new.pdf}}
% \vspace{-0.3cm}
% \caption{\small D\&D or D\&D++ obtain the lowest gender and skintone bias for ArcFace in IJB-C at most FPRs.  \textit{Please see Sec. \ref{sec:ArcFace} and supplementary material for our implementation of baselines.}}\vspace{-0.5cm}
% % \caption{\small ArcFace-based D\&D/D\&D++ obtain lowest (a)Gender and (b)Skintone bias in IJB-C than other baselines using ArcFace backbone at most FPRs.}\vspace{-0.75cm}
% \label{fig:afbiasplot}
% }
% \end{figure}
To evaluate the generalizability of baselines and our D\&D variants, we repeat the aforementioned experiments using Crystalface network \cite{ranjan2019fast}. \textit{D\&D and D\&D++ generate more similar average attention maps for male and female faces (Fig. \ref{fig:amgender}); and for faces with dark and light skintone (Fig. \ref{fig:amst})}, than the original Crystalface network. More qualitative examples are provided in Figs. \ref{fig:cfqualresgender},\ref{fig:cfqualresst}. We also implement the gender-debiasing baselines (Hair obscuring, IVE(g), PASS-g) and skintone-debiasing baselines (IVE(s), PASS-s) using Crystalface to make fair comparison with D\&D-based methods. \textit{Crystalface networks trained with D\&D/D\&D++ obtain the lowest gender and skintone bias (Figs. \ref{fig:cfgenplot},\ref{fig:cfstplot}), and highest BPC scores (Table \ref{tab:cfallbias})}. Also, D\&D++ achieves considerably higher TPR than D\&D (Table \ref{tab:cfallbias}). Also, similar to Sec. \ref{subsubsec:third}, we evaluate our Crystalface-based D\&D(s), D\&D++(s) and baselines (trained on light and dark skintones) on the intermediate skintone category (i.e. `medium'). In Fig. \ref{fig:lmdcrys}, we provide the verification plots for all three skintone categories and also report and report the STD values among them. \textit{D\&D(s)/D\&D++(s) obtain considerably lower STD than existing baselines, thus mitigating skintone bias}. The hyperparameter information and detailed results for all the methods are provided in the supplementary material.\vspace{-0.1cm}
\begin{table*}[]
\centering
%\scriptsize
\subfloat[Gender bias - Crystalface backbone]{\scalebox{0.73}{
\hspace{-10pt}\begin{tabular}{c|cc|cc|cc}
\toprule
 FPR & {  $10^{-5}$} & {  } & {  $10^{-4}$} & {  }  & {  $10^{-3}$}  & {  } \\
 \midrule
  Method & TPR & \hspace{-4pt} BPC\textsubscript{g}$(\uparrow)$ \hspace{-4pt} & TPR & \hspace{-4pt} BPC\textsubscript{g}$(\uparrow)$ \hspace{-4pt} & TPR & \hspace{-4pt} BPC\textsubscript{g}$(\uparrow)$ \hspace{-4pt} \\
  \midrule
Crystalface \hspace{-6pt}&  0.856  & 0 & 0.912 & 0 &  0.950 & 0 \\
IVE(g)\textsuperscript{$\dag$} \cite{terhorst2019suppressing} & 0.840 & \underline{0.768} & 0.910 & 0.365 & 0.952 & 0.389 \\
W/o hair\textsuperscript{$\dag$} \cite{albiero2020face} & 0.592 & -3.441 & 0.803 & 0.615 & 0.899 & 0.301 \\
PASS-g\textsuperscript{$\dag$} \cite{Dhar_2021_ICCV} &0.691  & 0.687& 0.842   &0.291  &0.914   &  0.027 \\
\midrule
OSD(g) & 0.721  & 0.482 & 0.817 & 0.631& 0.895  &0.297  \\
\rowcolor{Gray}
D\&D(g) & 0.705  & 0.650 & 0.805 & \textbf{0.719} & 0.888 &\underline{0.515} \\
\rowcolor{Gray}
D\&D++(g) & 0.754  & \textbf{0.854} & 0.844  & \underline{0.701}& 0.914 & \textbf{0.898}\\
\bottomrule
\end{tabular}
}\label{tab:cfgenbias}}
\subfloat[Skintone bias - Crystalface backbone]{\scalebox{0.73}{
\begin{tabular}{c|cc|cc|cc}
\toprule
 FPR &  {  $10^{-4}$} & {  }  & {  $10^{-3}$} & {  } & {  $10^{-2}$} & {  } \\
 \midrule
  Method & TPR & \hspace{-4pt} BPC\textsubscript{st}$(\uparrow)$ \hspace{-4pt} & TPR  & \hspace{-4pt} BPC\textsubscript{st}$(\uparrow)$ \hspace{-4pt} & TPR & \hspace{-4pt} BPC\textsubscript{st}$(\uparrow)$ \hspace{-4pt} \\
  \midrule
Crystalface & 0.912 & 0 &  0.950 & 0 & 0.973 & 0 \\
IVE(s)\textsuperscript{$\dag$} \cite{terhorst2019suppressing} & 0.910  &-0.371 &0.950 & -0.900 &0.974 & -2.49 \\
PASS-s\textsuperscript{$\dag$} \cite{Dhar_2021_ICCV} & 0.851 & 0.222 & 0.910 & 0.158 &0.953 & -0.187 \\
\midrule
OSD(s) & 0.848 & 0.351 & 0.916 &  0.214 & 0.961  & -0.012 \\
\rowcolor{Gray}
D\&D(s) & 0.850 & \textbf{0.643}& 0.916 & \textbf{0.914} & 0.961 &\textbf{0.821} \\
\rowcolor{Gray}
D\&D++(s) & 0.886 & \underline{0.629} &  0.934& \underline{0.733} & 0.967 &\underline{0.327}\\
\bottomrule
\end{tabular}
}\label{tab:cfstbias}}
%\vspace{-0.3cm}
\caption{\small \textbf{(a)}\textit{Gender} and \textbf{(b)}\textit{Skintone} bias analysis for \textit{Crystalface} network, and its de-biased counterparts on IJB-C. \textbf{Bold}=Best, \underline{Underlined}=Second best. D\&D variants obtain highest BPC\textsubscript{g} at most FPRs. \textsuperscript{$\dag$}=Our implementation of baselines (Refer to Sec. \ref{sec:baseline} for details). All the methods are trained on BUPT-BalancedFace \cite{wang2020mitigating} dataset.} \label{tab:cfallbias}
\end{table*}
\subsection{Bias vs. face verification performance}
\begin{figure*}
{
\centering
%\subfloat[\small TPR v/s Gender bias (CF)]{\includegraphics[width=0.25\linewidth]{images/cfgb.png}}~
%\subfloat[\small TPR v/s Skintone bias (CF)]{\includegraphics[width=0.25\linewidth]{images/cfstb.png}}
%\vspace{-0.1cm}
\subfloat[\small TPR vs. Gender bias ]{\includegraphics[width=0.25\linewidth]{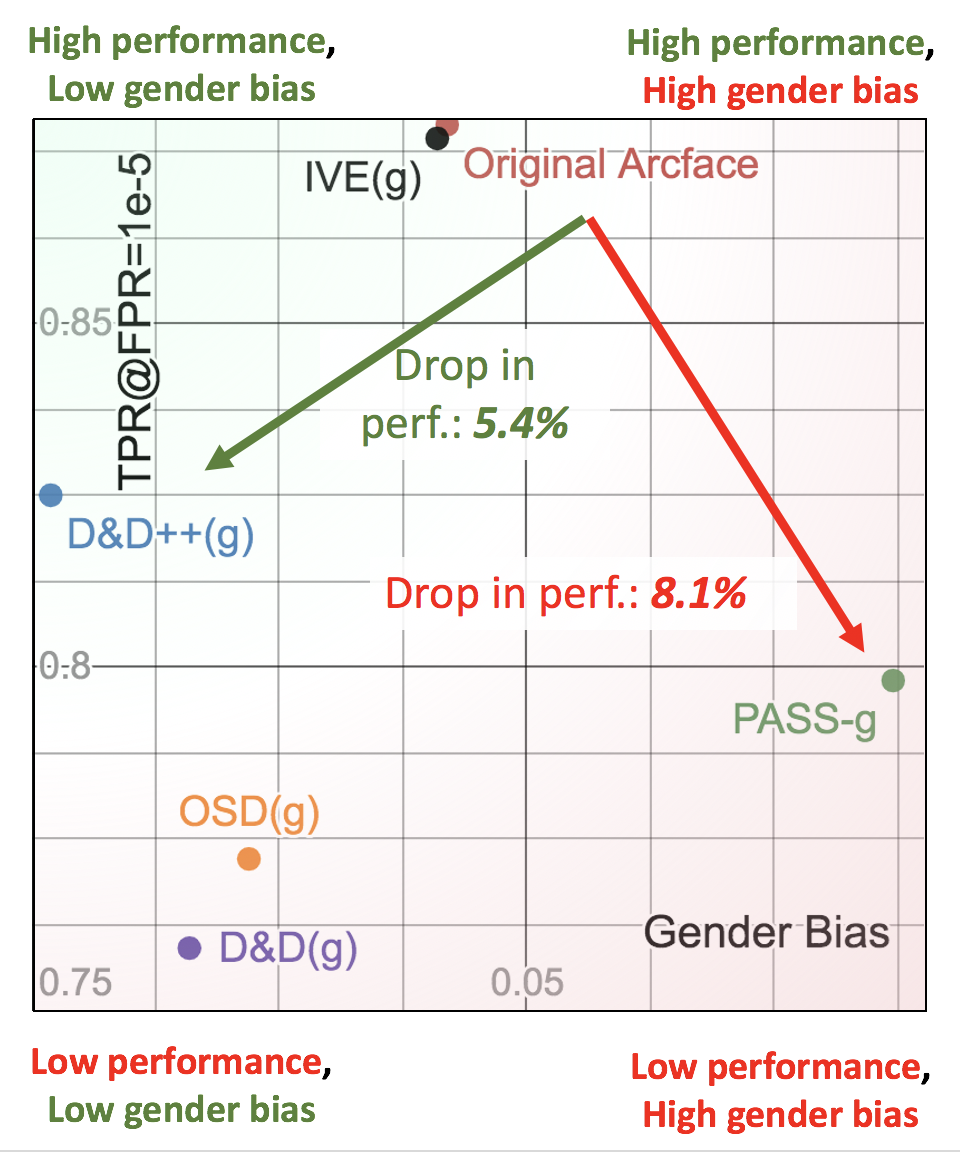}}
\subfloat[\small TPR vs. Gender bias]{\includegraphics[width=0.25\linewidth]{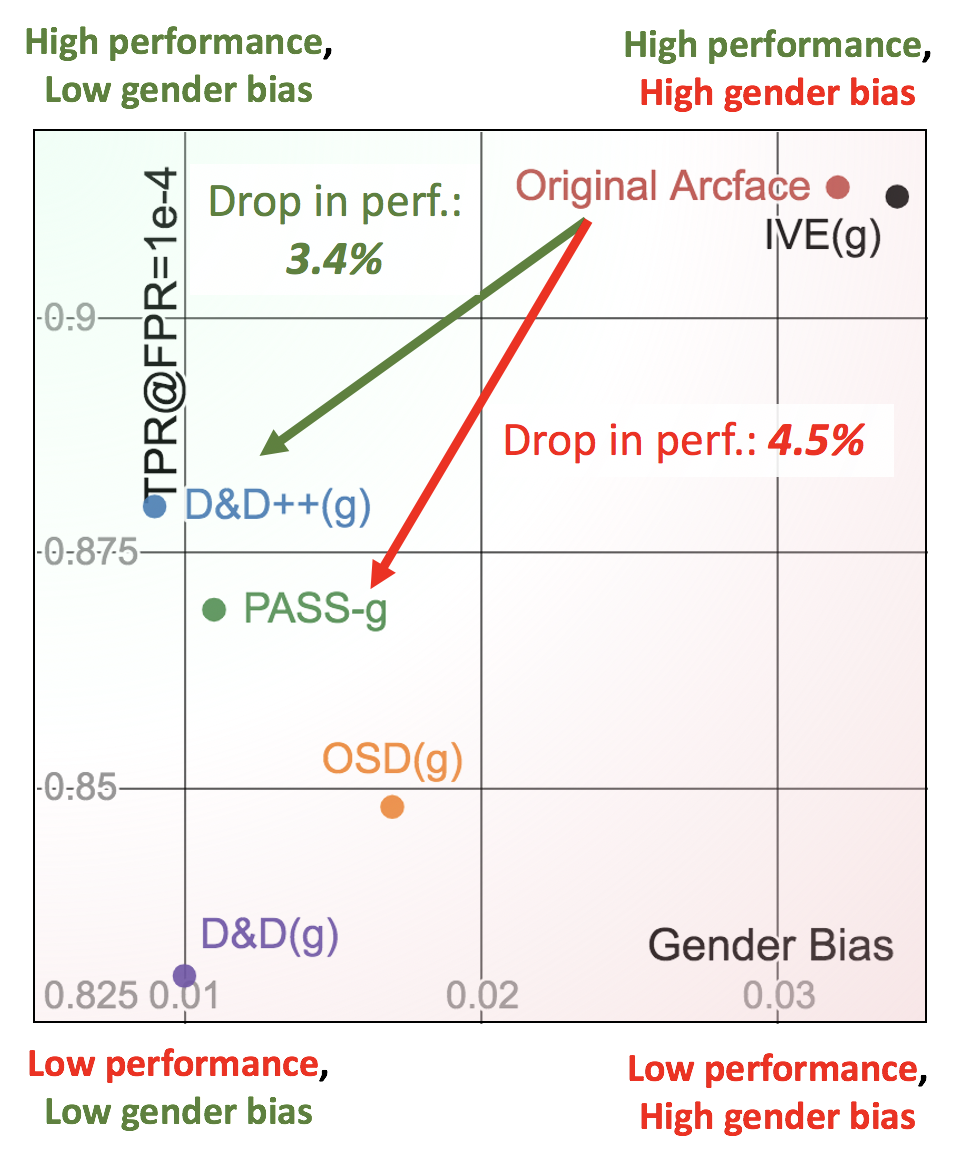}}
\subfloat[\small TPR vs. Skintone bias ]{\includegraphics[width=0.25\linewidth]{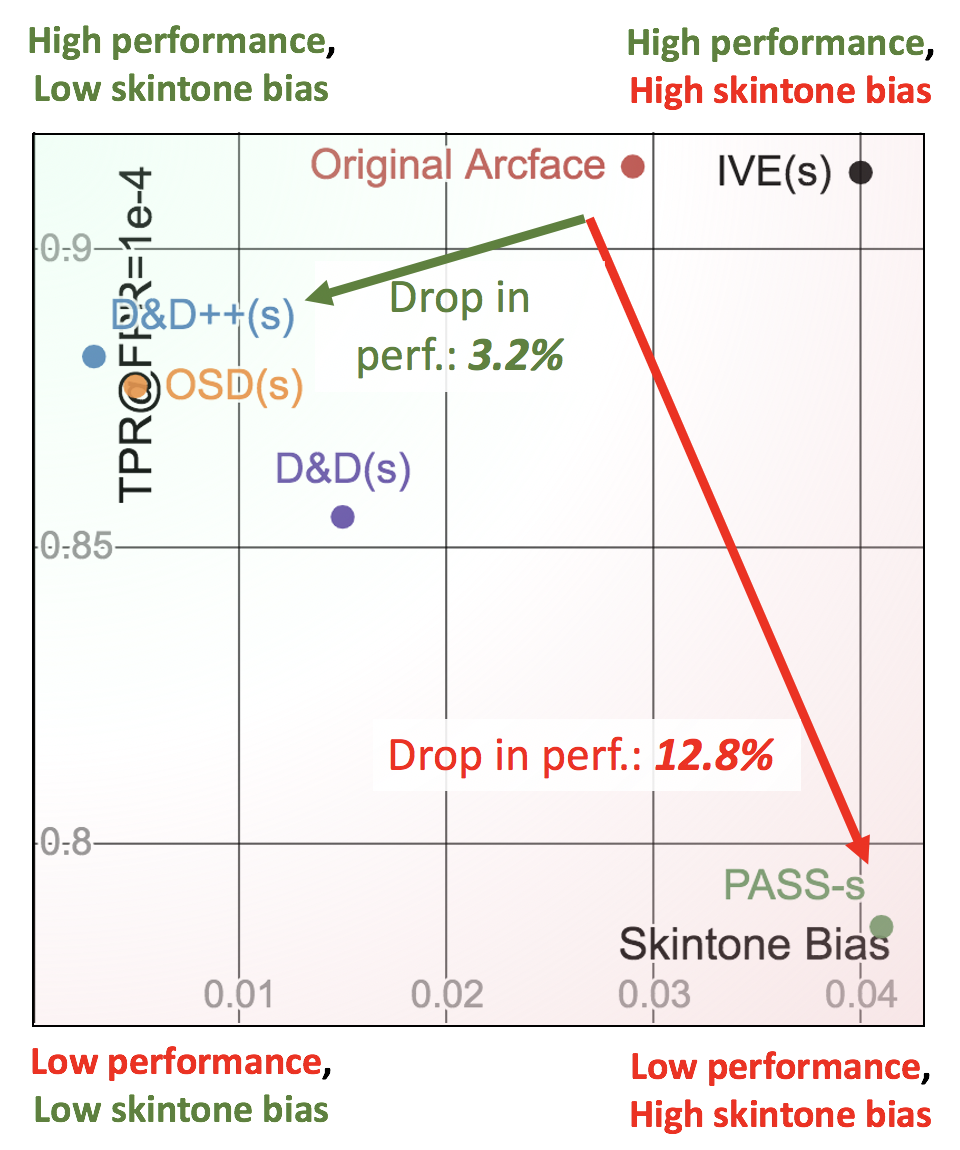}}
\subfloat[\small TPR vs. Skintone bias ]{\includegraphics[width=0.25\linewidth]{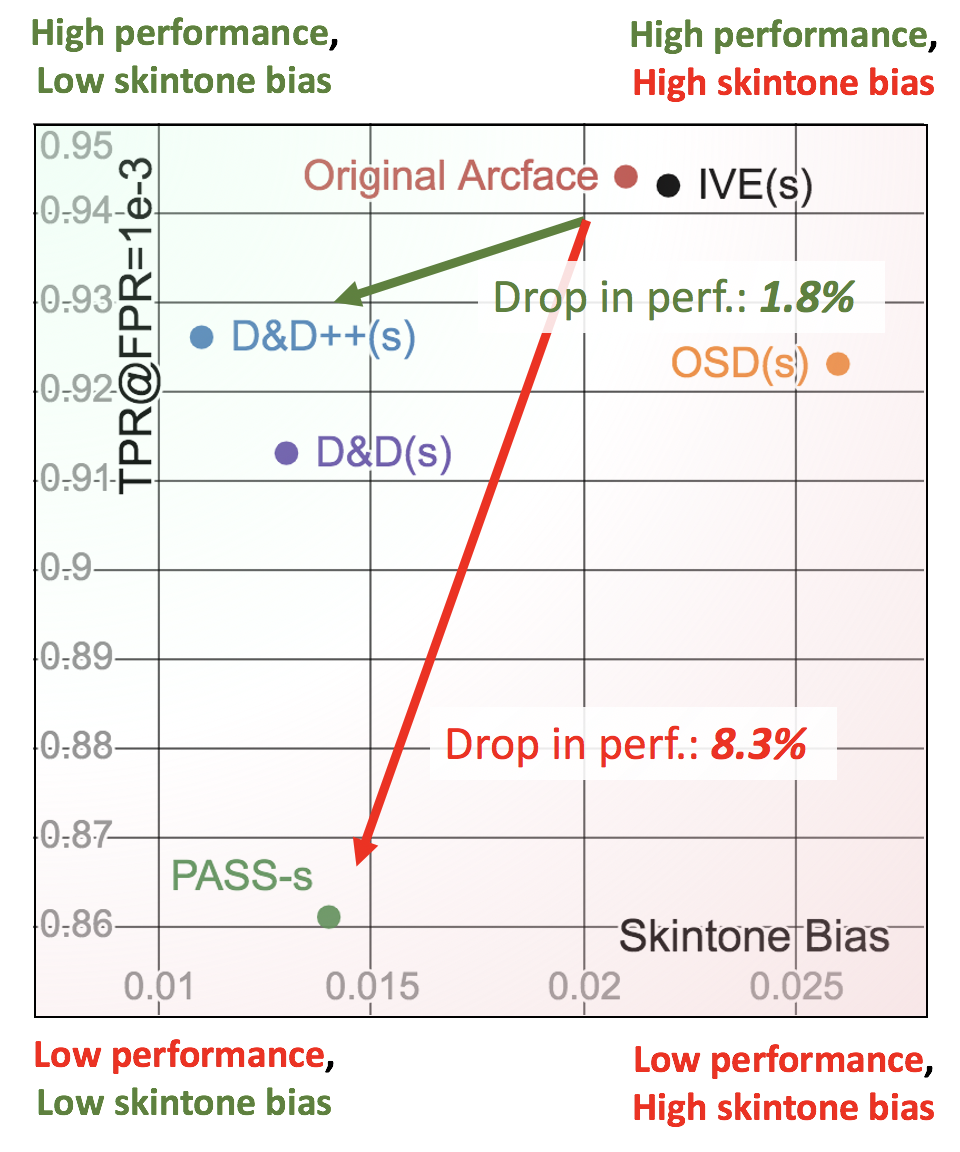}}
\vspace{-0.3cm}
\caption{\small TPR at a fixed FPR vs. Gender/Skintone bias demonstrated by de-biasing methods. Ideally, a de-biasing method would occupy the upper-left-hand corner, where performance is high, and bias is low (\textit{mostly occupied by D\&D++}). The lower right-hand is the worst case which decreases performance without reducing bias. The red and green arrows indicate the drop in TPR in PASS and D\&D++, respectively.}
\label{fig:tradeoff}
}
\end{figure*}

Although an ideal de-biasing system should reduce bias while maintaining face verification performance, existing adversarial de-biasing algorithms such as PASS \cite{Dhar_2021_ICCV}, DebFace \cite{gong2020jointly} etc. demonstrate significant drop in face verification performance. In this work, we provide insights for reducing bias while minimizing the drop in face verification performance by presenting D\&D++, which is a non-adversarial approach. We recognize that D\&D++(s/g) also demonstrates a slight drop in face verification performance. This drop may be caused by the explicit distillation constraint ($L_{dis}$) imposed on the D\&D++ student $M^{*}_s$ that restricts $M^{*}_s$ from learning all the gender or skintone specific details. However, compared to de-biasing methods such as PASS \cite{Dhar_2021_ICCV} that explicitly remove protected attributes from face representations, \textit{the drop in verification performance with D\&D++ is considerably lower}. Moreover, the goal of this work is \textit{\underline{not}} to obtain SOTA verification accuracy, but to provide a better alternative to adversarial de-biasing techniques. D\&D++ clearly achieves this goal, as shown in Fig. \ref{fig:tradeoff}. Hence, we believe that non-adversarial methods like D\&D++ are more practical for reducing bias, since adversarial methods significantly lower the identity classifying capability of a network to achieve `fairness through blindness'.
\section{Conclusion}
We present two novel knowledge distillation-based techniques (D\&D and D\&D++) to incrementally learn different categories of an attribute. We observe that our proposed methods enforce the networks to attend to similar spatial regions of the face for both categories of an attribute and consequently reduce bias w.r.t. that attribute. Both D\&D and D\&D++ outperform the existing baselines in reducing gender and skintone bias. D\&D++, while being less biased than baselines, generally obtains better face verification performance than SOTA adversarial de-biasing algorithms. We also show the generalizability of our methods on two SOTA face recognition networks. 
%--------------------------------------------------------------------
{\small
\bibliographystyle{ieee_fullname}
\bibliography{egbib}
}
\setcounter{section}{0}
\renewcommand{\thesection}{A\arabic{section}}  
\setcounter{table}{0}
\renewcommand{\thetable}{A\arabic{table}}  
\setcounter{figure}{0}
\renewcommand{\thefigure}{A\arabic{figure}}
\setcounter{equation}{0}
\renewcommand{\theequation}{A\arabic{equation}}
\section*{Supplementary material}
In this supplementary material, we provide the following information:\\
\textbf{Section \ref{supp_sec:relation}}: Relation between face recognition and face verification.\\
\textbf{Section \ref{supp_sec:fairness}}: A novel interpretation of the bias measure (introduced in \cite{Dhar_2021_ICCV}) as a metric for Equality of Odds.\\
\textbf{Section \ref{supp_sec:dndinfo}}: Training details for OSD, D\&D, and D\&D++. \\
\textbf{Section \ref{supp_sec:detailedresult}}: Detailed results with ArcFace (Section \ref{supp_sec:afdetailresult}) and Crystalface (Section \ref{supp_sec:cfdetailresult}) backbones, including verification ROCs. \\
\textbf{Section \ref{supp_sec:passinfo}}: Training details for PASS \cite{Dhar_2021_ICCV} baselines, following the official implementation \cite{passcode}.\\
\textbf{Section \ref{supp_sec:iveinfo}}: Training details for IVE, following the official implementation \cite{ivecode}.\\
\textbf{Section \ref{supp_sec:bowyerinfo}}: Pipeline for obscuring hair (similar to \cite{albiero2020face}). \\
\begin{table}[]
\centering
%\scriptsize
\scalebox{0.8}{
\hskip-0.2cm\begin{tabular}{c|c}
\toprule

    Table/Fig. & Summary  \\
  \midrule
  Table \ref{supp_tab:hpdnd} & Hyperparameters for D\&D,D\&D++,OSD \\
  \midrule
  Fig. \ref{supp_fig:afgstwise}& \thead{Gender \& Skintone-wise verification plots for\\ ArcFace and its debiasing counterparts}\\
  \midrule
  Table \ref{supp_tab:arcstd}& \thead{Tabular values for \textbf{Figure 6} from the main paper}\\
  \midrule
  Fig. \ref{supp_fig:cfwise}& \thead{Gender \& Skintone-wise verification plots for\\ Crystalface and its debiasing counterparts}\\
  \midrule
  Table \ref{supp_tab:cfallbias} & \thead {Gender and skintone bias analysis for Crystyalface-based methods \\ (Extension of \textbf{Table 3} from the main paper)}\\
  \midrule
  Table \ref{supp_tab:crystd}& \thead{Tabular values for \textbf{Figure 7} from the main paper}\\
  \midrule
  Table \ref{supp_tab:hppass} & Hyperparameters for PASS\\
\bottomrule
\end{tabular}
}
\caption{ \textbf{Summary}: For the readers' convenience, we provide a brief summary of the important tables and figures in this supplementary material.} \label{supp_tab:summary}
\end{table}

% \section{Bias measure as metric for Equality of Odds}
\section{Relation between face recognition and face verification}
A face recognition network is trained to classify the identities in a training dataset. Here, we briefly describe the relationship between face recognition and face verification. Any task (such as face veification, identification, authentication etc.) that requires a system to recognize a representation of the input face comes under the umbrella of face recognition.  The most common tasks in the literature \cite{ranjan2019fast,deng2018ArcFace,Swami_2016_triplet} that are used to evaluate a face recognition network are:\\
(i) Face Verification: As defined in \cite{ranjan2019fast}, the aim of this task is to determine if a given pair of templates (i.e. two sets of face representations) belong to the same or different identity. These representations are extracted using previously trained networks. This is also referred to as 1:1 verification.\\
(ii)Face Identification: The aim of this task is to match a probe template to a collection of templates corresponding to many identities; such a collection is referred to as a gallery. This is also referred to as `1:$N$ search'.\\

In the context of mitigating bias, most face recognition networks are evaluated in terms of their face verification performance on different demographic groups, as done in \cite{gong2020jointly,Dhar_2021_ICCV,gac}. Following this, we also evaluate the bias mitigation in face recognition with respect to the face verification task.
\label{supp_sec:relation}
\section{Zero Bias implies Equality of Odds}
\label{supp_sec:fairness}
We use the bias measures introduced in previous bias mitigation work \cite{Dhar_2021_ICCV}. Here, we show that it may be viewed as a measure of \textit{equality of odds} \cite{hardt2016equality} for pair-wise matching in the sense that achieving zero bias (as defined in Eq~\ref{supp_eq:abias}) allows us to achieve equality of odds.

First, we define \text{bias} at a false positive rate (FPR) of $F$ with respect to attribute $A$ as
\begin{equation}
    \text{Bias}^{(F)} = |\text{TPR}^{(F)}_{a_0} - \text{TPR}^{(F)}_{a_1}|,
    \label{supp_eq:abias}
\end{equation}
where $\text{TPR}^{(F)}_{a_*}$ denotes the true positive rate (TPR) on pairs of faces with attribute $A=a_*$ at FPR of $F$.

Next, we show how achieving zero bias in equation~\ref{supp_eq:abias} satisfies equalized odds. First let $\mathcal{F}$ be the set of all face images and let $A : \mathcal{F}\rightarrow \{0, 1\}$ be an indicator on a binary attribute of a face where $0$ corresponds to $a_0$ and $1$ corresponds to $a_1$. Let $\Omega_A \equiv \{(f_1,f_2)\in \mathcal{F}\times\mathcal{F}~|~A(f_1) = A(f_2)\}$ be the set of all pairs of faces with matching attributes and let $Y : \Omega_A \rightarrow \{0,1\}$ indicate identity equivalence for a pair of faces. Since all pairs in $\Omega_A$ consist of faces with equal values of $A$ we extend $A$ onto $\Omega_A$ such that $A(f_1,f_2) = A(f_1)$ for all $(f_1,f_2) \in \Omega_A$. Finally, let $\hat{Y} : \Omega_A \rightarrow \{0,1\}$ be a predictor of $Y$. Supposing $\omega \in \Omega_A$ is random pair of faces sampled from $\Omega_A$, then we have \textit{equality of odds} if and only if
\begin{multline}
    \label{supp_eq:eqoddstpr}
    P(\hat{Y}(\omega)=1|A(\omega)=0,Y(\omega)=1) = \\ P(\hat{Y}(\omega)=1|A(\omega)=1,Y(\omega)=1)
\end{multline}
and
\begin{multline}
    \label{supp_eq:eqoddsfpr}
    P(\hat{Y}(\omega)=1|A(\omega)=0,Y(\omega)=0) = \\ P(\hat{Y}(\omega)=1|A(\omega)=1,Y(\omega)=0).
\end{multline}
Equation~\ref{supp_eq:eqoddstpr} is equivalent to $\text{TPR}^{(F)}_{a_0} = \text{TPR}^{(F)}_{a_1}$ for a fixed FPR of $F$, while equation~\ref{supp_eq:eqoddsfpr} corresponds to a equal FPR for both $a_0$ and $a_1$ pairs.

Equation~\ref{supp_eq:eqoddstpr} is clearly satisfied when the bias measure in equation~\ref{supp_eq:abias} is zero. Equation~\ref{supp_eq:eqoddsfpr} is satisfied by selecting two appropriate thresholds, one for pairs with attribute $A=a_0$ and another for pairs with attribute $A=a_1$. In this way, we have shown that minimizing the bias term defined in equation~\ref{supp_eq:abias} works towards achieving equalized odds in pair-wise face matching.
\section{Training details for D\&D, D\&D++ and OSD}
In this section, we provide hyperparameter and training details for our proposed methods (D\&D and D\&D++) and OSD.
\label{supp_sec:dndinfo}
\begin{table}%
  \centering
\begin{tabular}{ccc}
\toprule
Method/Backbone  & ArcFace & Crystalface\\
\midrule
OSD(g)&$\lambda_{osd}=1.0$&$\lambda_{osd}=0.8$\\
D\&D(g)&$\lambda_1=1.0$& $\lambda_1=1.0$\\
D\&D++(g)&$\lambda_2=1.0$&$\lambda_2=1.0$\\
\midrule
OSD(s)&$\lambda_{osd}=0.5$&$\lambda_{osd}=0.3$\\
D\&D(s)&$\lambda_1=1.0$&$\lambda_1=0.5$\\
D\&D++(s)&$\lambda_2=1.0$&$\lambda_2=0.5$\\
\bottomrule
\vspace{-6pt}
\end{tabular}
\caption{\small  Hyperparameters for training D\&D, D\&D++ and OSD.}
  \label{supp_tab:hpdnd}
\end{table}
\subsection{D\&D and D\&D++}
In Section 4.2 of the main paper, we explain the D\&D and D\&D++ approaches in detail. We define a bias reducing classification loss $L_{br}$ to train the student network $M_s$ in step 2 as
\begin{equation}
    L_{br} = L_{class} + \lambda_1 L_{dis},
\end{equation}
where $\lambda_1$ is used to weight $L_{dis}$ in D\&D. Once $M_s$ is trained, we add another step (step 3) called D\&D++ and initialize a new student network $M^{*}_s$ with $M_s$ and train it on both categories of the binary attribute $A$. During this phase, we constrain $M^{*}_s$ to mimic the teacher $M_s$. So, we train $M^{*}_s$ using the bias reducing classification loss $L_{br}$ defined as
\begin{equation}
    L_{br} = L_{class} + \lambda_2 L_{dis},
\end{equation}
where $\lambda_2$ is used to weight $L_{dis}$ in D\&D++. We list the hyperparameters $\lambda_1$ and $\lambda_2$ in Table \ref{supp_tab:hpdnd}.
\subsection{OSD}
We construct a baseline called One Step Distillation (OSD) by skipping Step 2 of D\&D++. Here, we first train a teacher network $M_t$ that is trained on only one category of attribute $A$ (category $a_{high}$). Then, we initialize a new student network $M^{*}_s$ and train it on both attribute categories of $A$. During this phase, we constrain $M^{*}_{s}$ to mimic $M_t$. To realize this, we feed the given image to both $M^{*}_s$ and $M_t$ and obtain features $f^{*}_s$ and $f_t$, respectively and compute their cosine distance using $L_{dis}$
\begin{equation}
    L_{dis}(f_t,f^{*}_s) =1 - \frac{f_t\cdot f^{*}_s}{\|f_t\|\|f^{*}_s\|}.
\end{equation}
Combining $L_{dis}$ with $L_{class}$, we train $M^{*}_s$ with a bias reducing classification loss $L_{br}$ defined as
\begin{equation}
    L_{br} = L_{class} + \lambda_{osd} L_{dis},
\end{equation}
where $\lambda_{osd}$ is used to weight $L_{dis}$ in OSD.  The difference
between D\&D++ and OSD is that in D\&D++, $M^{*}_s$ uses a teacher ($M_s$) that has information about both of the attribute categories, whereas in OSD, $M^{*}_s$ uses a teacher ($M_s$) with information about only one category.\\

We provide the hyperparameter $\lambda_{osd}$ in Table \ref{supp_tab:hpdnd}. For all the steps in training networks in D\&D, D\&D++ and OSD, we use a batch size of 128. We train the networks for 300 epochs. We start with a learning rate of 0.1 and reduce the learning rate by 10\% after every fifty epochs. We use SGD for optimization.
\section{Detailed results}
\label{supp_sec:detailedresult}
In this section, we provide detailed versions of the results presented in the main paper.
\subsection{Results with ArcFace}
\label{supp_sec:afdetailresult}
\begin{figure}
{
\centering
\subfloat[]{\includegraphics[width=0.5\linewidth]{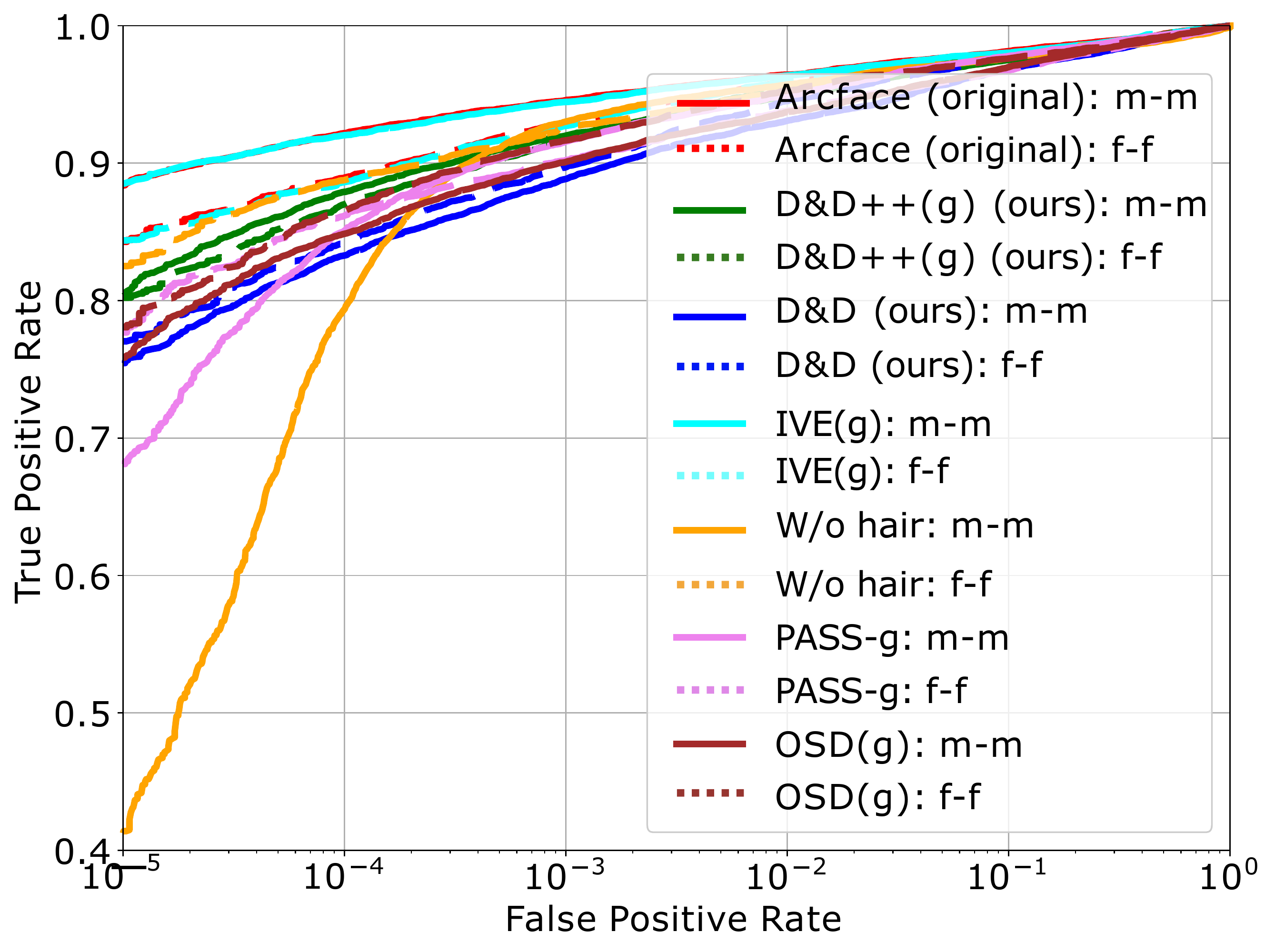}\label{supp_fig:afgwise}}
%\subfloat[Gender bias plots]{\includegraphics[width=0.5\linewidth]{images/aaai_cf_gender_bupt_dnd_barplot_all7.pdf}}
\subfloat[]{\includegraphics[width=0.5\linewidth]{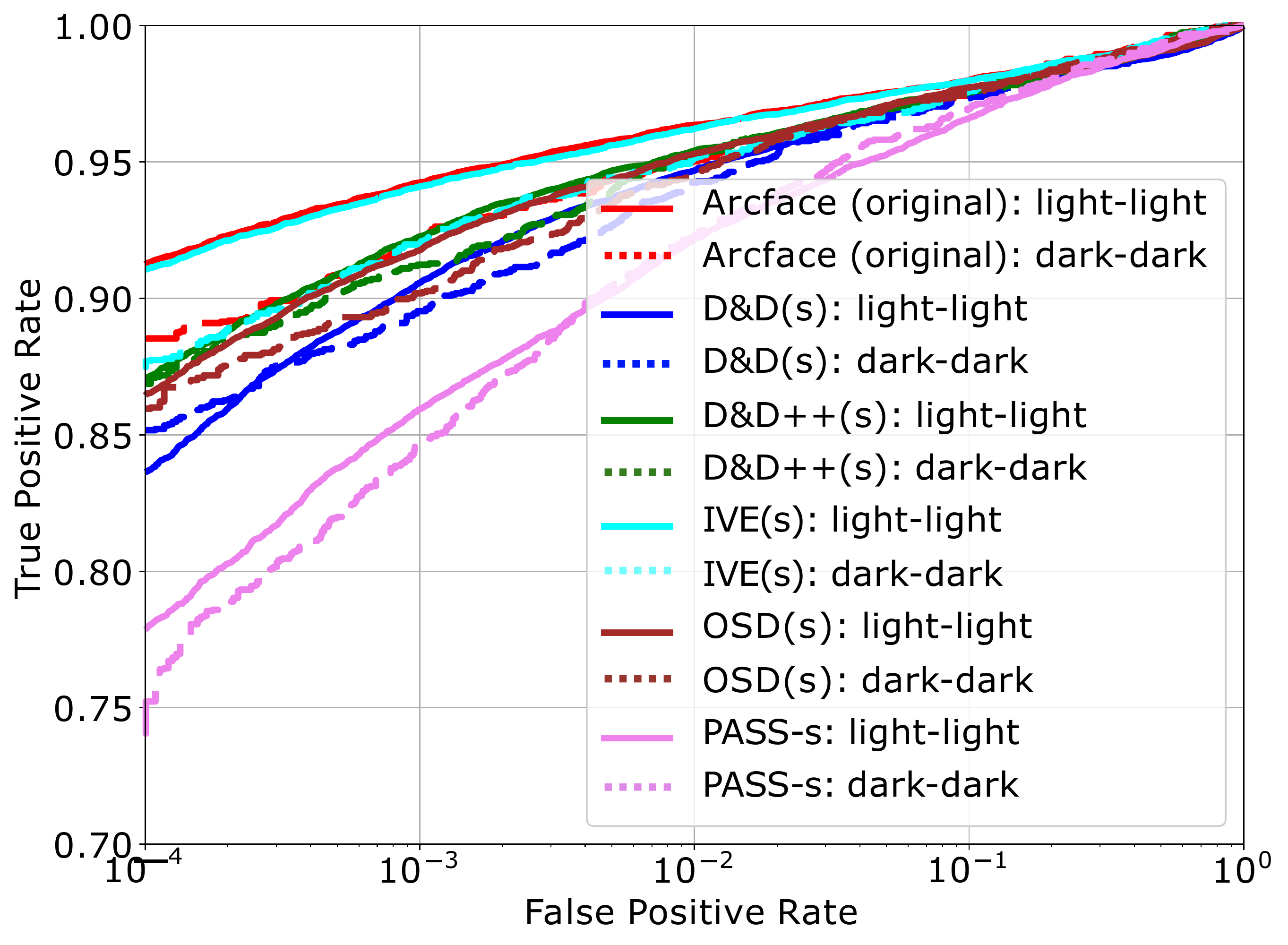}\label{supp_fig:afstwise}}
%\subfloat[Skintone bias plots]{\includegraphics[width=0.5\linewidth]{images/aaai_cf_skintone_bupt_dnd_barplot_all6_new.pdf}}
\caption{\small (a) Gender-wise verification plots for ArcFace and its gender-debiasing counterparts. `m-m'=male-male pairs, `f-f'=female-female pairs. For a given method, a \textit{high degree of separation between the male-male curve and female-female curve indicates high gender bias and vice versa}. (b) Skintone-wise verification plots for ArcFace and its skintone-debiasing counterparts. For a given method, \textit{a high degree of separation between the light-light curve and dark-dark curve indicates high skintone bias and vice versa}.}
\label{supp_fig:afgstwise}
}
\end{figure}

As mentioned in Section 5.3.2 (Table 2) of the main paper, we apply D\&D, D\&D++ and all other de-biasing baselines on the Resnet-50 version of ArcFace \cite{deng2018ArcFace}, and evaluate the gender and skintone bias reduction. In Figure \ref{supp_fig:afgstwise}, we provide the gender-wise and skintone-wise  ROCs for IJB-C, obtained using ArcFace and its debiasing counterparts. In Fig. 6 of the main paper, we also provide the verification plots for all three skintone categories (light, medium, dark) and standard deviation (STD) among these categories, obtained using ArcFace network and its skintone debiasing counterparts. Here, in Table \ref{supp_tab:arcstd}, we present the tabular values of this figure.
\begin{table*}[]
\centering
\scalebox{1.0}{
\begin{tabular}{c|ccc|ccc|ccc}
 \toprule
 FPR &    & $10^{-4}$   & &  &  $10^{-3}$   &  &  &   $10^{-2}$   &  \\
 \midrule
    Method& TPR\textsubscript{med}& Avg & STD ($\downarrow$)& TPR\textsubscript{med}& Avg & STD ($\downarrow$)&  TPR\textsubscript{med}& Avg & STD ($\downarrow$)\\
  \midrule
ArcFace & 0.883&0.893&0.014&0.921&0.928&0.009&0.954&0.956&0.006\\
IVE(s)$\dag$\cite{terhorst2019suppressing}&0.872&0.956&0.006&0.921&0.928&0.011& 0.964&0.960&0.006\\
PASS-s$\dag$\cite{Dhar_2021_ICCV} & 0.754&0.757&0.016&0.852&0.852&0.005&0.918&0.920&0.002 \\
\midrule
OSD(s) &0.861&0.861&\textbf{0.002}&0.904&0.908&0.007&0.944&0.947&0.004\\
\rowcolor{Gray}
D\&D(s)  & 0.852&0.846&0.007&0.901&0.901&\textbf{0.004}&0.939&0.943&0.003\\
\rowcolor{Gray}
D\&D++(s) & 0.867 & 0.869 & \textbf{0.002}&0.922&0.919&\textbf{0.004}&0.954&0.953&\textbf{0.001} \\
\bottomrule
\end{tabular}
}
\caption{\small Average and Standard deviation (STD) among the verification TPRs of light-light pairs, medium-medium pairs and dark-dark pairs, obtained using ArcFace and its de-biased counterparts. TPR\textsubscript{med}: medium-medium TPR. \textbf{Bold}=Best, \underline{Underlined}=Second best. \textsuperscript{$\dag$}=Our implementation of baselines. All methods are trained on BUPT-BalancedFace \cite{wang2020mitigating} data.}\label{supp_tab:arcstd} \vspace{-0.12cm}
\end{table*}
\begin{table*}[]
\centering
\scalebox{1.0}{
\begin{tabular}[h]{cccccc}
\hline
 Method/FPR & $10^{-5}$ & $10^{-4}$ & $10^{-3}$  &Training attributes &Training Dataset (\# images)\\ 
\hline
Debface-ID  &0.820 & 0.881&0.895&Race,age,gender&MS-Celeb-1M ($\sim$ 5.8 Million)\\
D\&D++(g)  &0.825&0.880&0.920&Gender&BUPT-BF ($\sim$ 1.2 Million)\\
D\&D++(s) &0.823&0.882&0.926&Race&BUPT-BF($\sim$ 1.2 Million)\\
\bottomrule
\end{tabular}
}
\caption{IJB-C verification performance of ArcFace-based D\&D++ vs. Debface\cite{gong2020jointly}. Debface-ID numbers are obatined from the original paper \cite{gong2020jointly}.}\label{supp_tab:comp}
\label{supp_tab:compsota}
\end{table*}
% \begin{figure*}
% {
% \centering
% %\vspace{-0.6cm}
% \subfloat[\scriptsize TPR@FPR=$10^{-5}$ v/s GB (CF)]{\includegraphics[width=0.25\linewidth]{images/cface_gb_1e-5.png}}
% \subfloat[\scriptsize TPR@FPR=$10^{-3}$ v/s GB (CF)]{\includegraphics[width=0.25\linewidth]{images/cface_gb_1e-3.png}}
% \subfloat[\scriptsize TPR@FPR=$10^{-4}$ v/s STB (CF)]{\includegraphics[width=0.25\linewidth]{images/cfstb_1e-4.png}}
% \subfloat[\scriptsize TPR@FPR=$10^{-2}$ v/s STB (CF)]{\includegraphics[width=0.25\linewidth]{images/cfacestb_1e-2.png}}\\
% \subfloat[\scriptsize TPR@FPR=$10^{-5}$ v/s GB (AF)]{\includegraphics[width=0.25\linewidth]{images/aface_gb_1e-5.png}}
% \subfloat[\scriptsize TPR@FPR=$10^{-3}$ v/s GB (AF)]{\includegraphics[width=0.25\linewidth]{images/arcface_gb_1e-3.png}}
% \subfloat[\scriptsize TPR@FPR=$10^{-4}$ v/s STB (AF)]{\includegraphics[width=0.25\linewidth]{images/arcface_stb_1e-4.png}}
% \subfloat[\scriptsize TPR@FPR=$10^{-2}$ v/s STB (AF)]{\includegraphics[width=0.23\linewidth]{images/arcface_stb_1e-2.png}}
% \caption{TPR at a fixed FPR \textbf{(a,b)} v/s Gender bias(GB) in Crystalface(CF) based models. \textbf{(c,d)} v/s Skintone bias(STB) in CF-based models. \textbf{(e,f)} v/s GB in ArcFace(AF) based models. \textbf{(g,h)} v/s STB in AF-based models.}\label{supp_fig:tradeoffsupp}}

% \end{figure*}

\subsubsection{Comparison with Debface \cite{gong2020jointly}:} In Table 2 of the main paper we compare D\&D++ with other methods including the recently proposed adversarial method PASS \cite{Dhar_2021_ICCV}, and show that D\&D++ consistently obtains higher face verification performance and lower bias than PASS. We note that DebFaceID \cite{gong2020jointly} is another adversarial method proposed for removing protected attributes like gender and race from face representations, that uses a ResNet50 ArcFace backbone (similar to our ArcFace based D\&D++). For the IJB-C dataset, this work reports the overall face verification performance. So, in Table \ref{supp_tab:compsota}, we compare the overall face verification performance obtained by D\&D++ with that obtained by DebfaceID on IJB-C. D\&D++ obtains higher face verication performance than DebfaceID at most FPRs. It should be noted that DebfaceID uses a cleaned version of MS-Celeb-1M (MS1M) dataset \cite{guo2016ms} (provided by \cite{deng2018ArcFace}) for training, which consists of approximately 5.8 million images, whereas  D\&D-based systems are trained on the BUPT-BalancedFace dataset that consists of $\sim$ 1.2 million images. We do not use MS1M dataset as it does not contain race labels. On the other hand, BUPT-BalancedFace \cite{wang2020mitigating} contains race labels, making it easier to train skintone-debiasing models.
% \subsection{Bias v/s face verification performance}
% \label{supp_sec:biasvsperf}
% An ideal de-biasing system must reduce bias and maintain reasonable face verification performance. In Figure 8 of the main paper, we show the TPR at a fixed FPR and gender/skintone bias for several de-biasing methods, including our proposed methods. We fixed the FPR=$10^{-4}$ for gender de-biasing methods, and FPR=$10^{-3}$ for skintone de-biasing methods. Here in Figure \ref{supp_fig:tradeoffsupp}, we provide the same plots for all the other FPRs. In these plots, an ideal de-biasing method would occupy the upper-left-hand corner, where the face verification performance is high, and bias is low, whereas in the worst case scenario, the de-biasing method would occupy the lower left corner which indicates high bias and low face verification performance.
\begin{figure}
{
\centering
\subfloat[]{\includegraphics[width=0.5\linewidth]{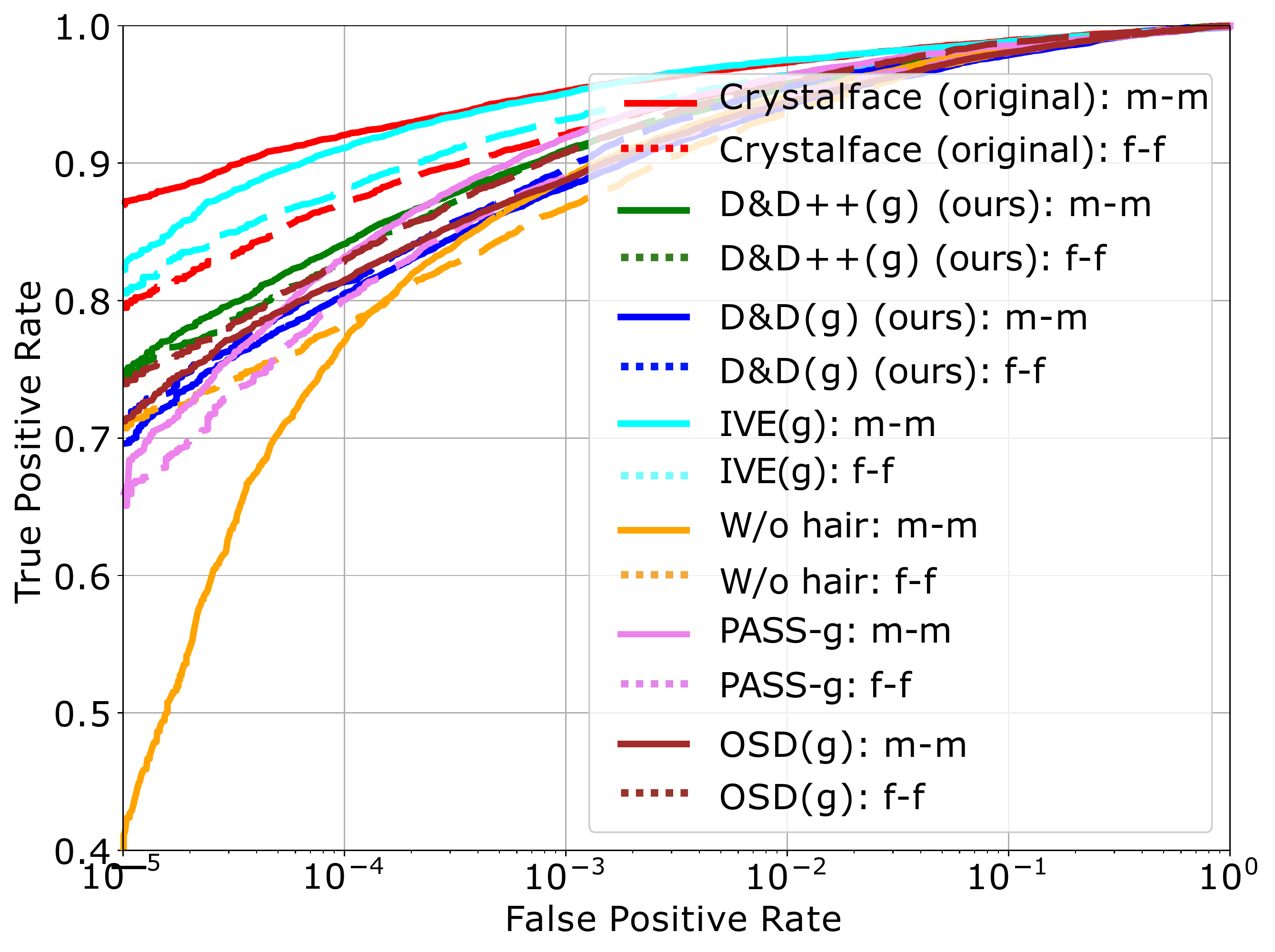}\label{supp_fig:cfgwise}}
%\subfloat[Gender bias plots]{\includegraphics[width=0.5\linewidth]{images/aaai_cf_gender_bupt_dnd_barplot_all7.pdf}}
\subfloat[]{\includegraphics[width=0.5\linewidth]{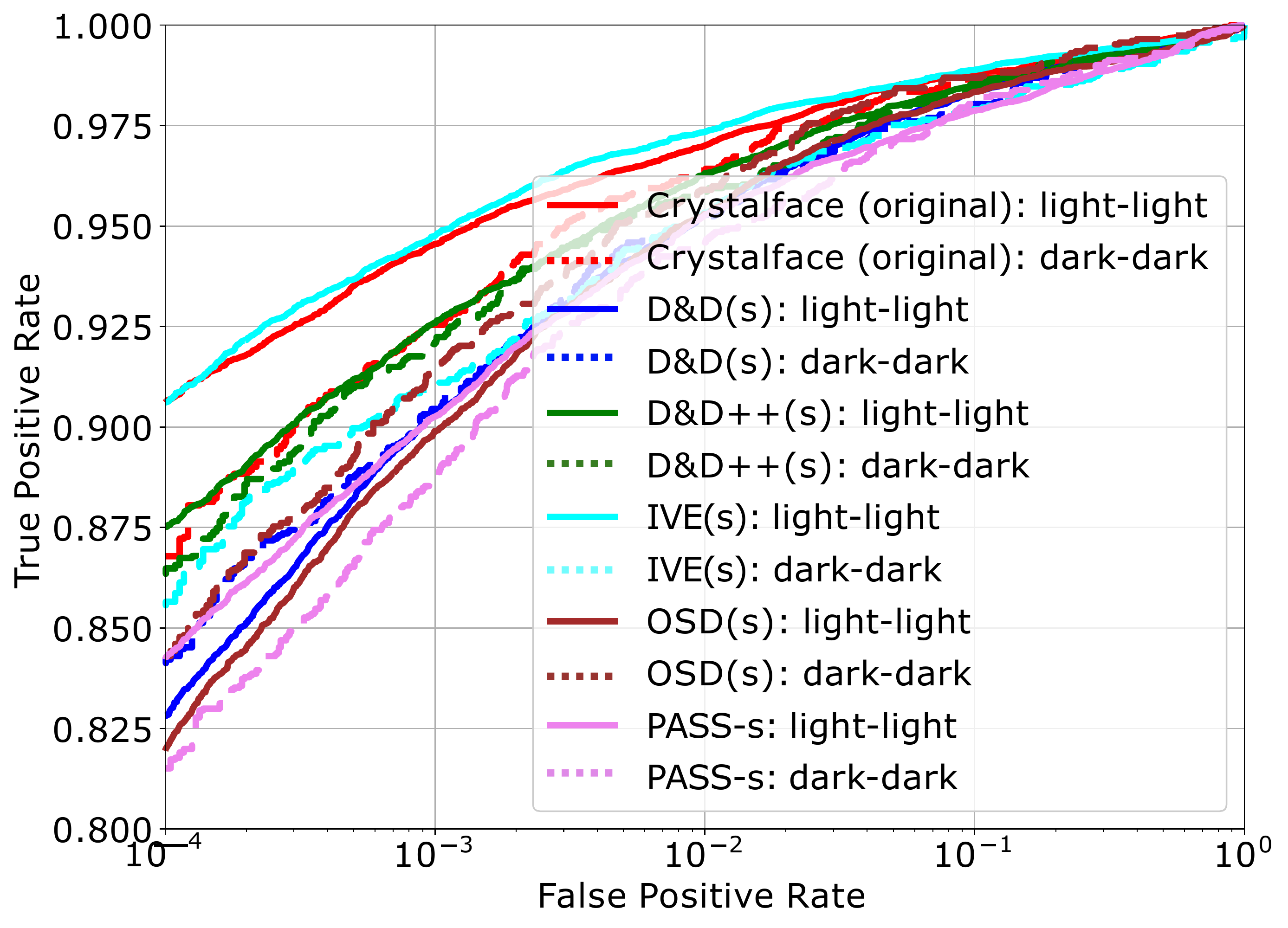}\label{supp_fig:cfstwise}}
%\subfloat[Skintone bias plots]{\includegraphics[width=0.5\linewidth]{images/aaai_cf_skintone_bupt_dnd_barplot_all6_new.pdf}}
\caption{\small (a) Gender-wise verification plots for Crystalface and its gender-debiasing counterparts. `m-m'=male-male pairs, `f-f'=female-female pairs.  For a given method, \textit{a high degree of separation between the male-male curve and female-female curve indicates high gender bias and vice versa}. (b) Skintone-wise verification plots for Crystalface and its skintone-debiasing counterparts. For a given method, \textit{a high degree of separation between the light-light curve and dark-dark curve indicates high skintone bias and vice versa}.}\label{supp_fig:cfwise}
\label{supp_fig:cfgstwise}
}
\end{figure}
\subsection{Results with Crystalface}
\label{supp_sec:cfdetailresult}
For evaluating the generalizability of D\&D, D\&D++ and other baselines, we implement all the methods using the Crystalface \cite{ranjan2019fast} backbone and present the results in Section 5.4 of the main paper. Here, in Tables \ref{supp_tab:cfgenbias},\ref{supp_tab:cfstbias} we extend Tables 3a, 3b (respectively) from the main paper. We also provide the gender-wise and skintone-wise verification ROCs for IJB-C, obtained using all of these methods in Figure \ref{supp_fig:cfgstwise}. In Fig. 7 of the main paper, we also provide the verification plots for all three skintone categories (light, medium, dark) and standard deviation (STD) among these categories, obtained using Crystalface network and its skintone debiasing counterparts. Here, in Table \ref{supp_tab:crystd}, we present the tabular values of this figure.
\begin{table*}[]
\centering
%\scriptsize
\subfloat[Gender bias - Crystalface backbone]{\scalebox{0.73}{
\begin{tabular}{c|ccccc|ccccc|ccccc}
\toprule
 FPR &  &  & $10^{-5}$ &  & &  & &  $10^{-4}$ &  &  &  &  & $10^{-3}$ &  &  \\
 \midrule
  Method & TPR & TPR\textsubscript{m} & TPR\textsubscript{f} & Bias$(\downarrow)$ \hspace{-4pt} & \hspace{-4pt} BPC\textsubscript{g}$(\uparrow)$ \hspace{-4pt} & TPR & TPR\textsubscript{m} & TPR\textsubscript{f} & Bias$(\downarrow)$ \hspace{-4pt} & \hspace{-4pt} BPC\textsubscript{g}$(\uparrow)$ \hspace{-4pt} & TPR & TPR\textsubscript{m} & TPR\textsubscript{f} & Bias$(\downarrow)$ \hspace{-4pt} & \hspace{-4pt} BPC\textsubscript{g}$(\uparrow)$ \hspace{-4pt} \\
 \midrule
Crystalface \hspace{-6pt}&  0.856 &  0.869 & 0.794 & 0.075 & 0 & 0.912 &  0.920 & 0.871 &   0.049 & 0 &  0.950& 0.953& 0.921 & 0.031 & 0 \\
IVE(g)\textsuperscript{$\dag$} \cite{terhorst2019suppressing} & 0.840 & 0.820 & 0.804 & 0.016 & \underline{0.768} & 0.910 & 0.911 & 0.880 & 0.031 & 0.365 & 0.952 & 0.951 & 0.932 & 0.019 & 0.389 \\
W/o hair\textsuperscript{$\dag$} \cite{albiero2020face} & 0.592 & 0.396 & 0.706 & 0.310 & -3.441 & 0.803 & 0.770 & 0.783 & 0.013 & 0.615 & 0.899 & 0.888 & 0.868 & 0.020 & 0.301 \\
PASS-g\textsuperscript{$\dag$} \cite{Dhar_2021_ICCV} &0.691  &0.656 &0.647  &\underline{0.009}  & 0.687& 0.842 &0.832  &0.800  &0.031  &0.291  &0.914  &0.918  &0.890 &0.029  &  0.027 \\
\midrule
OSD(g) & 0.721 & 0.712 & 0.738 & 0.027 & 0.482 & 0.817 & 0.815 & 0.828 & 0.013 & 0.631& 0.895 & 0.888 & 0.908 & 0.020 &0.297  \\
\rowcolor{Gray}
D\&D(g) & 0.705 & 0.693 & 0.706 & 0.013 & 0.650 & 0.805 & 0.805 & 0.813 & \textbf{0.008} & \textbf{0.719} & 0.888 & 0.883 & 0.896 & \underline{0.013} & \underline{0.515} \\
\rowcolor{Gray}
D\&D++(g) & 0.754 & 0.744 & 0.741 & \textbf{0.002} & \textbf{0.854} & 0.844 & 0.841 & 0.830 & \underline{0.011} & \underline{0.701}& 0.914 & 0.910 & 0.907 & \textbf{0.002} & \textbf{0.898}\\
\bottomrule
\end{tabular}
}\label{supp_tab:cfgenbias}}\\
\subfloat[Skintone bias - Crystalface backbone]{\scalebox{0.73}{
\begin{tabular}{c|ccccc|ccccc|ccccc}
 \toprule
 FPR &  &  & $10^{-4}$ &  & &  & &  $10^{-3}$ &  &  &  &  & $10^{-2}$ &  &  \\
 \midrule
    Method & TPR & TPR\textsubscript{l} & TPR\textsubscript{d} & Bias$(\downarrow)$ \hspace{-4pt} & \hspace{-4pt} BPC\textsubscript{st}$(\uparrow)$ \hspace{-4pt} & TPR & TPR\textsubscript{l} & TPR\textsubscript{d} & Bias$(\downarrow)$ \hspace{-4pt} & \hspace{-4pt} BPC\textsubscript{st}$(\uparrow)$ \hspace{-4pt} & TPR & TPR\textsubscript{l} & TPR\textsubscript{d} & Bias$(\downarrow)$ \hspace{-4pt} & \hspace{-4pt} BPC\textsubscript{st}$(\uparrow)$ \hspace{-4pt} \\
  \midrule
Crystalface & 0.912 & 0.906 & 0.867 & 0.038 & 0 &  0.950& 0.945 & 0.925 & 0.020 & 0 & 0.973 & 0.970 & 0.963 & 0.006 & 0 \\
IVE(s)\textsuperscript{$\dag$} \cite{terhorst2019suppressing} & 0.910 & 0.906 & 0.854 & 0.072 &-0.371 &0.950 & 0.948 & 0.909 & 0.038 & -0.900 &0.974  & 0.974 & 0.953 & 0.021 & -2.49 \\
PASS-s\textsuperscript{$\dag$} \cite{Dhar_2021_ICCV} & 0.851 & 0.842& 0.815 & 0.027 & 0.222 & 0.910 & 0.903 &0.886  & 0.016 & 0.158 &0.953  &0.953 & 0.946 & 0.007 & -0.187 \\
\midrule
OSD(s) & 0.848 & 0.819 & 0.841 & 0.022 & 0.351 & 0.916 & 0.899 & 0.913 & 0.015 & 0.214 & 0.961 & 0.953 & 0.959 & 0.006 & -0.012 \\
\rowcolor{Gray}
D\&D(s) & 0.850 & 0.828 & 0.839 & \textbf{0.011} & \textbf{0.643}& 0.916 & 0.903 & 0.904 & \textbf{0.001} & \textbf{0.914} & 0.961 & 0.953 & 0.952 & \textbf{0.001} &\textbf{0.821} \\
\rowcolor{Gray}
D\&D++(s) & 0.886 & 0.875 & 0.862 & \underline{0.013} & \underline{0.629} &  0.934& 0.926 & 0.921 & \underline{0.005} & \underline{0.733} & 0.967 & 0.963 & 0.959 & \underline{0.004} &\underline{0.327}\\
\bottomrule
\end{tabular}}\label{supp_tab:cfstbias}}
\caption{\small Bias analysis for \textit{Crystalface} network, and its de-biased counterparts on IJB-C. TPR: overall True Positive rate, TPR\textsubscript{m}: male-male TPR, TPR\textsubscript{f}: female-female TPR. TPR\textsubscript{l}: light-light TPR, TPR\textsubscript{d}: dark-dark TPR. \textbf{Bold}=Best, \underline{Underlined}=Second best. D\&D variants obtain higher BPC and lower gender bias at most FPRs. \textsuperscript{$\dag$}=Our implementation of baselines (See Sections \ref{supp_sec:passinfo}, \ref{supp_sec:iveinfo}, \ref{supp_sec:bowyerinfo} for details). All methods are trained on BUPT-BalancedFace \cite{wang2020mitigating} data.} \label{supp_tab:cfallbias}
\vspace{-0.2cm}
\end{table*}
\begin{table*}[]
\centering
\scalebox{1.0}{
\begin{tabular}{c|ccc|ccc|ccc}
 \toprule
 FPR   &  & $10^{-4}$   & &  &   $10^{-3}$  &  &  &   $10^{-2}$   &  \\
 \midrule
    Method&  TPR\textsubscript{med} & Avg & STD ($\downarrow$)&  TPR\textsubscript{med}&  Avg & STD ($\downarrow$) & TPR\textsubscript{med}& Avg & STD ($\downarrow$)\\
  \midrule
Crystalface  &0.906  &0.893 & 0.018 &0.939 &0.936  & 0.008  &0.968  &0.967 &  0.003\\
IVE(s)$\dag$\cite{terhorst2019suppressing}   &0.889  &0.883 & 0.022  & 0.941 & 0.933&0.017  &0.967  &0.965  &0.009  \\
PASS-s)$\dag$\cite{Dhar_2021_ICCV}  &0.844 &0.834  &0.013   &0.904   &0.898  &  0.008  &0.946  &0.948  & 0.003 \\
\midrule
OSD(s)  &0.834 & 0.831 &\underline{0.009}  & 0.899 &0.904 &  \underline{0.007}  &0.947  & 0.953 & 0.005 \\
\rowcolor{Gray}
D\&D(s)   &0.829 &0.832  &\textbf{0.005}   &0.889  &0.899 &\underline{0.007}   &0.948  & 0.951&\textbf{0.002} \\
\rowcolor{Gray}
D\&D++(s) & 0.888 &0.875 &0.011  & 0.927 & 0.925 & \textbf{0.003}  &0.963 & 0.962 &\textbf{0.002}\\
\bottomrule
\end{tabular}
}
\caption{\small Average and Standard deviation (STD) among the verification TPRs of light-light pairs, medium-medium pairs and dark-dark pairs, obtained using Crystalface and its de-biased counterparts. TPR\textsubscript{med}: medium-medium TPR. \textbf{Bold}=Best, \underline{Underlined}=Second best. D\&D variants obtain the lowest STD (bias) among the performance of the three skintones. \textsuperscript{$\dag$}=Our implementation of baselines. All methods are trained on BUPT-BalancedFace \cite{wang2020mitigating} data. \vspace{-0.8cm}}\label{supp_tab:crystd}
\end{table*}
\section{Training details for PASS \cite{Dhar_2021_ICCV}}
\label{supp_sec:passinfo}
\subsection{Brief summary of PASS}
PASS \cite{Dhar_2021_ICCV} is composed of three components:\\
(1) \textbf{Generator model $M$}: A model that accepts face recognition feature $f_{in}$ from a pre-trained network, and generates a lower dimensional feature $f_{out}$ that is supposed to be agnostic to sensitive attribute (gender or skintone). $M$ consists of a single linear layer, followed by a PReLU \cite{he2015deep} layer. \\
(2) \textbf{Classifier} $C$: A classifier that takes in $f_{out}$ and generates a prediction vector for identity classification.\\
(3) \textbf{Ensemble of attribute classifiers $E$}: An ensemble of $K$ attribute prediction models. Each of these models is a two layer MLP with 128 and 64  hidden units respectively with SELU activations, followed by a sigmoid activated output layer with $N_{att}$ units,  where $N_{att}$ = the number of classes in the attribute being considered.

 We use the official implementation of PASS \cite{passcode} to build PASS-g (for reducing gender information in face recognition feature) and PASS-s (for reducing skintone information). We provide a brief summary of PASS training, and specify the hyperparameters used. More details are provided on the original paper \cite{Dhar_2021_ICCV} \\
\textbf{Stage 1 - Initializing and training $M$ and $C$}: Using input features $f_{in}$ from a pre-trained network, we train $M$ and $C$ from scratch for $T_{fc}$ iterations using $L_{class}$. $L_{class}$ is a the standard cross entropy classification loss. The learning rate used to train $M$ and $C$ in this stage is denoted by $\alpha_1$. \\
\textbf{Stage 2 - Initializing and training $E$}: Once $M$ is trained to perform classification, we feed the outputs $f_{out}$ of $M$ to ensemble $E$ of $K$ attribute prediction models. $E$ is then trained to classify attribute for $T_{atrain}$ iterations using $L_{att}$. $L_{att}$ is a cross-entropy classification loss for classifying attributes. The learning rate used to train the models in $E$ in this stage is denoted by $\alpha_2$. Model $M$ remains frozen in this step.\\ 
\textbf{Stage 3 - Update model $M$ and classifier $C$}: Here, $M$ is trained to generate features $f_{out}$ that can classify identities and have reduced encoding of  sensitive attribute under consideration. We feed $f_{out}$ to $E$ and $C$, the outputs of which result in an adversarial de-biasing loss $L_{deb}$ and $L_{class}$ respectively. We combine them to compute the bias reducing classification loss in PASS denoted as $L^{(PASS)}_{br}$
\begin{equation}
    L^{(PASS)}_{br} = L_{class} + \lambda L_{deb},
\end{equation}
$L^{(PASS)}_{br}$ is used for training $M$ and $C$ for $T_{deb}$ iterations, while $E$ remains locked. $\lambda$ is used to weight the adversarial loss $L_{deb}$. The learning rate used to train $M$ and $C$ in this stage is denoted by $\alpha_3$.\\
\textbf{Stage 4 - Update ensemble $E$ (discriminator)}:
% In stage 3, $M$ is trained to generate attribute-debiased descriptors $f_{out}$ to fool the models in $E$, whereas in stage 4
In stage 4, members of $E$ are trained to classify attribute using $f_{out}$. So, stages 3 and 4 are run alternatively, for $T_{ep}$ episodes, after which all the models in $E$ are re-initialized and re-trained (as done in stage 2). Here, one episode indicates an instance of running stages 3 and 4 consecutively. In stage 4, following the discriminator-training strategy introduced by \cite{Dhar_2021_ICCV}, we choose one of the models in $E$, and train it for $T_{plat}$ iterations or until it reaches an accuracy of $A^*$ on the validation set. $M$ and $C$ remain frozen in this stage.
\begin{table}%
  \small
  \centering
\begin{tabular}{cccc|cc}
\toprule
Backbone  & &\multicolumn{2}{c|}{ArcFace} & \multicolumn{2}{c}{Crystalface}\\
\midrule
Hyperparam & Stage & PASS-g & PASS-s& PASS-g & PASS-s \\
\midrule
$\lambda$ & 3&10&10&1&10\\
$K$ & 2, 3, 4 &3&2&4&2\\
$T_{fc}$ & 1  &10000&10000&16000&16000\\
$T_{deb}$ & 3 &1200&1200&1200&1200\\
$T_{atrain}$ & 2 &30000&30000&30000&30000\\
$T_{plat}$ & 4  &2000&2000&2000&2000\\
$A^*$ &  4&0.95&0.95&0.90&0.95\\
$\alpha_1$ & 1 &$10^{-2}$&$10^{-2}$&$10^{-2}$&$10^{-2}$\\
$\alpha_2$ &  2,4&$10^{-3}$&$10^{-3}$&$10^{-3}$&$10^{-3}$\\
$\alpha_3$ &  3&$10^{-4}$&$10^{-4}$&$10^{-4}$&$10^{-4}$\\
$T_{ep}$ & 3,4 &40&40&40&40\\
\bottomrule
\vspace{-6pt}
\end{tabular}
\caption{\small  Hyperparameters for training PASS-g and PASS-s on ArcFace and Crystalface features}
  \label{supp_tab:hppass}
  \vspace{-11pt}
\end{table}
\subsection{Datasets and Hyperparameters for PASS}
\begin{figure}
{
\centering
\subfloat[]{\includegraphics[width=0.5\linewidth]{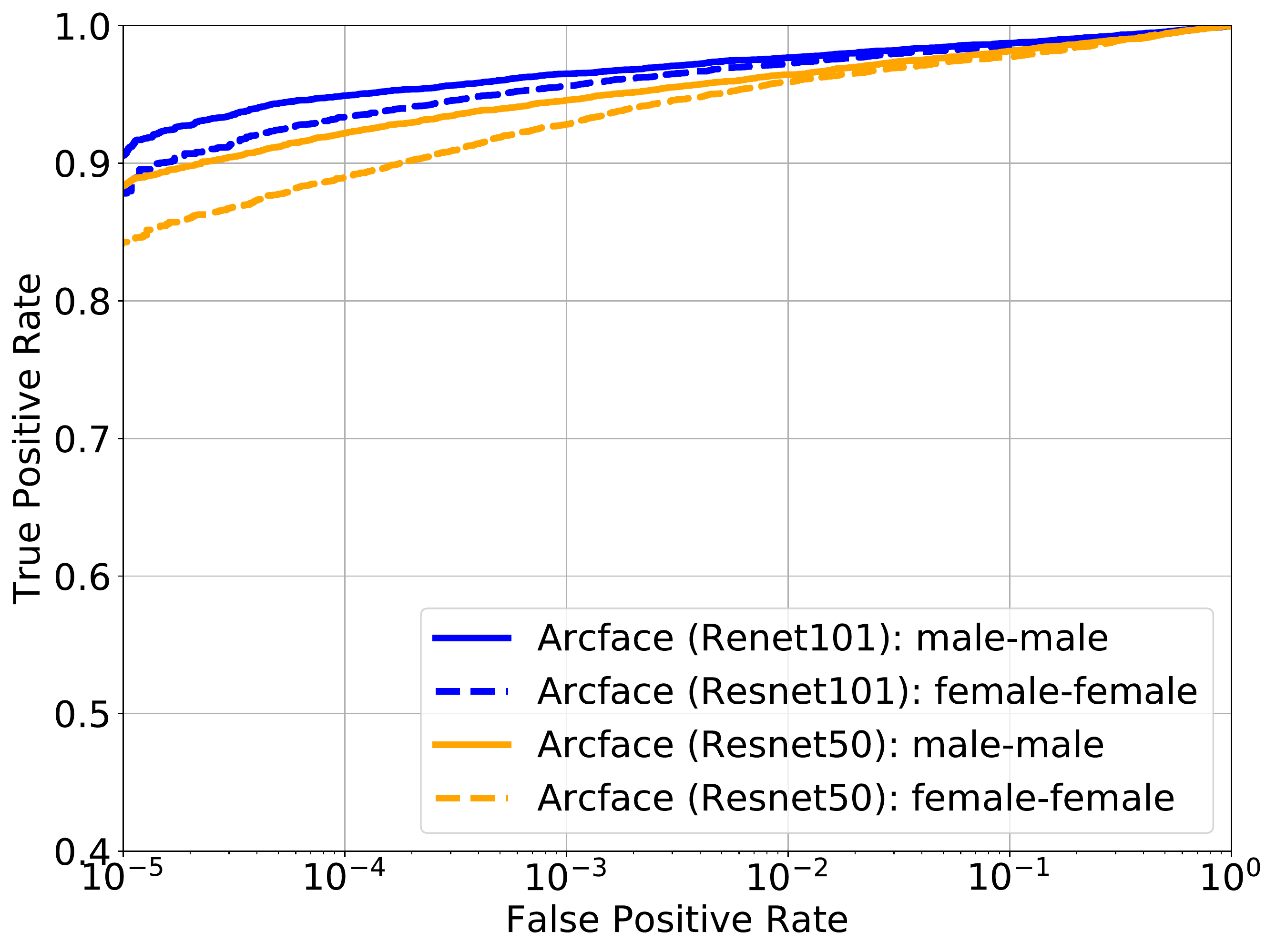}\label{supp_fig:gen50vs100}}~
\subfloat[]{\includegraphics[width=0.5\linewidth]{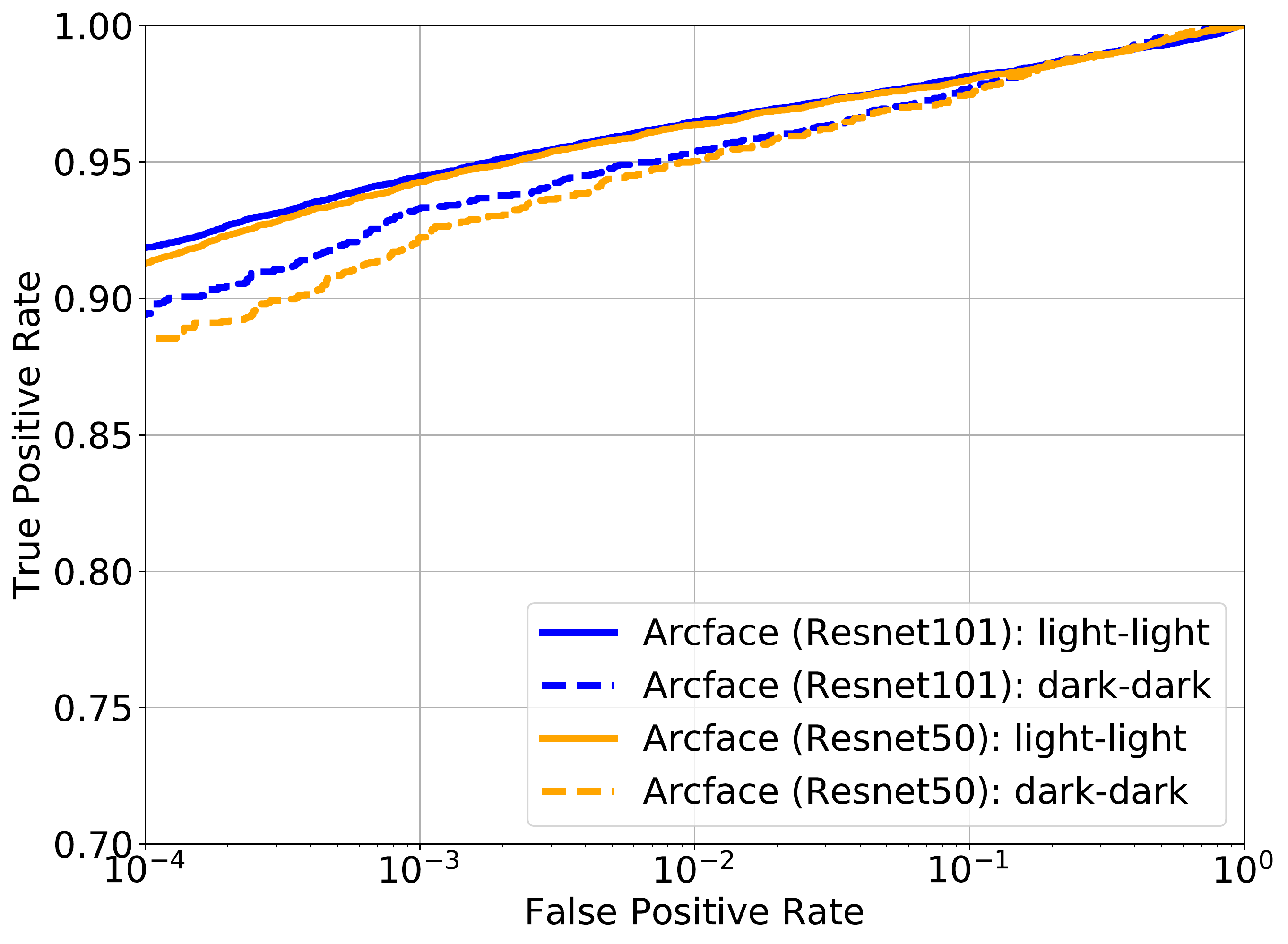}\label{supp_fig:st50vs100}}
\vspace{-1em}
\caption{\small (a) Gender-wise and (b) Skintone-wise verification ROCs on the IJB-C dataset, for Resnet50 and Resnet101 version of the ArcFace networks, trained on BUPT-BalancedFace dataset. }
%\vspace{-0.7cm}
\label{supp_fig:50vs100}
}
\end{figure}

\begin{figure}
{
\centering
\subfloat[]{\includegraphics[width=0.5\linewidth]{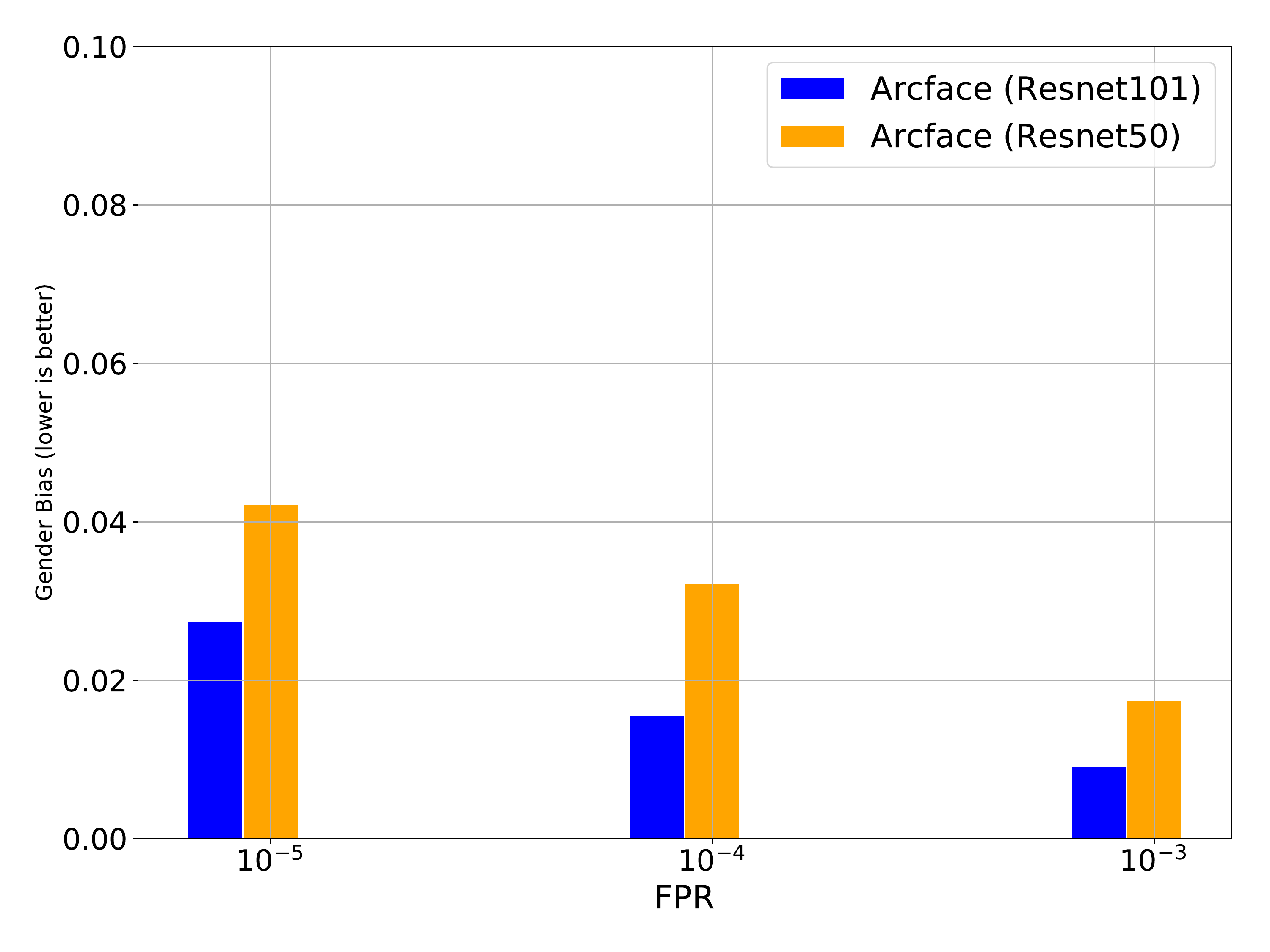}\label{supp_fig:gen50vs100bar}}~
\subfloat[]{\includegraphics[width=0.5\linewidth]{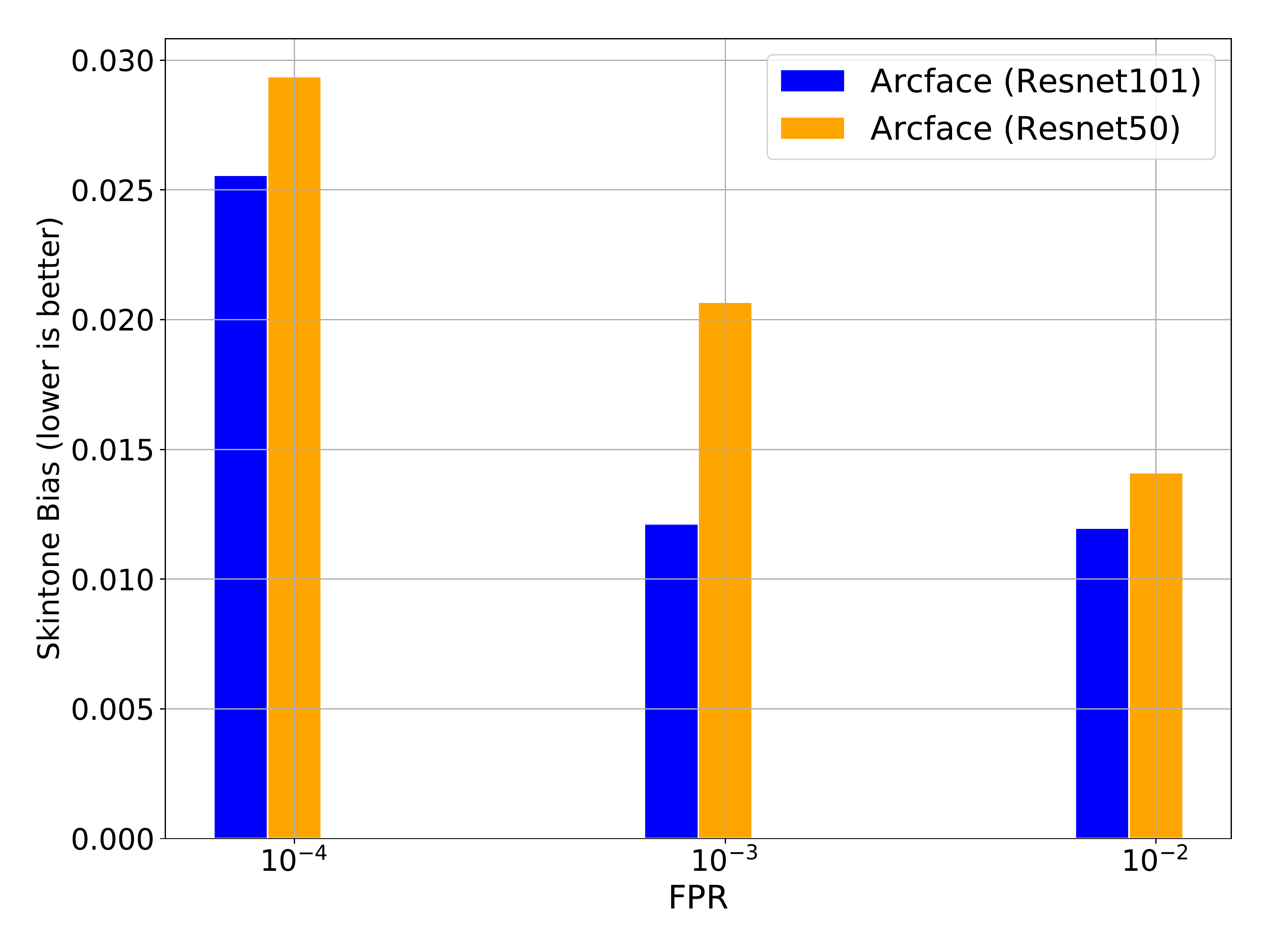}\label{supp_fig:st50vs100bar}}
\vspace{-1em}
\caption{\small (a) Gender bias and (b) Skintone bias on the IJB-C dataset, for Resnet50 and Resnet101 version of the ArcFace networks, trained on BUPT-BalancedFace dataset. }
%\vspace{-0.7cm}
\label{supp_fig:50vs100bias}
}
\end{figure}

In the original paper \cite{Dhar_2021_ICCV}, $f_{in}$ is obtained from a pre-trained ArcFace network that has been trained on the MS1MV2 \cite{ms1mv2} dataset. We note that the authors of PASS \cite{Dhar_2021_ICCV} perform experiments using the ResNet 101 version of the ArcFace network. But, in our preliminary experiments, we found that the ResNet50 version of ArcFace network demonstrates more gender and skintone bias, as shown in Figures \ref{supp_fig:50vs100} and \ref{supp_fig:50vs100bias}. With this reasoning, and following some other previous works \cite{gong2020jointly,gac}, we use the Resnet 50 version of ArcFace in our experiments, which is unlike the experiments in PASS\cite{Dhar_2021_ICCV}.

Also, in the original PASS \cite{Dhar_2021_ICCV} paper, PASS-g is trained on a mixture of UMDFaces\cite{bansal2017umdfaces}, UMDFaces-Videos\cite{bansal2017s} and MS1M \cite{guo2016ms}. However, due to the current unavailability of the UMDFaces and UMDFaces-Videos dataset and to make PASS variants comparable with D\&D (and OSD) variants,  we obtain $f_{in}$ from a ArcFace network that has been trained on the BUPT-BalancedFace \cite{wang2020mitigating}, following which we train both PASS-g and PASS-s using the BUPT-BalancedFace dataset as well.  

We note that the authors of PASS\cite{Dhar_2021_ICCV} also perform experiments on Crystalface \cite{ranjan2019fast} trained on the aforementioned  `mixture' dataset. Due to the current unavailability of this dataset, in our implementation, we extract $f_{in}$ using a pre-trained Crystalface network trained on BUPT-BalancedFace dataset. Following that, we train both PASS-g and PASS-s using the BUPT-BalancedFace dataset as well.

We use the same hyperparameters specified in the original paper \cite{Dhar_2021_ICCV} and present them in Table \ref{supp_tab:hppass}.  We use a batch size of 400 in all these experiments. In our work, we use the official implementation of PASS \cite{passcode}.
\section{Training details for IVE \cite{terhorst2019suppressing}}
\label{supp_sec:iveinfo}
IVE \cite{terhorst2019suppressing} is an attribute suppression algorithm that assigns a score to each variable in face representations using a decision tree ensemble. This score of a variable indicates the importance of that variable for a specific recognition task. Variables that affect attribute classification considerably are then excluded from the representation. In every exclusion step, $n_e$ variables are removed from the representation. The algorithm is run for $n_s$ steps, thus resulting in exclusion of $n_s \times n_e$ variables from the representation. We follow the re-implementation of IVE by \cite{Dhar_2021_ICCV} and construct two variants of IVE: IVE(g) for reducing gender information and IVE(s) for reducing skintone information. \\

IVE is a feature-based system that reduces information of sensitive attribute from features obtained using a pre-trained network (like ArcFace or Crystalface). So before training IVE, we first train a Resnet 50 version of the ArcFace network on the BUPT-BalancedFace dataset. After that, we train IVE(g) and IVE(s) (separately) as follows:\\
\textbf{Training IVE(g)} : We extract ArcFace features for the images in BUPT-BalancedFace dataset. We also obtain the gender labels for these images using \cite{ranjan2017all}. Then we use the IVE system (explained in \cite{terhorst2019suppressing}) to remove variables in the features that encode gender information. During inference, we use the trained IVE(g) system to transform the ArcFace features extracted for the evaluation dataset (IJB-C).\\
\textbf{Training IVE(s)}: We follow the same experimental setup for training IVE(s). The only difference is that in IVE(s), instead of gender labels, we feed the race label (already provided in the BUPT-BalancedFace dataset) alongwith the ArcFace features extracted for the images in BUPT-BalancedFace dataset. For inference, we use the trained IVE(s) system to transform ArcFace features for IJB-C.\\ 

We perform the same experiment by replacing the pre-trained ArcFace network with a Crystalface network trained on BUPT-BalancedFace dataset, for our Crystalface-based experiments. The official implementation for training IVE is publicly available \cite{ivecode}. In all of our IVE experiments, we use the parameters values mentioned in the code, i.e. $n_s=20$ and $n_e=5$, thus resulting in 100 eliminations. Since face recognition features from ArcFace or Crystalface are 512-dimensional, the trained IVE(s/g) framework transforms the input features for test images into 412 dimensional features, which are then used to perform face verification.
\begin{figure}
\centering
{\includegraphics[width=\linewidth]{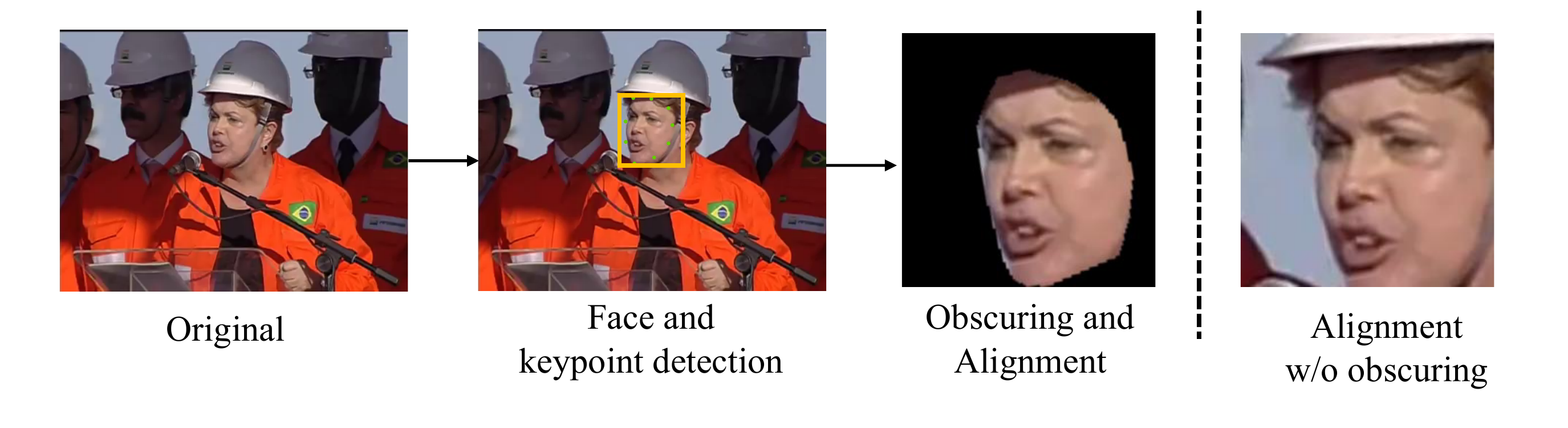}}
%\vspace{-1em}
\caption{\small Our method for obscuring hair (Similar to \cite{albiero2020face}). On the right, we show an aligned image without obscuring hair.}
\label{supp_fig:bowyerbaseline}
\vspace{-6pt}
\end{figure}
\section{Pipeline for obscuring hair}
\label{supp_sec:bowyerinfo}
In \cite{albiero2020face}, the authors obscure hair regions of images in the evaluation dataset. This is done to get an equal fraction of pixels in the images for each gender. The authors use a segmentation network \cite{yu2018bisenet} to obscure the hair.  As a result of obscuring hair, it is shown that the resulting face recognition features extracted using ArcFace demonstrate lower gender bias.  However, as pointed out by \cite{Dhar_2021_ICCV}, such experiments are only performed on datasets with clean frontal faces in MORPH \cite{ricanek2006morph} and Notre-Dame \cite{phillips2005overview} datasets.  But, complex datasets like IJB-C contain varied and cluttered poses, which is why segmentation cannot be used (especially for images with extreme poses). So, following \cite{Dhar_2021_ICCV}, we compute the face border keypoints using \cite{ranjan2017all} and obscure all the regions outside the polygon formed by these keypoints. Our hair obscuring pipeline is presented in Fig \ref{supp_fig:bowyerbaseline}. Note that, this baseline method cannot be used for mitigating skintone bias. After obscuring hair regions for images in the IJB-C dataset, we extract their features using pre-trained Crystalface/ArcFace networks trained on BUPT-BalancedFace, and perform 1:1 face verification.

\end{document}